\documentclass[11pt,a4paper]{article}

\usepackage{arxiv}

\usepackage{amsmath,amssymb,amsthm}

\usepackage{graphicx}
\usepackage{booktabs,array,longtable,multirow,tabularx}
\usepackage{xcolor}
\usepackage{tikz}
\usetikzlibrary{arrows.meta,positioning,shapes.geometric,shapes.symbols,fit,backgrounds,calc,decorations.pathreplacing}

\usepackage{enumitem}
\setlist[itemize]{topsep=4pt,itemsep=2pt,parsep=0pt,partopsep=0pt}
\setlist[enumerate]{topsep=4pt,itemsep=2pt,parsep=0pt,partopsep=0pt}

\usepackage{hyperref}
\usepackage{url}
\hypersetup{
  colorlinks=true,
  linkcolor=black,
  urlcolor=blue!60!black,
  citecolor=blue!60!black,
  breaklinks=true,
}

\definecolor{ThesisBlue}{HTML}{1F4E79}
\definecolor{TakeawayOrange}{HTML}{C55A11}
\definecolor{SoftGray}{HTML}{F2F2F2}

\newcommand{\thesisbox}[2]{%
  \par\medskip\noindent
  \fcolorbox{ThesisBlue}{SoftGray!40!white}{%
    \begin{minipage}{\dimexpr\linewidth-2\fboxsep-2\fboxrule}
      {\color{ThesisBlue}\bfseries #1}\par\smallskip #2
    \end{minipage}%
  }\par\medskip
}
\newcommand{\takeaway}[1]{%
  \par\medskip\noindent
  \fcolorbox{TakeawayOrange}{SoftGray!40!white}{%
    \begin{minipage}{\dimexpr\linewidth-2\fboxsep-2\fboxrule}
      {\color{TakeawayOrange}\bfseries Takeaway.\ }#1
    \end{minipage}%
  }\par\medskip
}

\newcommand{\eg}{e.g.,\ }

\newcommand{\artifactrepo}{\url{https://github.com/xxzcc/awesome-llm-mas-rl}}

\title{
  \textbf{Reinforcement Learning for LLM-based Multi-Agent Systems} \\[4pt]
  \large through Orchestration Traces
}
\author{
  Chenchen Zhang \\[2pt]
  \small Independent Researcher \\
  \small \texttt{zcc1959339538@gmail.com}
}
\date{May 2026}

\begin{document}
\sloppy           
\maketitle

\begin{abstract}
\noindent
As large language model (LLM) agents evolve from isolated tool users
into coordinated teams, reinforcement learning (RL) must optimize not
only individual actions, but also how work is spawned, delegated,
communicated, aggregated, and stopped. This paper studies RL for
LLM-based multi-agent systems through \emph{orchestration traces}:
temporal interaction graphs whose events include sub-agent spawning,
delegation, communication, tool use, return, aggregation, and stopping
decisions. The trace view provides a common unit for auditing reward
design, credit and signal assignment, and orchestration learning.

Using this lens, we identify three technical axes. First, reward
design falls into eight families; orchestration rewards target
system-level properties such as parallelism speedup, split
correctness, and aggregation quality. Second, reward and credit
signals attach to eight credit- or signal-bearing units from token to
team; explicit counterfactual message-level credit remains especially
sparse in our curated pool, while agent-, role-, turn-, and
orchestrator-level signals are beginning to fill in. Third,
orchestration learning decomposes into five sub-decisions (when to
spawn, whom to delegate to, how to communicate, how to aggregate, when
to stop); within our curated pool as of May 4, 2026, we found no
explicit RL training method for the stopping decision.

We connect academic methods to public industrial evidence from Kimi
Agent Swarm, OpenAI Codex, and Anthropic Claude~Code. The resulting
scale gap should be read as a gap between publicly reported deployment
envelopes and open academic evaluation regimes, not as independent
verification of industrial training traces: Kimi is the clearest
public trained-orchestrator anchor, while Codex and Claude~Code mainly
document deployment shape and harness constraints. We release the
artifact at \artifactrepo, including an $84$-entry tagged paper pool, a
$32$-record exclusion log, scripted corpus statistics, and a minimal
JSON schema for replayable orchestration traces, then close with
fifteen research directions spanning algorithms, rewards, systems,
safety, and evaluation.
\end{abstract}

\clearpage
\tableofcontents
\clearpage

\section{Introduction}
\label{sec:intro}

\thesisbox{Thesis}{
Single-agent reinforcement learning (RL) for large language models (LLMs)
optimizes \emph{trajectories}: a sequence of tokens, tool calls, and
environment observations produced by one policy. As LLM agents evolve from
isolated tool users into coordinated teams, we use
\emph{orchestration traces} as a working abstraction for taxonomy and
audit in addition to per-agent trajectories: a temporal interaction
graph in which an orchestrator decides \emph{when} to spawn sub-agents,
\emph{whom} to delegate to, \emph{how} they should communicate, \emph{which}
tools they may call, and \emph{how} their partial outputs are aggregated.
This re-frames the central technical challenges as (i)~reward design across
team, individual, process, tool, and verifier signals; (ii)~credit
assignment across agents, turns, messages, tool calls, and orchestrator
decisions; and (iii)~learning the orchestration process itself.
}

\begin{figure}[t]
\centering
\resizebox{0.98\linewidth}{!}{%
\begin{tikzpicture}[
  font=\footnotesize,
  box/.style={draw=black!35, rounded corners=3pt, thick, align=center, minimum height=0.9cm},
  source/.style={box, fill=gray!12, text width=2.9cm},
  core/.style={box, fill=green!12, draw=green!45!black, text width=3.6cm, minimum height=1.4cm},
  reward/.style={box, fill=blue!10, draw=blue!55!black, text width=2.6cm, minimum height=1.25cm},
  credit/.style={box, fill=orange!13, draw=orange!70!black, text width=2.6cm, minimum height=1.25cm},
  orch/.style={box, fill=green!12, draw=green!50!black, text width=2.6cm, minimum height=1.25cm},
  support/.style={box, fill=gray!10, draw=black!30, text width=2.85cm, minimum height=0.95cm},
  risk/.style={box, fill=red!8, draw=red!50!black, text width=2.85cm, minimum height=0.95cm},
  arrow/.style={-Latex, thick, black!55}
]

  \node[source] (single)   at (0,  2.7) {single-agent\\LLM RL};
  \node[source] (marl)     at (0,  0.9) {classical\\MARL};
  \node[source] (industry) at (0, -0.9) {industrial\\agent systems};

  \node[core] (trace) at (4.3, 0.9)
    {\textbf{Orchestration Trace}\\[1pt]
     temporal event graph\\
     spawn / message / tool / aggregate};

  \node[reward] (rew)    at (8.2,  2.7) {\textbf{Reward}\\8 families\\R1--R8};
  \node[credit] (cred)   at (8.2,  0.9) {\textbf{Credit}\\8 units\\team--token};
  \node[orch]   (orches) at (8.2, -0.9) {\textbf{Orchestration}\\5 decisions\\O1--O5};

  \node[support] (bench) at (12.0,  2.7) {benchmarks\\E1--E4};
  \node[risk]    (safe)  at (12.0,  0.9) {safety / risks\\attack surfaces};
  \node[support] (open)  at (12.0, -0.9) {open problems\\P1--P15};

  \draw[arrow] (single.east) -- (trace.west);
  \draw[arrow] (marl.east) -- (trace.west);
  \draw[arrow] (industry.east) -- (trace.west);

  \draw[arrow] (trace.east) -- (rew.west);
  \draw[arrow] (trace.east) -- (cred.west);
  \draw[arrow] (trace.east) -- (orches.west);

  \draw[arrow] (rew.east) -- (bench.west);
  \draw[arrow] (cred.east) -- (safe.west);
  \draw[arrow] (orches.east) -- (open.west);
  \draw[arrow, black!35] (cred.east) -- (bench.west);
  \draw[arrow, black!35] (rew.east) -- (open.west);

  \node[anchor=west, font=\scriptsize, black!65] at (0, -2.25)
    {Color key: reward = blue, credit = orange, orchestration = green, safety/risk = red, systems/evidence = gray.};

\end{tikzpicture}%
}
\caption{Paper map. Reading: the survey takes three input traditions
(single-agent LLM RL, classical MARL, and industrial agent systems),
foregrounds the orchestration trace as the shared object, and then
organizes the literature into reward design, credit assignment, and
orchestration learning. Benchmarks, safety, and open problems are
downstream because they inherit the same trace structure.}
\label{fig:paper-map}
\end{figure}

\subsection{Why now: recent developments}
\label{sec:intro:why-now}

Three concurrent signals make May 4, 2026 a useful cutoff for this
paper.

\textbf{Public industrial evidence exposes larger deployment envelopes.}
Moonshot's Kimi~K2.5 introduced an Agent~Swarm trained with
Parallel-Agent Reinforcement Learning (PARL), scaling to up to
$100$ sub-agents and $1{,}500$ coordinated steps / tool calls as
reported~\cite{kimi-k2-5-2026}; K2.6 expanded this to $300$
sub-agents and $4{,}000$ coordinated steps, adding a ``Claw Groups''
research preview of
cross-vendor coordination~\cite{kimi-k2-6-2026}. We treat these
numbers as a publicly reported deployment envelope rather than an
independently reproduced training trace. Kimi PARL is the clearest
public example in our pool of \emph{trained} multi-agent
orchestration. OpenAI's Codex app is described in official materials
as a command center managing parallel software-engineering
agents~\cite{openai-codex2025}, and Anthropic's Claude~Code ships
built-in and user-defined sub-agents~\cite{claude-code-subagents-2025},
with an engineering post-mortem of sixteen parallel Claudes jointly
building a C compiler~\cite{anthropic-c-compiler2026}; in both cases
the public material documents the deployment form---parallel
workflows, harness boundaries, dynamic spawn---without disclosing
whether multi-agent coordination itself is an RL training target. We
treat Kimi as the published-training anchor and Codex / Claude~Code as
deployment-shape and engineering-pressure evidence (\S\ref{sec:systems}).

\textbf{Academic methods are catching up with the right primitives.}
In the window from 2025-Q2 through May 2026, the literature in our
pool produced a systematic multi-agent RFT
paradigm~\cite{marft2025,maporl2025,magrpo2025}, a
hierarchical GRPO decomposition for LLM teams~\cite{m-grpo2025},
a single-LLM dual-role policy optimization with tool
integration~\cite{matpo2025}, a stability analysis of
multi-agent GRPO~\cite{dr-mas2026}, and credit-assignment
methods targeting message-level
counterfactuals~\cite{c3-2026} and Shapley-based agent-level
credit~\cite{sharp2026}. A May 2026 coverage refresh added
closely related OpenReview, arXiv, and project-page entries on
meta-thinking and deliberation~\cite{rema2025,learning-to-deliberate2025},
UI-agent credit re-assignment~\cite{collabuiagents2025},
interaction-derived rewards and self-evolution~\cite{comas2026,sirius2025,multiagent-finetuning2025},
planner/workforce optimization~\cite{owl2025}, and zero-supervision
MAS design~\cite{mas-zero2025}. A May 2026 refresh added
actor-critic decentralized collaboration~\cite{collm-maac2026},
width-scaling search teams~\cite{wideseek-r1-2026},
communication/topology learning~\cite{agent-qmix2026}, language-space
credit assignment~\cite{langmarl2026}, multi-agent self-search for
code~\cite{marti-mars2-2026}, GUI role orchestration~\cite{lamo2026},
attacker--defender safety training~\cite{magic2026}, and
self-play / hierarchical interaction entries from OpenReview
submissions and proceedings~\cite{spiral2026,marshal2026,depart2026,sage2026}.
These are not isolated tricks---they
collectively formalize LLM collaboration as cooperative MARL with
new credit- and signal-bearing units. Figure~\ref{fig:method-timeline} visualizes
the corpus across an 18-month window.

\begin{figure}[t]
\centering
\resizebox{\linewidth}{!}{%
\begin{tikzpicture}[font=\footnotesize,
  agent/.style={circle, draw=blue!65, thick, fill=blue!12,
                inner sep=1pt, minimum size=5.2mm, font=\scriptsize},
  orch/.style={circle, draw=red!70, thick, fill=red!12,
               inner sep=1pt, minimum size=5.2mm, font=\scriptsize},
  role/.style={circle, draw=orange!80, thick, fill=orange!15,
               inner sep=1pt, minimum size=5.2mm, font=\scriptsize},
  turn/.style={circle, draw=green!50!black, thick, fill=green!12,
               inner sep=1pt, minimum size=5.2mm, font=\scriptsize},
  msg/.style={circle, draw=purple!70, thick, fill=purple!12,
              inner sep=1pt, minimum size=5.2mm, font=\scriptsize},
  framew/.style={circle, draw=gray!70, thick, fill=gray!18,
                 inner sep=1pt, minimum size=5.2mm, font=\scriptsize},
  tick/.style={black!50, thick},
  qlabel/.style={font=\scriptsize\itshape, black!60, anchor=north},
  pbelow/.style={anchor=north, font=\tiny, black!70,
                 fill=white, fill opacity=0.88, text opacity=1,
                 inner sep=0.6pt},
  pabove/.style={anchor=south, font=\tiny, black!70,
                 fill=white, fill opacity=0.88, text opacity=1,
                 inner sep=0.6pt}
]

  \draw[->, thick, black!65] (0, 0) -- (10.8, 0)
    node[right]{\small time};

  \foreach \x/\lab in {0.5/Q4'24, 2.0/Q1'25, 3.5/Q2'25, 5.0/Q3'25,
                      6.5/Q4'25, 8.0/Q1'26, 9.5/Q2'26} {
    \draw[tick] (\x, -0.12) -- (\x, 0.12);
    \node[qlabel] at (\x, -0.15) {\lab};
  }


  \node[framew] at (-0.9, 3.2) {F};
  \node[anchor=east, black!65, font=\scriptsize] at (-1.18, 3.2)
    {framework / system};
  \node[orch] at (-0.9, 2.6) {O};
  \node[anchor=east, black!65, font=\scriptsize] at (-1.18, 2.6)
    {orchestrator credit};
  \node[role] at (-0.9, 2.0) {R};
  \node[anchor=east, black!65, font=\scriptsize] at (-1.18, 2.0)
    {role credit};
  \node[agent] at (-0.9, 1.4) {A};
  \node[anchor=east, black!65, font=\scriptsize] at (-1.18, 1.4)
    {agent credit};
  \node[turn] at (-0.9, 0.8) {T};
  \node[anchor=east, black!65, font=\scriptsize] at (-1.18, 0.8)
    {turn credit};
  \node[msg] at (-0.9, 0.2) {M};
  \node[anchor=east, black!65, font=\scriptsize] at (-1.18, 0.2)
    {message credit};

  \node[role, anchor=center] (malt) at (1.0, 2.0) {};
  \node[pbelow] at (1.0, 1.64) {MALT};

  \node[agent, anchor=center] (maporl) at (2.3, 1.4) {};
  \node[pbelow] at (2.3, 1.04) {MAPoRL};

  \node[orch, anchor=center] (pup) at (3.7, 2.6) {};
  \node[pbelow] at (3.7, 2.24) {Puppeteer};
  \node[role, anchor=center] (halo) at (4.3, 2.0) {};
  \node[pbelow] at (4.3, 1.64) {HALO};
  \node[agent, anchor=center] (marft) at (4.3, 1.4) {};
  \node[pbelow] at (4.3, 1.04) {MARFT};

  \node[agent, anchor=center] (magrpo) at (5.0, 1.4) {};
  \node[pbelow] at (5.0, 1.04) {MAGRPO};
  \node[framew, anchor=center] (agl) at (5.6, 3.2) {};
  \node[pbelow] at (5.6, 2.84) {Agent Lt.};

  \node[role, anchor=center] (matpo) at (6.3, 2.0) {};
  \node[pbelow] at (6.3, 1.64) {MATPO};
  \node[turn, anchor=center] (cf) at (6.3, 0.8) {};
  \node[pbelow] at (6.3, 0.44) {Ctx-Fold};
  \node[role, anchor=center] (mg) at (7.2, 2.0) {};
  \node[pbelow] at (7.2, 1.64) {M-GRPO};
  \node[turn, anchor=center] (marsrl) at (7.2, 0.8) {};
  \node[pbelow] at (7.2, 0.44) {MarsRL};

  \node[framew, anchor=center] (kimi25) at (8.0, 3.2) {};
  \node[pbelow] at (8.0, 2.84) {Kimi K2.5};
  \node[agent, anchor=center] (drmas) at (8.4, 1.4) {};
  \node[pbelow] at (8.4, 1.04) {Dr.MAS};
  \node[agent, anchor=center] (sharp) at (8.4, 1.55) {};

  \node[msg, anchor=center] (c3) at (9.1, 0.2) {};
  \node[pbelow] at (9.1, -0.16) {C3};
  \node[orch, anchor=center] (para) at (9.6, 2.6) {};
  \node[pbelow] at (9.55, 2.24) {ParaMgr};
  \node[orch, anchor=center] (hera) at (9.95, 2.82) {};
  \node[pabove, anchor=south east] at (9.85, 3.14) {HERA};
  \node[framew, anchor=center] (kimi26) at (10.35, 3.2) {};
  \node[pbelow] at (10.35, 2.84) {K2.6};

  \draw[decorate, decoration={brace, amplitude=3pt, mirror}, thick, red!60]
    (4.5, 3.65) -- (10.45, 3.65)
    node[midway, above=4pt, red!70, font=\scriptsize\bfseries]
    {recent method cluster in an 18-month window};

\end{tikzpicture}%
}
\caption{Timeline of selected representative LLM-MAS entries from Q4~2024 to
Q2~2026, plotted by arXiv submission date and grouped vertically
by the credit-bearing unit they target
(\S\ref{sec:credit:hierarchy}). Nearly the entire corpus sits in an
18-month window, motivating the timing claim in
\S\ref{sec:intro:why-now}. The orchestrator and message rows
remain sparsely populated throughout; agent- and role-level credit
has received the most attention.}
\label{fig:method-timeline}
\end{figure}
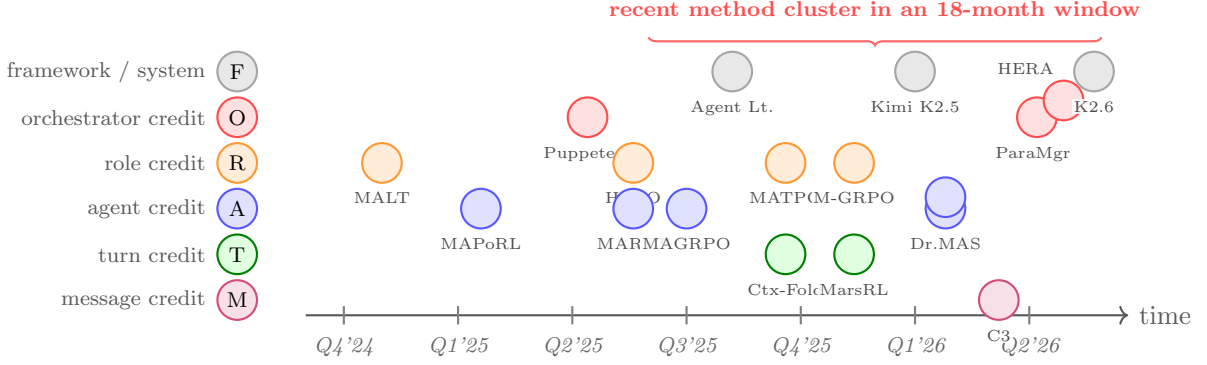

\begin{figure}[t]
\centering
\small
\renewcommand{\arraystretch}{1.08}
\begin{tabular}{@{}lccccc@{}}
\toprule
\textbf{Entry} & \textbf{Reward} & \textbf{Credit} & \textbf{Orch.} & \textbf{Eval.} & \textbf{Safety} \\
\midrule
MAGRPO / MARFT / MAPoRL & \(\bullet\) & \(\bullet\) & \(\circ\) & \(\circ\) & -- \\
M-GRPO / MATPO / MALT & \(\bullet\) & \(\bullet\) & \(\bullet\) & \(\circ\) & -- \\
Dr.\ MAS / C3 / SHARP & \(\bullet\) & \(\bullet\) & \(\circ\) & \(\circ\) & -- \\
Puppeteer / HALO / ParaManager & \(\bullet\) & \(\circ\) & \(\bullet\) & \(\circ\) & -- \\
Agent Lightning / MarsRL & \(\bullet\) & \(\bullet\) & \(\circ\) & \(\bullet\) & -- \\
ReMA / Learning to Deliberate & \(\bullet\) & \(\circ\) & \(\bullet\) & \(\circ\) & -- \\
CollabUIAgents / OWL & \(\bullet\) & \(\bullet\) & \(\bullet\) & \(\bullet\) & -- \\
CoMAS / SiriuS / Multiagent FT & \(\bullet\) & \(\circ\) & \(\bullet\) & \(\circ\) & -- \\
MAE / MAS-Zero & \(\bullet\) & \(\circ\) & \(\bullet\) & \(\circ\) & -- \\
WideSeek / Agent Q-Mix / LangMARL & \(\bullet\) & \(\bullet\) & \(\bullet\) & \(\circ\) & -- \\
MAGIC / SPIRAL / MARSHAL / DEPART & \(\bullet\) & \(\bullet\) & \(\bullet\) & \(\circ\) & \(\circ\) \\
Kimi / Codex / Claude Code & \(\circ\) & \(\circ\) & \(\bullet\) & \(\circ\) & \(\circ\) \\
TAMAS / AgentDojo / WASP & -- & \(\circ\) & \(\circ\) & \(\bullet\) & \(\bullet\) \\
SWE-Bench / WebArena / MultiAgentBench & -- & -- & \(\circ\) & \(\bullet\) & \(\circ\) \\
\bottomrule
\end{tabular}
\caption{Compact coverage map for representative retained entries.
\(\bullet\) means the entry directly studies the dimension; \(\circ\)
means it supplies indirect evidence, a benchmark substrate, or a system
constraint. The sparsity is intentional: it shows why the survey treats
reward, credit, orchestration, evaluation, and safety as coupled but
unevenly supported dimensions.}
\label{fig:taxonomy-heatmap}
\end{figure}

\begin{table}[t]
\centering
\small
\renewcommand{\arraystretch}{1.12}
\begin{tabularx}{\linewidth}{@{}p{2.5cm}p{2.8cm}X@{}}
\toprule
\textbf{Facet} & \textbf{Dominant counts} & \textbf{Interpretation} \\
\midrule
Source category
 & 42 RL methods; 18 benchmarks; 10 classical-MARL foundations; 6 industrial cases
 & The pool is method-centered but includes benchmarks, systems, and classical primitives only when they are load-bearing for the taxonomy. \\
RL relevance
 & 49 yes; 9 partial; 26 no
 & Non-RL entries are retained only as contrastive foundations, benchmarks, safety cases, or industrial deployment anchors. \\
Reward family
 & 15 hybrid; 10 shared; 7 orchestration; 6 verifier; 33 NA
 & Explicit reward design is concentrated in the LLM-MAS RL and critic/process-supervision entries. \\
Signal / credit granularity
 & 23 agent; 10 role; 8 orchestrator; 5 turn; 2 message; 36 NA
 & Message-level signals are rare, and explicit counterfactual message-level credit is rarer still; orchestrator-level signals are growing but still narrow relative to agent- and role-level signals. \\
Orchestration form
 & 18 centralized; 13 hierarchical; 8 debate; 4 swarm; 3 harness; 34 NA
 & The retained methods emphasize centralized and hierarchical controllers; many supporting entries do not define an orchestration policy. \\
\bottomrule
\end{tabularx}
\caption{Coverage statistics computed from the $84$ retained entries
in the artifact repository. These counts support the qualitative
claims about sparsity in message-level explicit credit, orchestrator-level signals, and
MAS-native evaluation; they should not be read as field-wide
prevalence estimates.}
\label{tab:coverage-statistics}
\end{table}

\begin{table}[t]
\centering
\small
\renewcommand{\arraystretch}{1.08}
\begin{tabular}{@{}lrrrrrr@{}}
\toprule
\textbf{Reward family} & \textbf{NA} & \textbf{agent} & \textbf{msg.} & \textbf{orch.} & \textbf{role} & \textbf{turn} \\
\midrule
shared        & 0  & 9 & 1 & 0 & 0 & 0 \\
hybrid        & 0  & 6 & 0 & 1 & 4 & 4 \\
orchestration & 0  & 0 & 0 & 6 & 1 & 0 \\
individual    & 0  & 4 & 0 & 0 & 0 & 0 \\
debate        & 0  & 3 & 1 & 0 & 0 & 0 \\
process       & 0  & 1 & 0 & 0 & 0 & 1 \\
role          & 0  & 0 & 0 & 0 & 3 & 0 \\
verifier      & 4  & 0 & 0 & 0 & 2 & 0 \\
NA            & 32 & 0 & 0 & 1 & 0 & 0 \\
\bottomrule
\end{tabular}
\caption{Reward-family by signal/credit-granularity cross-tab generated from
the retained-entry CSV in the artifact repository. The table makes the sparsity claim more
concrete within the retained pool: the finest retained tag is \texttt{message} for only two
entries, and only one of those (C3) explicitly estimates
message-level counterfactual credit; orchestrator-level tags now include
width-scaling and orchestration-reward entries, but remain far fewer
than agent-level tags. The counts are reproducible with the statistics script in the
accompanying artifact repository.}
\label{tab:reward-credit-crosstab}
\end{table}

\textbf{Existing surveys cover pairwise intersections but not the
triple.} Surveys of LLM-based multi-agent systems~\cite{survey-mas2024}
and their collaboration mechanisms~\cite{survey-collab2025} cover
architectures and applications; the recent
$500+$-paper agentic RL survey~\cite{survey-agentic-rl2025} and the
LLM-lifecycle RL survey~\cite{survey-rl-meets-llm2025} cover
single-agent agentic RL extensively; the agentic reasoning
survey~\cite{survey-agentic-reasoning2026} covers reasoning agents
broadly. None triangulates the three: \emph{multi-agent} and
\emph{RL/post-training} and \emph{LLM agents}. That is the gap this
paper targets.

\subsection{Scope and positioning}
\label{sec:intro:position}

The contribution is a taxonomy paper with an explicit position:
LLM-MAS RL is most usefully organized around the orchestration trace.
The paper therefore does not try to be a neutral catalogue of all
multi-agent LLM systems. It asks a narrower question: when LLM agents
are trained or post-trained as teams, which parts of the interaction
graph can be rewarded, credited, and learned? The benchmark
requirements in \S\ref{sec:bench:good} are consequently framed as
reporting recommendations derived from gaps in the retained pool, not
as a new benchmark release.

\textbf{Relative to LLM-MAS architecture surveys.}
\cite{survey-mas2024,survey-collab2025} catalogue agent profiles,
perception, action, and interaction mechanisms, but say comparatively
little about how these components are \emph{trained}. Our focus is
the post-training stage only: given an architecture, what rewards
drive it, how is credit assigned, and how is the orchestration
process itself learned?

\textbf{Relative to agentic RL surveys.} \cite{survey-agentic-rl2025}
covers the full single-agent agentic-RL landscape; we cover the step
that follows: what changes when the policy is no longer a single
agent but an orchestrated team of them. Many primitives carry over
(PPO, GRPO, verifiable rewards) and we do not re-derive them here.

\textbf{Relative to classical MARL.} We treat classical MARL as a
\emph{conceptual toolkit} (Dec-POMDP, CTDE, COMA, Shapley credit,
VDN/QMIX, MAPPO/IPPO) rather than as a field summary. \S\ref{sec:background}
covers only the MARL concepts that are load-bearing for later sections;
we refer the reader elsewhere for comprehensive MARL treatment.

The scope is deliberately bounded. The retained pool is not an
exhaustive list of every multi-agent LLM paper, a benchmark
leaderboard, or a system-design manual. We curate $51$ \emph{focal}
LLM-MAS entries---RL methods, industrial cases, and directly adjacent
surveys---that populate the three taxonomies that structure the paper,
supplemented by $33$ classical-MARL, safety, single-agent-RL,
benchmark, and critic / tool-use evaluation references
($84$ retained entries total). The accompanying artifact snapshot and
repository (\artifactrepo) contain the retained-entry CSV, the
$32$-record exclusion log, the statistics script, and the trace-schema
files. Together these expose $116$ audited records; Appendix~\ref{app:artifact-protocol}
gives the search strings, screening stages, borderline examples, and
tag definitions.

\subsection{Corpus construction and evidence levels}
\label{sec:intro:methodology}

The paper pool is deliberately \emph{curated}, not exhaustive. We
constructed it in four passes. First, we seeded the pool from adjacent
surveys on LLM-based multi-agent systems, agentic RL, and RL for
LLMs~\cite{survey-mas2024,survey-collab2025,survey-agentic-rl2025,survey-rl-meets-llm2025}.
Second, we searched arXiv, ACL Anthology, OpenReview, and official
project pages for combinations of
\texttt{multi-agent LLM}, \texttt{reinforcement learning},
\texttt{post-training}, \texttt{credit assignment},
\texttt{orchestration}, \texttt{agent swarm}, \texttt{tool use}, and
\texttt{prompt injection}. Third, we added backward and forward
citation links when they supplied a load-bearing concept for one of
our three taxonomies. Fourth, we audited the resulting set against
three inclusion rules: the work must either (i) train or post-train an
LLM-MAS component, (ii) document an industrial system whose public
interface constrains RL design, or (iii) provide a benchmark, safety
case, or classical-MARL primitive used later in the paper.

We exclude papers that use multiple LLM calls only as an implementation
detail but do not expose multi-agent interaction, reward design, credit
assignment, or orchestration as a study object. Tags in
the retained-entry CSV were assigned by manual reading of the
abstract, method section, and public artifact when available. The CSV
records an explicit \texttt{verified} field and source status through
its category, venue, and notes fields; we do
not claim formal inter-annotator agreement. Instead, we treat the
18-column schema as a structured taxonomy artifact whose entries can be
corrected as the literature changes.

Because several load-bearing systems are industrial, we separate
evidence levels throughout the paper. Peer-reviewed and arXiv methods
are used for algorithmic claims; company technical reports are used
only where they disclose training or evaluation details; product
documentation and blogs are used as deployment-shape evidence unless
they explicitly disclose a training mechanism. This distinction is
made explicit in \S\ref{sec:systems:industry}.

\begin{table}[htbp]
\centering
\small
\renewcommand{\arraystretch}{1.15}
\begin{tabularx}{\linewidth}{@{}p{3.0cm} X@{}}
\toprule
\textbf{Protocol item} & \textbf{This paper} \\
\midrule
Cutoff
 & Literature and public-system audit through May 4, 2026. \\
Search sources
 & arXiv, ACL Anthology, OpenReview, Semantic Scholar / citation links,
   and official company documentation or technical blogs for deployed
   systems. \\
Query families
 & \texttt{multi-agent LLM} $\times$
   \{\texttt{reinforcement learning}, \texttt{post-training},
   \texttt{credit assignment}, \texttt{orchestration},
   \texttt{agent swarm}, \texttt{tool use}, \texttt{prompt injection}\}. \\
Screening outcome
 & Internal audit count after the May 2026 coverage refresh:
   $116$ candidate records were considered; $84$ were retained in the
   tagged pool and $32$ exclusion decisions were logged.
   The retained-entry CSV is the reusable taxonomy
   artifact; the exclusion log is a screening-decision log,
   not a retained-paper bibliography or a full systematic-review
   reproducibility package. \\
Inclusion rule
 & Include work that trains or post-trains an LLM-MAS component,
   documents a deployment interface that constrains RL design, or
   supplies a benchmark, safety case, or MARL primitive used later. \\
Tagging rubric
 & Each retained entry is tagged for source category, RL use, reward
   family, finest credit granularity, orchestration form, scenario,
   core/case/supporting status, verification status, and notes. \\
\bottomrule
\end{tabularx}
\caption{Curation protocol for the paper pool. The counts are internal
audit counts for this paper rather than a claim of exhaustive
coverage or independently reproducible screening.}
\label{tab:survey-protocol}
\end{table}

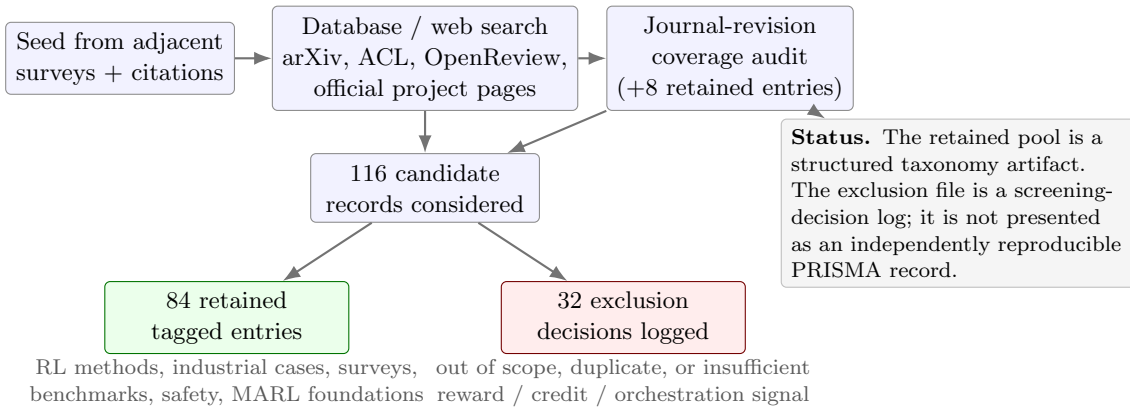
\begin{figure}[t]
\centering
\resizebox{0.94\linewidth}{!}{%
\begin{tikzpicture}[
  font=\footnotesize,
  box/.style={draw=black!45, rounded corners=2pt, fill=blue!5,
              minimum width=3.0cm, minimum height=0.9cm, align=center},
  keep/.style={draw=green!45!black, rounded corners=2pt, fill=green!8,
               minimum width=3.2cm, minimum height=0.9cm, align=center},
  drop/.style={draw=red!45!black, rounded corners=2pt, fill=red!7,
               minimum width=3.2cm, minimum height=0.9cm, align=center},
  note/.style={draw=black!25, rounded corners=2pt, fill=gray!8,
               text width=4.4cm, align=left, font=\scriptsize},
  arr/.style={-Latex, thick, black!55}
]

\node[box] (seed) at (0, 0) {Seed from adjacent\\surveys + citations};
\node[box] (search) at (4.0, 0) {Database / web search\\arXiv, ACL, OpenReview,\\official project pages};
\node[box] (audit) at (8.0, 0) {Journal-revision\\coverage audit\\(+8 retained entries)};

\node[box] (cand) at (4.0, -1.7) {116 candidate\\records considered};
\node[keep] (keep) at (1.4, -3.4) {84 retained\\tagged entries};
\node[drop] (drop) at (6.6, -3.4) {32 exclusion\\decisions logged};

\node[note] (scope) at (11.0, -1.9) {
\textbf{Status.} The retained pool is a structured taxonomy artifact.
The exclusion file is a screening-decision log; it is not presented as
an independently reproducible PRISMA record.};

\draw[arr] (seed) -- (search);
\draw[arr] (search) -- (audit);
\draw[arr] (search) -- (cand);
\draw[arr] (audit) -- (cand);
\draw[arr] (cand) -- (keep);
\draw[arr] (cand) -- (drop);
\draw[arr] (audit) -- (scope);

\node[font=\scriptsize, black!65, align=center] at (1.4, -4.25)
  {RL methods, industrial cases, surveys,\\benchmarks, safety, MARL foundations};
\node[font=\scriptsize, black!65, align=center] at (6.6, -4.25)
  {out of scope, duplicate, or insufficient\\reward / credit / orchestration signal};

\end{tikzpicture}%
}
\caption{Corpus construction flow. Counts are internal audit counts
after the journal-revision coverage audit, not a claim of exhaustive
coverage or independently reproducible screening.}
\label{fig:corpus-flow}
\end{figure}

\subsection{Contributions}
\label{sec:intro:contributions}

\begin{itemize}
  \item \textbf{A unifying thesis.} We argue that LLM-MAS RL is
    usefully analyzed through the \emph{orchestration trace},
    understood as an event graph, rather than only through per-agent
    trajectories; this reframing reorganizes a large fraction of the
    recent literature.
  \item \textbf{A lightweight taxonomy formalism.} We extend the
    Dec-POMDP to a dynamic-Dec-POMDP that accommodates spawn
    and despawn actions (\S\ref{sec:formalism}) and state two
    informal observations: credit diffusion under uniform credit,
    and non-iden\-tifi\-ability of orchestrator spawn decisions.
    These organize the rest of the paper. The formalism is intended
    as an \emph{organizing abstraction} for taxonomy and auditability,
    not a new MARL theory; concrete algorithmic forms and tight rates are
    open (\S\ref{sec:open}).
  \item \textbf{Three taxonomies.} We organize methods along
    (a)~reward design across eight families (\S\ref{sec:rewards}),
    (b)~credit and signal assignment across eight credit- or
    signal-bearing units
    (\S\ref{sec:credit}), and
    (c)~orchestration learning across five sub-decisions
    (\S\ref{sec:orchestration}).
  \item \textbf{An industrial--academic bridge.} We connect open
    methods to Kimi PARL, OpenAI Codex, and Anthropic Claude Code
    (\S\ref{sec:systems}), identify which design choices in these
    systems have---and have not---been published, and characterize the
    gap between publicly reported industrial deployment envelopes and
    open academic evaluation regimes in rollout cost and trace length
    (\S\ref{sec:engineering}).
  \item \textbf{An open, tagged paper pool.} We release an $84$-entry
    curated pool ($51$ focal LLM-MAS entries plus $33$ supporting
    references) with $18$-column taxonomy tags, synchronised with
    the paper bibliography and summarised as a single table in
    Appendix~\ref{app:summary-table}. The broader artifact contains
    $116$ audited records when the exclusion log is included
    (Appendix~\ref{app:artifact-protocol}). It is intended as a
    reusable taxonomy substrate that follow-up work can extend without
    re-curating from scratch.
  \item \textbf{Scripted corpus statistics and trace schema.}
    The artifact includes a statistics script, a static statistics
    snapshot, a machine-readable orchestration-trace JSON Schema, a
    valid example trace, and a dependency-free trace validator
    (\S\ref{sec:artifact-statement}). These make the sparsity claims
    and benchmark-reporting recommendations mechanically inspectable.
  \item \textbf{Entry cards.} Appendix~\ref{app:method-cards}
    gives one-card summaries for thirteen core methods, frameworks,
    and industrial anchors under a uniform template,
    suitable as a quick reference complementing the main taxonomies.
  \item \textbf{Open problems.} We identify fifteen open problems
    (\S\ref{sec:open}), organized along algorithmic, reward,
    systems, safety, and evaluation axes.
\end{itemize}

\subsection{Roadmap}
\label{sec:intro:roadmap}

\S\ref{sec:intro:methodology} defines the corpus and evidence levels.
\S\ref{sec:background} gives the minimal MARL and agentic-RL
background; \S\ref{sec:formalism} extends the Dec-POMDP to the
dynamic-agent setting needed for the rest of the paper.
\S\ref{sec:systems} covers industrial and academic system forms;
\S\ref{sec:engineering} quantifies the engineering constraints
(rollout cost, harness boundary, trace-length dependence) that
discipline algorithm choice.
\S\ref{sec:rewards}--\S\ref{sec:orchestration} are the three
pillars of the thesis: reward design, credit assignment, and
orchestration learning.
\S\ref{sec:benchmarks} argues that current benchmarks fail to
measure the very properties (parallelism efficiency, collaboration
quality, error amplification) that LLM-MAS RL is supposed to
optimize.
\S\ref{sec:open} lists fifteen open problems and
\S\ref{sec:conclusion} returns to the thesis.
Appendices~\ref{app:method-cards}--\ref{app:summary-table} contain
the method cards and the complete paper-pool summary table.

\section{Background: From MARL to LLM-MARL}
\label{sec:background}

This section gives the minimal background needed for the rest of the
survey. We cover classical MARL (\S\ref{sec:background:marl}) and
single-agent LLM RL (\S\ref{sec:background:llmrl}) compactly, then
spend the rest of this section on what makes LLM-MAS genuinely
different from either (\S\ref{sec:background:whats-different}).

\subsection{Classical MARL in one page}
\label{sec:background:marl}

A \emph{Markov game}~\cite{markov-games1994} generalizes an MDP to $n$ agents: each agent
$i$ has an action space $\mathcal{A}_i$, observation space
$\mathcal{O}_i$, and policy $\pi_i$; transitions are driven by the
joint action $(a_1, \ldots, a_n)$ and yield per-agent rewards
$r_i$ (cooperative, competitive, or mixed-motive). When observations
are partial, the setting is a
\emph{decentralized partially-observable MDP} (Dec-POMDP)~\cite{dec-pomdp2002}.

Two design choices organize most classical MARL algorithms:

\begin{itemize}
  \item \textbf{Centralized training, decentralized execution (CTDE).}
    A central critic that sees the joint $(s, a_1, \ldots, a_n)$ is
    used only during training; at deployment each agent runs on its
    own observation $o_i$.
    VDN~\cite{vdn2018}, QMIX~\cite{qmix2018},
    MADDPG~\cite{maddpg2017}, and MAPPO~\cite{mappo2022}
    all live in this family.
  \item \textbf{Value decomposition vs.\ counterfactual baselines.}
    VDN/QMIX decompose a team value function into per-agent
    contributions additively or monotonically.
    COMA~\cite{coma2018} replaces that
    with a counterfactual baseline: agent $i$'s advantage is the
    difference between the team return and the return under a
    counterfactual where $i$'s action is marginalized.
    Shapley-value credit~\cite{shapley-q2020}
    generalizes this to a fair marginal-contribution attribution
    over all subsets of agents; difference rewards~\cite{difference-rewards2001}
    are the closely-related earlier formulation.
\end{itemize}

Two practical algorithms recur in LLM-MAS papers: \textbf{IPPO}~\cite{ippo2020}
(independent PPO per agent, no centralized critic) and \textbf{MAPPO}~\cite{mappo2022}
(shared policy with centralized critic). Dr.\,MAS~\cite{dr-mas2026}
is the most visible recent paper to reopen the
IPPO-vs-MAPPO-vs-GRPO question in the LLM-MAS setting; its central
observation is that GRPO's group-normalized advantage, borrowed
unchanged from single-agent reasoning RL, becomes unstable at the
agent level without explicit agent-wise normalization.

\subsection{Single-agent LLM RL in one page}
\label{sec:background:llmrl}

Single-agent LLM RL has evolved rapidly:
\textbf{RLHF}~\cite{instructgpt2022} (preference rewards from human labels)
$\to$
\textbf{RLAIF} (preference rewards from AI judges)
$\to$
\textbf{RLVR} (verifiable rewards against ground truth)
$\to$
\textbf{Reasoning RL}~\cite{deepseek-r12025} (o1-/R1-style long-CoT with GRPO)
$\to$
\textbf{Agentic RL}~\cite{react2023} (multi-turn tool use, web browsing, code
execution).

Two axes organize this progression. Along the \emph{reward} axis, the
signal shifts from sparse preference (one label per rollout) to
dense verifiable (per-step check) to hybrid. Along the \emph{credit}
axis, the unit shifts from trajectory-level PPO to token-level GAE to
step- or turn-level process rewards (PRM). By the time one reaches
agentic RL, the policy already produces actions at three natural
granularities---token, action, tool call---and credit must be
assigned across all three. The multi-agent extension adds further
granularities \emph{above} the single-agent trajectory (agent, role,
orchestrator), which is the subject of \S\ref{sec:credit}.

Two representative methods are load-bearing below. \textbf{PPO}~\cite{ppo2017} and
\textbf{GRPO}~\cite{deepseekmath2024} are the dominant policy-optimization choices: PPO uses
a learned value baseline, GRPO normalizes advantages within a group
of $K$ rollouts from the same prompt, eliminating the value network.
GRPO's simplicity makes it the default in most multi-agent papers in
our pool, but as Dr.\,MAS~\cite{dr-mas2026} documents, its
group-normalization is what needs to change at the multi-agent level.

\subsection{Why LLM-MAS is not classical MARL}
\label{sec:background:whats-different}

Seven differences separate LLM-MAS from the classical MARL setting
in \S\ref{sec:background:marl}. Each has direct consequences for
algorithm design in later sections.

\begin{enumerate}
  \item \textbf{Action space is natural language.}
    A sub-agent's action is a generated message, a tool invocation,
    or a sub-agent spawn. This makes the action space combinatorial
    and ill-defined for classical MARL machinery (VDN's additive
    decomposition, MADDPG's continuous-control assumptions).
  \item \textbf{Observation is long and partially summarized.}
    An agent may see a conversation transcript of thousands of
    tokens, a tool-returned document, or a summarized report from
    another agent. Observation shape varies within and across
    episodes; this is why orchestration traces are
    graph-structured rather than sequence-structured
    (\S\ref{sec:orchestration}).
  \item \textbf{Number of agents is dynamic and learnable.}
    Kimi K2.5 discloses PARL training of an orchestrator that can
    spawn up to $100$ sub-agents; K2.6 extends the reported
    deployment envelope to $300$ sub-agents. We use the latter as a
    scale-pressure signal, not as independent evidence of a new
    RL-training objective. In the disclosed K2.5 setting the count is
    the output of a learned policy, not a fixed hyperparameter.
    Classical MARL fixes $n$ and trains with $n$ fixed; Shapley
    credit over a dynamic agent set is still open
    (\S\ref{sec:credit:open}).
  \item \textbf{Communication is free-form.}
    Classical MARL communication is typically a small discrete or
    continuous channel. In LLM-MAS every message is a natural-language
    utterance. This both widens the channel (agents can transmit
    plans, critiques, counterfactuals) and creates a new
    signal/credit-assignment unit (message-level signal or credit,
    \S\ref{sec:credit:hierarchy}).
  \item \textbf{Episode length is long and asynchronous.}
    Thousands of steps, hours of wall-clock time, parallel sub-agent
    execution. Rollout cost dominates RL wall-clock
    (\S\ref{sec:orchestration}), and the slowest sub-agent gates the
    whole trace.
  \item \textbf{Agents are heterogeneous by role.}
    Planner / executor / critic / verifier / summarizer. Role-based
    heterogeneity introduces role-level credit
    (MALT~\cite{malt2025}, M-GRPO~\cite{m-grpo2025}) that has no
    clean counterpart in homogeneous MARL.
  \item \textbf{Credit- and signal-bearing units are new.}
    Beyond (state, action) and agent, LLM-MAS introduces message,
    tool call, role, and orchestrator-decision as credit- and
    signal-bearing
    units (\S\ref{sec:credit}). This is the single most important
    structural difference.
\end{enumerate}

\takeaway{Classical MARL gives the language (Dec-POMDP, CTDE, COMA,
Shapley). Single-agent LLM RL gives the algorithms (PPO, GRPO,
verifiable reward, agentic rollouts). LLM-MAS adds new
credit- and signal-bearing units that neither body of work handles natively, and
that is what the rest of this survey is about.}

\section{A Working Abstraction for the Orchestration Trace}
\label{sec:formalism}

The background in \S\ref{sec:background} kept the formalism deliberately
classical. The rest of this paper rests on a thesis that does not fit
the classical mould: LLM multi-agent RL is usefully analyzed through
an \emph{orchestration trace}, a temporal interaction graph whose
vertices are \emph{events} (orchestrator decisions, sub-agent
invocations, tool calls, messages, summary returns, aggregations)
and whose vertex set itself is determined by the policy. This
section fixes the vocabulary for that object and states two informal
observations that are referenced throughout
\S\ref{sec:rewards}--\S\ref{sec:credit}.

\textbf{Scope.} Our intent here is a \emph{taxonomy formalism} for the
survey, not a fully axiomatized new MARL framework. We
introduce the minimum formal vocabulary needed to make subsequent
taxonomy claims unambiguous, and we flag the technical gaps
(off-policy evaluation of unrealized branches, exact value-function
forms over variable-shape graphs) that a follow-up theory paper
would need to address. We do not define a new solution concept,
establish equivalence to existing dynamic-agent MARL formalisms, or
prove new convergence / identifiability results.

\subsection{Relation to existing formalisms}
\label{sec:formalism:relation}

The abstraction above is closest to four existing families of
formalisms, but does not reduce cleanly to any one of them.
\textbf{Dec-POMDPs and Markov games}~\cite{dec-pomdp2002,markov-games1994}
provide the fixed-agent cooperative setting, but assume a fixed
agent index set and a joint action at each time step; LLM-MAS
orchestration must additionally represent spawn/despawn, delegation,
and aggregation events. \textbf{Hierarchical RL and options-style
controllers} supply the idea of a high-level policy choosing
temporally extended actions, but they usually treat options as
predefined action abstractions rather than as natural-language
sub-agents that communicate, call tools, and return summaries.
\textbf{Dynamic-population or open multi-agent systems} address the
changing agent set, but typically abstract away the language/tool
event graph that determines where reward and safety failures occur.
\textbf{Graph-conditioned MARL critics and communication learning}
handle graph-structured interaction, but generally assume the graph is
observed or learned as a communication topology rather than produced
by an orchestrator whose actions create and remove subgraphs.

Table~\ref{tab:formalism-comparison} makes the boundary explicit.
Our working abstraction occupies a narrower role than these
formalisms: it fixes the event vocabulary needed by this survey. The
objects that matter for later taxonomy are not only states and joint
actions, but \emph{credit- and signal-bearing events}: spawn, message, tool,
return, and aggregation nodes whose existence depends on earlier
orchestrator decisions. This is why we call the section a working
abstraction rather than a new solution concept.

\begin{table}[t]
\centering
\small
\renewcommand{\arraystretch}{1.12}
\begin{tabularx}{\linewidth}{@{}p{2.7cm}p{3.0cm}XX@{}}
\toprule
\textbf{Formalism} & \textbf{What it handles} & \textbf{What is missing for this survey} & \textbf{Role here} \\
\midrule
Dec-POMDP / Markov game
 & Fixed-agent decentralized control
 & Spawn/despawn, variable joint-action shape, language/tool events
 & Classical baseline. \\
Hierarchical RL / options
 & Meta-actions and temporally extended skills
 & Natural-language sub-agents that communicate, call tools, and return summaries
 & Analogy for orchestration actions. \\
Open / dynamic-population MAS
 & Changing agent populations
 & Trace-level message, tool, return, and aggregation events
 & Closest dynamic-agent precedent. \\
Graph-conditioned MARL critic
 & Graph-structured state or communication
 & Topology created by the orchestrator rather than only observed
 & Critic architecture precedent. \\
This paper's $\mathcal{M}^{+}$
 & Dynamic agents plus language/tool event graph
 & A new optimality theory, convergence result, or solution concept
 & Bookkeeping abstraction for taxonomy. \\
\bottomrule
\end{tabularx}
\caption{Relation between the working abstraction in this paper and
existing formalisms. The purpose of $\mathcal{M}^{+}$ is to make the
taxonomy auditable over event graphs; it is not proposed as a new
solution concept.}
\label{tab:formalism-comparison}
\end{table}

\subsection{Orchestration trace as a Dec-POMDP extension}
\label{sec:formalism:decpomdp}

\textbf{Definition 1 (Dec-POMDP, recap).}
A decentralized POMDP~\cite{dec-pomdp2002} over a Markov
game~\cite{markov-games1994} is a tuple
$\mathcal{M} = (\mathcal{I}, \mathcal{S}, \{\mathcal{A}_i\}_{i \in \mathcal{I}},
P, \{\mathcal{O}_i\}_{i \in \mathcal{I}}, \Omega, r, \gamma)$,
where $\mathcal{I} = \{1, \ldots, n\}$ is a \emph{fixed} set of agents,
$\mathcal{S}$ is the state space, $P(s' \mid s, \mathbf{a})$ is the
transition kernel under joint action
$\mathbf{a} = (a_1, \ldots, a_n)$, $\Omega(o_i \mid s, i)$ is the
observation model, $r : \mathcal{S} \times \prod_i \mathcal{A}_i \to \mathbb{R}$
is a shared reward, and $\gamma \in [0,1)$ is the discount.

\textbf{Definition 2 (Dynamic-Dec-POMDP).}
\label{def:dyn-decpomdp}
We extend $\mathcal{M}$ to accommodate spawn / despawn dynamics:
\begin{equation}
\mathcal{M}^{+} \;=\; \bigl(\mathcal{I}_t,\, \mathcal{S},\, \mathcal{A}(\cdot),\, \mathcal{A}_{\mathrm{spawn}},\, P,\, \Omega,\, r,\, \gamma\bigr),
\label{eq:dyn-decpomdp}
\end{equation}
where $\mathcal{I}_t \subseteq \mathbb{N}$ is a time-indexed agent set,
$\mathcal{A}(i,r_i,h_t)$ is the action space available to agent
instance $i$ under role $r_i$ and trace history $h_t$,
$\mathcal{A}_{\mathrm{spawn}}$ is a discrete action space over
\texttt{spawn}$(\text{role}, \text{context})$ and
\texttt{despawn}$(i)$ operations, and the global state is augmented
with the current agent count $N(s_t) = |\mathcal{I}_t|$. The
orchestrator at time $t$ is a privileged agent whose action lies in
$\mathcal{A}_{\mathrm{spawn}}$; sub-agents act in their own
$\mathcal{A}(i,r_i,h_t)$. This notation is deliberately permissive:
tool permissions, role prompts, memory access, and harness-imposed
constraints can all change the action set without requiring a new
agent identity.

\textbf{Definition 3 (Orchestration trace as event graph).}
\label{def:trace}
An orchestration trace produced by a rollout under $\mathcal{M}^{+}$
is a rooted, edge-labelled, vertex-labelled temporal graph
\begin{equation}
G \;=\; (V, E, \ell_V, \ell_E),
\label{eq:trace}
\end{equation}
whose components are:
\begin{itemize}
  \item $V = V_{\mathrm{orch}} \cup V_{\mathrm{spawn}} \cup V_{\mathrm{msg}} \cup V_{\mathrm{tool}} \cup V_{\mathrm{ret}} \cup V_{\mathrm{agg}}$: a set of \emph{events}---orchestrator decisions, sub-agent spawns, inter-agent messages, tool calls, summary returns, and aggregation steps;
  \item $E \subseteq V \times V$: a set of temporal/causal dependency edges (e.g., a tool-call event depends on the orchestrator-decision event that authorized it);
  \item $\ell_V : V \to (\text{agent}, \text{role}, \text{content})$: a vertex label assigning each event to an agent instance (drawn from $\mathcal{I}_t$ at the relevant $t$), a role, and structured content;
  \item $\ell_E : E \to \{\texttt{spawn},\, \texttt{msg},\, \texttt{return},\, \texttt{aggregate}\}$: an edge-type label.
\end{itemize}
A classical Dec-POMDP trajectory corresponds to the special case
$|V_{\mathrm{spawn}}| = 0$, $|\mathcal{I}_t| \equiv n$, and the
event sequence linearizes into the standard
$(s_0, a_0, s_1, a_1, \ldots)$ form. Definition~\ref{def:trace} is
the same object used in the visual schematic of
Figure~\ref{fig:orchestration-trace} and the credit hierarchy of
\S\ref{sec:credit:hierarchy}: events, not agents, are the carriers
of credit.

\subsection{Value function under variable-shape traces}
\label{sec:formalism:value}

In the fixed-$n$ setting the value of a joint policy
$\boldsymbol{\pi}$ is
$V^{\boldsymbol{\pi}}(s) = \mathbb{E}\bigl[\sum_{t \ge 0} \gamma^t r_t \mid s_0 = s\bigr]$,
and CTDE methods such as MADDPG~\cite{maddpg2017} and
MAPPO~\cite{mappo2022} learn $V^{\boldsymbol{\pi}}(s)$ or
$Q^{\boldsymbol{\pi}}(s, \mathbf{a})$ over a fixed-shape joint.
Under $\mathcal{M}^{+}$ neither $s$ nor $\mathbf{a}$ has a fixed
dimension: the set of active sub-agents, and hence the shape of
$\mathbf{a}$, is itself a random variable. We therefore parameterize
the value object by the trace prefix:
\begin{equation}
V^{\boldsymbol{\pi}}(G_{\le t}) \;=\; \mathbb{E}_{\boldsymbol{\pi}}\!\left[\sum_{\tau \ge t} \gamma^{\tau - t} r_\tau \;\Big|\; G_{\le t}\right],
\label{eq:value-graph}
\end{equation}
where $G_{\le t}$ is the sub-graph of $G$ induced by events with
timestamp $\le t$. For orientation, the corresponding one-step
bookkeeping identity can be written in graph-conditioned form,
\begin{equation}
V^{\boldsymbol{\pi}}(G_{\le t}) \;=\; \mathbb{E}_{\boldsymbol{\pi}}\!\bigl[r_t + \gamma\, V^{\boldsymbol{\pi}}(G_{\le t+1}) \,\big|\, G_{\le t}\bigr],
\label{eq:bellman-graph}
\end{equation}
whose essential difference from the standard Bellman equation is that
$G_{\le t+1}$ may contain a \emph{larger} vertex set than
$G_{\le t}$ (when a spawn event fires): the expectation is taken
over transitions that grow or shrink the event graph, not only over
transitions in a fixed joint state space.
Equation~\eqref{eq:bellman-graph} is a trace-conditioned value
identity that motivates the graph- or trace-conditioned critics in
\S\ref{sec:credit}; it is not a new solution concept or convergence
claim. Concrete algorithmic forms (which sub-graph features
are sufficient, how to amortize $V$ across variable-shape inputs)
are open and are discussed in \S\ref{sec:open}.

\subsection{Two organizing observations}
\label{sec:formalism:claims}

The remainder of this paper invokes two informal observations
about $\mathcal{M}^{+}$. They are intended to organize the taxonomy,
not to function as theorem statements: each is supported by the
qualitative argument below and by the empirical evidence cited, but
neither is accompanied by a formal proof or rate in this paper.

\textbf{Observation 1 (Credit diffusion under uniform credit).}
\label{claim:diffusion}
\emph{Under a shared terminal team reward $R$, uniform credit
allocation across $n$ credit- or signal-bearing units of an orchestration
trace (e.g., tokenwise GAE with a single team baseline, as in naive
GRPO on the concatenated trace), and no structure-specific baseline,
the effective per-decision signal available to any single unit tends
to become less distinguishable as trace length grows.}

\emph{Argument.}
A uniform allocation distributes the terminal reward equally over
the $n$ units; the shared baseline removes the mean. The remaining
per-unit signal is dominated by the residual variance of $R$ shared
among $n$ units. As $n$ grows with trace length, the per-unit
signal becomes increasingly difficult to distinguish from
baseline-estimation noise. Concrete order-of-magnitude scaling
depends on the exact noise model (we leave a precise rate to a
follow-up theory paper); what matters here is the qualitative
fragility: longer traces with unstructured credit make it harder to
identify the contribution of any individual decision. This is the
same pathology that motivated COMA-style counterfactual
baselines~\cite{coma2018}, Shapley credit~\cite{shapley-q2020}, and
difference rewards~\cite{difference-rewards2001} in classical MARL,
and that motivates the credit-decomposition methods surveyed in
\S\ref{sec:credit}. Empirically, Dr.\,MAS~\cite{dr-mas2026}
documents the resulting training instability when GRPO is applied
naively to multi-agent rollouts.

\textbf{Observation 2 (Non-identifiability of same-prefix
counterfactual orchestrator credit).}
\label{claim:nonid}
\emph{Let $\pi_{\mathrm{orch}}$ be the orchestrator policy and let
$d_t \in \{\texttt{spawn}, \texttt{no-spawn}\}$ be its decision at
time $t$. Without an off-policy evaluation mechanism, the
same-prefix counterfactual effect
$\mathbb{E}[R \mid G_{\le t}, \texttt{spawn}]
- \mathbb{E}[R \mid G_{\le t}, \texttt{no-spawn}]$
is not identifiable from realized on-policy traces alone unless both
branches have coverage or additional structural assumptions are made.}

\emph{Argument.} A marginal association such as
$\mathbb{E}[R \mid d_t]$ can be estimated from logged on-policy data,
but it mixes different trace prefixes and therefore does not isolate
the causal contribution of the decision at a fixed prefix. Classical
$Q$-learning identifies $Q(s, a)$ only when every action $a$ is
occasionally sampled at comparable states $s$. For an orchestrator's
spawn decision, the un-taken branch
($\texttt{no-spawn}$, when $\texttt{spawn}$ was chosen) produces a
structurally different trace---no sub-graph is generated for it---so
on-policy rollouts furnish no realizations of the counterfactual.
Estimating the same-prefix contribution of the spawn decision
therefore requires either an explicit off-policy mechanism (e.g., a
learned counterfactual value function trained on alternative-branch
data) or strong structural assumptions. This is the conceptual
anchor of the ``counterfactual ambiguity'' discussion in
\S\ref{sec:credit:fails} and of the open problem flagged in
\S\ref{sec:open}.

\subsection{The reward--credit dual}
\label{sec:formalism:dual}

The two claims above motivate a trade-off that organizes the
next three sections. Each reward family in \S\ref{sec:rewards}
picks a \emph{privileged layer} in the credit hierarchy: outcome
rewards privilege the terminal-trajectory layer; process rewards
privilege the step or turn layer; orchestration rewards deliver a
dense signal directly at the orchestrator-decision event. The
denser the reward, the smaller the effective number of units over
which the diffusion in Claim~\ref{claim:diffusion} acts, and the
less the credit decomposition in \S\ref{sec:credit} must do to
recover a usable per-unit signal. Conversely, sparse terminal
rewards shift the burden onto credit assignment---counterfactual
baselines, role-level critics, Shapley-style attribution---because
the reward does not itself pick a layer. This duality explains why
two apparently opposite design choices (process rewards with weak
credit decomposition, vs.\ sparse outcome rewards with strong
credit machinery) can both be empirically competitive: they place
the same total work on different sides of the reward--credit ledger.

\takeaway{The orchestration trace $G = (V, E, \ell_V, \ell_E)$ is
an event graph drawn from the dynamic-Dec-POMDP $\mathcal{M}^{+}$
of Definition~\ref{def:dyn-decpomdp}. Value functions are naturally
conditioned on $G$ rather than on a fixed-shape state; uniform
credit on long shared-reward traces can make per-unit signal
fragile (Claim~\ref{claim:diffusion}); and
orchestrator spawn decisions are non-identifiable from on-policy
rollouts alone (Claim~\ref{claim:nonid}). Dense rewards and strong
credit decomposition are duals: entries in
\S\ref{sec:rewards}--\S\ref{sec:credit} differ in where they place
that burden.
The framework here is intended as an organizing abstraction for the
survey, not a new MARL theory; concrete algorithmic forms are open
(\S\ref{sec:open}).}

\section{System Forms: How LLM Agent Teams Are Organized}
\label{sec:systems}

Before algorithms, we fix the system object: the concrete topologies
in which LLM agents are organized, both in open literature and in
deployed products. This ordering matters for two reasons. First,
every RL method in later sections optimizes \emph{some} system form;
the form determines what can be rewarded and where credit can flow.
Second, public industrial systems expose engineering pressures
(rollout cost, asynchrony, harness design) that discipline which RL
methods are practical---a constraint missing from most academic
benchmarks.

\subsection{A typology of agent-team topologies}
\label{sec:systems:typology}

Six patterns recur across the paper pool
(Table~\ref{tab:topologies}). They are not mutually exclusive;
production systems typically combine two or three.

\begin{table}[htbp]
\centering
\small
\renewcommand{\arraystretch}{1.25}
\begin{tabularx}{\linewidth}{@{}p{3.4cm} X X@{}}
\toprule
\textbf{Topology} & \textbf{Defining feature} & \textbf{Representative instances} \\
\midrule
Centralized orchestrator + sub-agents
 & One orchestrator dispatches tasks to a pool of sub-agents and aggregates results.
 & Kimi Agent Swarm~\cite{kimi-k2-5-2026}, M-GRPO main/sub~\cite{m-grpo2025}, Puppeteer~\cite{puppeteer2025}, WideSeek-R1~\cite{wideseek-r1-2026} \\
Planner--executor--critic
 & Three specialized roles with distinct rubrics; critic closes the loop.
 & MALT~\cite{malt2025}, MATPO~\cite{matpo2025}, MAE~\cite{mae2025} \\
Debate / committee
 & Multiple agents argue, a resolver decides; credit is message-level.
 & Debate-as-Reward~\cite{debate-as-reward2026}, LatentMAS~\cite{latentmas2025} \\
Parallel swarm
 & Many near-homogeneous agents run concurrently, then aggregate.
 & Kimi PARL~\cite{kimi-k2-5-2026,kimi-k2-6-2026}, Anthropic parallel Claudes~\cite{anthropic-c-compiler2026} \\
Hierarchical agents
 & Multi-level spawn; agents at level $k$ can spawn level-$(k{+}1)$.
 & HALO~\cite{halo2025}, AgentSpawn~\cite{agentspawn2026}, DEPART~\cite{depart2026}, LAMO~\cite{lamo2026} \\
Managed / harness-based
 & System harness wraps model, tools, prompts, execution; agents live inside.
 & OpenAI Codex~\cite{openai-codex2025}, Claude Code~\cite{claude-code-subagents-2025}, Agent~Lightning~\cite{agent-lightning2025} \\
\bottomrule
\end{tabularx}
\caption{Six recurring agent-team topologies. Every method in our
pool fits one or more; the topology constrains which reward
families (\S\ref{sec:rewards}) and credit levels
(\S\ref{sec:credit}) are even definable.}
\label{tab:topologies}
\end{table}

\begin{figure}[t]
\centering
\resizebox{\linewidth}{!}{%
\begin{tikzpicture}[
    font=\scriptsize,
    orch/.style ={circle, draw=red!75, thick, fill=red!12,
                  minimum size=4mm, inner sep=0pt},
    sub/.style  ={circle, draw=blue!65, thick, fill=blue!10,
                  minimum size=3.5mm, inner sep=0pt},
    plan/.style ={rectangle, rounded corners=1pt, draw=red!75, thick,
                  fill=red!12, minimum width=7mm, minimum height=3.2mm,
                  inner sep=1pt, font=\tiny},
    exec/.style ={rectangle, rounded corners=1pt, draw=blue!65, thick,
                  fill=blue!10, minimum width=7mm, minimum height=3.2mm,
                  inner sep=1pt, font=\tiny},
    crit/.style ={rectangle, rounded corners=1pt, draw=orange!85, thick,
                  fill=orange!15, minimum width=7mm, minimum height=3.2mm,
                  inner sep=1pt, font=\tiny},
    judge/.style={diamond, draw=green!50!black, thick, fill=green!15,
                  minimum size=5mm, aspect=1.5, inner sep=0pt, font=\tiny},
    harness/.style={rectangle, draw=black!60, thick, dashed,
                    fill=gray!8, rounded corners=2pt,
                    inner sep=2.5mm},
    arr/.style  ={-{Stealth[length=1.6mm]}, thick, black!60},
    bidir/.style={{Stealth[length=1.4mm]}-{Stealth[length=1.4mm]}, thick, black!55},
    clab/.style  ={font=\footnotesize\bfseries, anchor=north, align=center},
    note/.style ={font=\tiny\itshape, black!60, anchor=north, align=center}
]


\begin{scope}[shift={(0, 0)}]
  \node[orch] (oA) at (1.6, 0.6) {O};
  \node[sub] (a1) at (0.4, -0.6) {};
  \node[sub] (a2) at (1.2, -0.6) {};
  \node[sub] (a3) at (2.0, -0.6) {};
  \node[sub] (a4) at (2.8, -0.6) {};
  \foreach \t in {a1,a2,a3,a4} \draw[arr] (oA) -- (\t);
  \node[clab] at (1.6, -1.05) {(a) centralized};
  \node[note] at (1.6, -1.55) {Kimi PARL, M-GRPO,\\Puppeteer};
\end{scope}

\begin{scope}[shift={(4.4, 0)}]
  \node[plan]  (p) at (0.4,  0.5) {planner};
  \node[exec]  (e) at (1.6,  0.5) {executor};
  \node[crit]  (c) at (2.8,  0.5) {critic};
  \draw[arr] (p) -- (e);
  \draw[arr] (e) -- (c);
  \draw[arr] (c.south) .. controls (2.8, -0.4) and (0.4, -0.4) ..
    node[below=-1pt, midway, font=\tiny\itshape, black!60]{revise} (p.south);
  \node[clab] at (1.6, -1.05) {(b) planner-executor-critic};
  \node[note] at (1.6, -1.55) {MALT, MATPO, MAE};
\end{scope}

\begin{scope}[shift={(8.8, 0)}]
  \node[sub] (d1) at (0.55, 0.2) {};
  \node[sub] (d2) at (1.65, 0.7) {};
  \node[sub] (d3) at (2.75, 0.2) {};
  \node[judge] (j) at (1.65, -0.8) {res.};
  \draw[bidir] (d1) -- (d2);
  \draw[bidir] (d2) -- (d3);
  \draw[bidir] (d1) -- (d3);
  \foreach \d in {d1, d2, d3} \draw[arr, dashed] (\d) -- (j);
  \node[clab] at (1.65, -1.4) {(c) debate / committee};
  \node[note] at (1.65, -1.9) {Debate-as-Reward,\\LatentMAS};
\end{scope}

\begin{scope}[shift={(0, -4.2)}]
  \node[orch] (oD) at (1.6, 0.7) {O};
  \foreach \i/\x in {1/0.2, 2/0.7, 3/1.2, 4/1.7, 5/2.2, 6/2.7} {
    \node[sub, scale=0.85] (s\i) at (\x, -0.15) {};
  }
  \foreach \i/\x in {1/0.45, 2/1.0, 3/1.55, 4/2.1, 5/2.65} {
    \node[sub, scale=0.85] (t\i) at (\x, -0.7) {};
  }
  \foreach \i in {1,...,6} \draw[arr, opacity=0.6] (oD) -- (s\i);
  \node[clab] at (1.6, -1.2) {(d) parallel swarm};
  \node[note] at (1.6, -1.7) {Kimi K2.5/K2.6,\\parallel Claudes};
\end{scope}

\begin{scope}[shift={(4.4, -4.2)}]
  \node[orch]  (h0) at (1.6,  0.7) {O};
  \node[sub]   (h1) at (0.6, -0.05) {};
  \node[sub]   (h2) at (1.6, -0.05) {};
  \node[sub]   (h3) at (2.6, -0.05) {};
  \node[sub, scale=0.85] (h21) at (1.2, -0.85) {};
  \node[sub, scale=0.85] (h22) at (2.0, -0.85) {};
  \draw[arr] (h0) -- (h1);
  \draw[arr] (h0) -- (h2);
  \draw[arr] (h0) -- (h3);
  \draw[arr] (h2) -- (h21);
  \draw[arr] (h2) -- (h22);
  \node[clab] at (1.6, -1.4) {(e) hierarchical};
  \node[note] at (1.6, -1.9) {HALO, AgentSpawn};
\end{scope}

\begin{scope}[shift={(8.8, -4.2)}]
  \node[harness, fit={(0.0, -1.05) (3.3, 1.0)}] (hbox) {};
  \node[orch] (oF) at (1.65, 0.55) {O};
  \node[sub]  (sF1) at (0.6, -0.4) {};
  \node[sub]  (sF2) at (1.65, -0.4) {};
  \node[sub]  (sF3) at (2.7, -0.4) {};
  \node[font=\tiny, black!55] at (0.4, -0.85) {tool};
  \node[font=\tiny, black!55] at (1.65, -0.85) {tool};
  \node[font=\tiny, black!55] at (2.9, -0.85) {tool};
  \draw[arr] (oF) -- (sF1);
  \draw[arr] (oF) -- (sF2);
  \draw[arr] (oF) -- (sF3);
  \node[font=\tiny\itshape, black!55, anchor=south west]
    at (hbox.north west) {harness};
  \node[clab] at (1.65, -1.4) {(f) managed / harness-based};
  \node[note] at (1.65, -1.9) {OpenAI Codex, Claude Code,\\Agent~Lightning};
\end{scope}

\end{tikzpicture}%
}
\caption{Visual schematics of the six recurring LLM-MAS topologies
catalogued in Table~\ref{tab:topologies}. Red ($\bigcirc$/box) =
orchestrator or planner; blue ($\bigcirc$) = sub-agent or executor;
orange (box) = critic; green diamond = debate resolver; dashed outer
box = managed harness. Solid arrows = delegation; dashed arrows =
voting / aggregation; double-headed arrows = bidirectional debate.
The topology constrains which credit-bearing units
(\S\ref{sec:credit:hierarchy}) are easiest to define: (a)/(d)/(e)
make orchestrator-level credit most natural; (c) makes
message-level credit a natural primary signal, although logged
messages can be credited in other topologies; only (f) admits a
harness-level boundary as a training-frozen interface.}
\label{fig:topologies}
\end{figure}
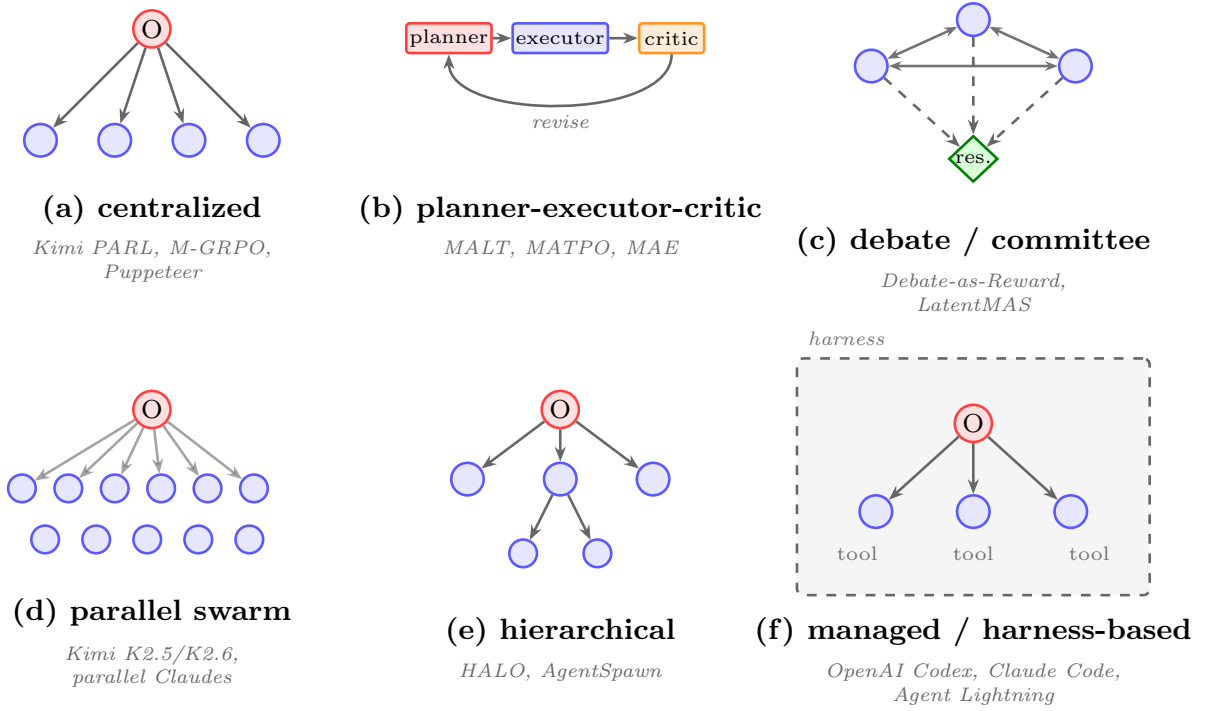

\noindent Two observations.

\begin{itemize}
  \item \textbf{Topology determines credit affordance.} Centralized
    orchestrator topologies make
    \emph{orchestrator-level credit} easiest to define; debate
    topologies make \emph{message-level credit} a natural primary
    signal. Methods that target these levels
    necessarily commit to a compatible topology.
  \item \textbf{Harness-based systems are a research-opaque majority.}
    Codex and Claude Code are in our pool as cases, not as methods,
    because their RL training recipes are not publicly disclosed.
    What \emph{is} disclosed is the harness: model $\oplus$ tools
    $\oplus$ prompts $\oplus$ execution logic. Any RL method that
    targets these systems must respect harness shape, even if the
    harness is not itself being trained.
\end{itemize}

\subsection{Public industrial evidence and selection rule}
\label{sec:systems:industry}

The industrial discussion is not intended as a census of all deployed
agent products. We retain only public industrial sources that satisfy
at least one load-bearing criterion for this paper: they disclose a
trained orchestration mechanism, expose a stable harness or
sub-agent interface that constrains future RL design, or document
long-running parallel workflows at a scale not represented in open
academic benchmarks. Under this rule, the retained industrial
\emph{anchors} are Kimi Agent Swarm, OpenAI Codex, and Anthropic
Claude Code / parallel-Claude engineering reports. They are
representative of three evidence roles rather than of the whole
commercial agent market: Kimi provides the public trained-orchestrator
anchor; Codex provides the cloud harness and parallel software-agent
workflow anchor; Claude Code provides the sub-agent interface and
parallel-team workflow anchor.

Several well-known frameworks and products are therefore not treated
as main industrial anchors. AutoGen, CAMEL, MetaGPT, CrewAI,
LangGraph, Devin-related product material, and OpenAI Swarm are
valuable context, but the screened public records either do not
disclose RL/post-training mechanisms, do not provide enough stable
technical detail for the evidence ledger, or are better used as
background framework examples. Appendix~\ref{app:artifact-protocol}
logs these borderline decisions. This selection rule is deliberately
conservative: it reduces coverage breadth in exchange for a clearer
claim boundary.

Table~\ref{tab:evidence-status} records how we use industrial
materials. This is important because blogs and documentation can
support claims about deployment shape, scale, interfaces, and user
workflow, but they do not by themselves make an algorithm
reproducible.

\begin{table}[htbp]
\centering
\small
\renewcommand{\arraystretch}{1.2}
\begin{tabularx}{\linewidth}{@{}p{2.8cm} p{2.4cm} X X@{}}
\toprule
\textbf{Source class} & \textbf{Examples} & \textbf{Used for} & \textbf{Not used for} \\
\midrule
Peer-reviewed / arXiv methods
 & MALT, MAPoRL, Puppeteer, Dr.\,MAS
 & Algorithmic mechanisms, training regimes, reported ablations
 & Claims beyond the paper's evaluation setting \\
Company technical reports
 & Kimi K2.5 / PARL
 & Publicly disclosed training shapes, reward components, system scale
 & Full reproducibility when optimizer, data, or ablations are absent \\
Product docs / launch blogs
 & Codex, Claude Code, Kimi K2.6, Claw Groups
 & Deployment form, harness boundary, scale, user-facing affordances
 & Undisclosed RL objectives or optimizer details \\
Engineering case studies
 & Anthropic parallel Claudes C-compiler case
 & Long-running agent-team workflow, cost, harness design pressure
 & General claims about model training or benchmark superiority \\
\bottomrule
\end{tabularx}
\caption{Evidence levels used for industrial systems. We distinguish
deployment-shape evidence from reproducible algorithmic evidence; this
prevents product documentation from being treated as equivalent to a
peer-reviewed RL method.}
\label{tab:evidence-status}
\end{table}

The more compact source-class matrix used during screening is moved
to Appendix~\ref{app:artifact-protocol}; Table~\ref{tab:evidence-status}
and Table~\ref{tab:claim-ledger} are the load-bearing evidence
controls in the main text.

\begin{table}[t]
\centering
\small
\renewcommand{\arraystretch}{1.12}
\begin{tabularx}{\linewidth}{@{}p{3.2cm}p{2.6cm}p{2.0cm}X@{}}
\toprule
\textbf{Claim used here} & \textbf{Primary source} & \textbf{Confidence} & \textbf{Boundary} \\
\midrule
Kimi K2.5 trains an orchestrator with PARL
 & Kimi technical report & high & Used as the only public industrial anchor in our pool that explicitly discloses RL training of the orchestrator. \\
Kimi K2.5 reports up to $100$ sub-agents and $1{,}500$ coordinated steps / tool calls
 & Kimi technical report & high & Used for publicly reported deployment-envelope evidence; full optimizer/data/ablation details are not reproducible. \\
Kimi K2.6 reports up to $300$ sub-agents and $4{,}000$ coordinated steps
 & official Kimi blog / product material & medium-high & Used as deployment-envelope evidence; not used as a reproducible training claim. \\
OpenAI Codex exposes parallel software-agent workflows and a harness boundary
 & OpenAI product material & high & Used as deployment-shape evidence; we do not claim a public multi-agent RL objective. \\
Claude Code exposes sub-agents and custom sub-agent interfaces
 & Anthropic documentation & high & Used as harness and spawn/delegation evidence; not as evidence of RL-trained orchestration. \\
Anthropic C-compiler project used parallel Claude Code sessions
 & Anthropic engineering case study & medium & Used as workflow-shape and cost-pressure evidence; not as a model-training result. \\
\bottomrule
\end{tabularx}
\caption{Claim-confidence ledger for industrial evidence. This table
makes explicit which claims are supported by public material and which
claims are intentionally \emph{not} made.}
\label{tab:claim-ledger}
\end{table}

\subsubsection{Kimi Agent Swarm (K2.5 / K2.6)}
\label{sec:systems:kimi}

Moonshot's Kimi~K2.5 is the most openly documented industrial instance
of trained orchestration in our pool. The K2.5 report introduces
Parallel-Agent Reinforcement Learning (PARL) and describes a swarm
scaling to up to $100$ sub-agents and $1{,}500$ coordinated steps /
tool calls as reported~\cite{kimi-k2-5-2026}. The K2.6 product and
technical materials scale the deployment envelope to $300$ sub-agents
and $4{,}000$ coordinated steps and add Claw Groups, a research preview of
cross-vendor and human-in-the-loop coordination~\cite{kimi-k2-6-2026}.

What makes Kimi the main industrial reference point under this
evidence boundary:

\begin{itemize}
  \item The orchestrator is a \emph{learned} policy, not a prompt
    template. Sub-agent creation is an action in its action space.
  \item The reward decomposes as
    $r_{\text{perf}} + \lambda_1 r_{\text{parallel}} + \lambda_2 r_{\text{finish}}$
    (\S\ref{sec:rewards:parl})---a published instance of an R7+R8
    composition with staged annealing.
  \item The ``Critical-Steps'' metric functions as an orchestrator-level
    credit signal (\S\ref{sec:credit}): it distinguishes real
    parallel progress from padded traces, penalizing
    pseudo-parallelism at the orchestrator level.
\end{itemize}

Kimi is therefore the industrial anchor for our thesis, but with an
important evidence boundary: the public materials disclose enough to
identify learned orchestration, reward shaping, and orchestrator-level
signals, but not enough to reproduce the full training recipe. More
importantly, the scale-gap argument is anchored by Kimi rather than
established uniformly across industrial systems; Codex, Claude~Code,
and related public systems are evidence for harness and workflow shape,
not for comparable disclosed RL trace scale.

\subsubsection{OpenAI Codex (app + harness)}
\label{sec:systems:codex}

Codex~\cite{openai-codex2025} is described in OpenAI's launch
material as a cloud-native parallel software-engineering agent
orchestrated from a single ``command center.'' Two features matter
for RL.

First, the \emph{harness}---model $\oplus$ tools $\oplus$ prompts
$\oplus$ execution logic---is the unit of deployment, not the model
alone. Any RL training for a harness-hosted agent must treat the
harness boundary as fixed during training, and as part of the
observation and action interface at inference. Agent~Lightning's
execution/training decoupling~\cite{agent-lightning2025} is, in
effect, an academic articulation of this constraint.

Second, the Codex UI makes parallel workflows and long-running tasks
first-class. This is significant because long-horizon parallel
rollouts are the most expensive rollouts to train on: a single
rollout can be minutes of wall clock and hundreds of tool calls.
RL algorithms that require many rollouts (GRPO in particular, with
its group of $K$) become impractical without the kind of
pipeline-parallel scheduling that MarsRL~\cite{marsrl2025} targets.

Codex's RL training recipe is not publicly disclosed; we cite it
here as a system-design anchor, not as an algorithmic data point.

\subsubsection{Anthropic Claude Code (sub-agents + agent teams)}
\label{sec:systems:claude-code}

Claude Code documentation~\cite{claude-code-subagents-2025}
specifies built-in sub-agents (Explore, Plan, general-purpose) and a
user-facing API for custom sub-agents, with a lead agent that
dispatches subtasks and aggregates results. Anthropic's engineering
case study of sixteen parallel Claudes jointly building a C compiler
over roughly $2{,}000$ sessions and $\sim\!100{,}000$ lines of
Rust~\cite{anthropic-c-compiler2026} is the largest public
multi-agent code-generation case study we are aware of.

Two aspects of Claude Code are load-bearing for the rest of this
paper.

\begin{itemize}
  \item \textbf{Sub-agent as a first-class object.} Claude Code's
    sub-agent API makes the spawn / delegate / aggregate pattern a
    concrete object that can be the subject of RL, not just a
    prompt-engineering convention.
  \item \textbf{Explicit steerability concern.} Anthropic's
    trustworthy-agents framework~\cite{anthropic-trustworthy2025}
    explicitly flags that sub-agents make it harder for users to
    understand and steer a workflow mid-execution. This is a
    credit-assignment-shaped concern: where in the trace can the
    human intervene, and what are the downstream consequences?
    We return to this in \S\ref{sec:open}.
\end{itemize}

\subsection{What systems reveal that papers do not}
\label{sec:systems:reveal}

Reading across the three systems against the paper pool surfaces
three gaps between what is deployed and what is published.

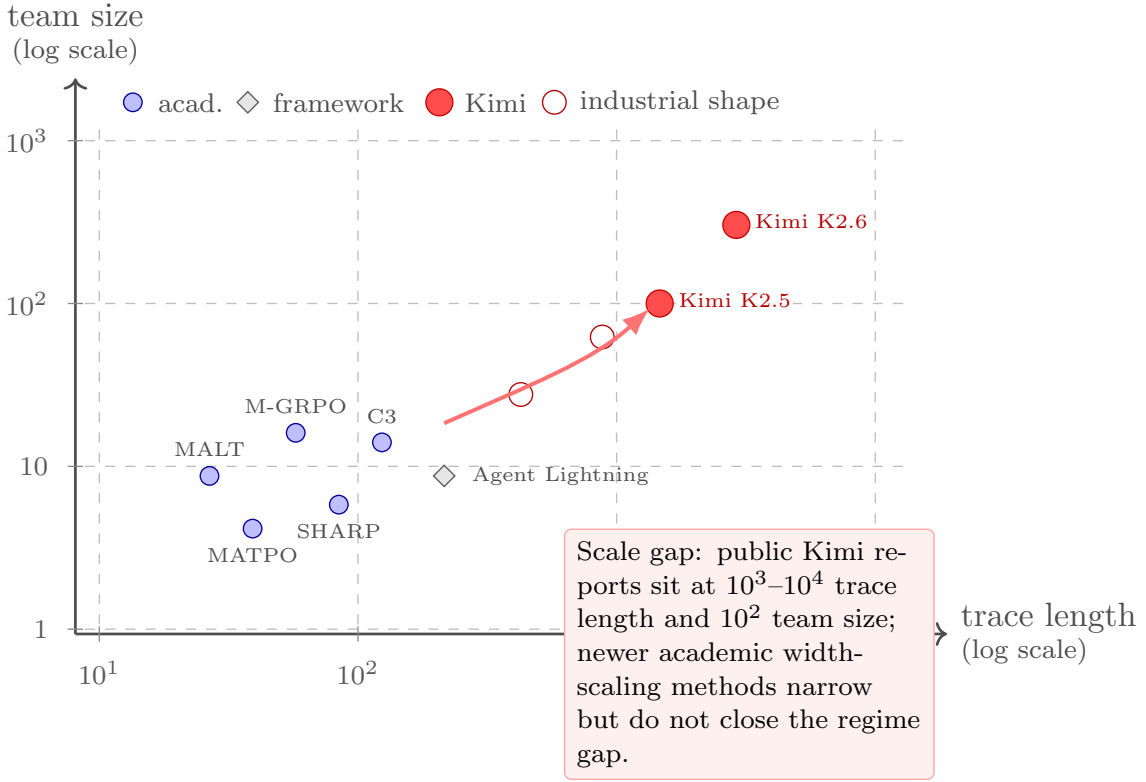
\begin{figure}[t]
\centering
\resizebox{0.96\linewidth}{!}{%
\begin{tikzpicture}[
  font=\footnotesize,
  academic/.style={circle, draw=blue!60!black, fill=blue!25, minimum size=5.5pt, inner sep=0pt},
  framework/.style={diamond, draw=gray!70!black, fill=gray!20, minimum size=6.5pt, inner sep=0pt},
  industry/.style={circle, draw=red!70!black, fill=red!30, minimum size=6.5pt, inner sep=0pt},
  indanchor/.style={circle, draw=red!80!black, fill=red!70, minimum size=8pt, inner sep=0pt},
  hollow/.style={circle, draw=red!70!black, fill=white, minimum size=7pt, inner sep=0pt},
  guide/.style={black!25, dashed, thin}
]

  \draw[->, thick, black!70] (0.45, 0.45) -- (9.55, 0.45)
    node[right, align=left] {\small trace length\\[-2pt]\scriptsize(log scale)};
  \draw[->, thick, black!70] (0.45, 0.45) -- (0.45, 6.25)
    node[above, align=center] {\small team size\\[-2pt]\scriptsize(log scale)};

  \foreach \x/\lab in {0.7/$10^1$, 3.4/$10^2$, 6.1/$10^3$, 8.8/$10^4$} {
    \draw[black!45] (\x,0.37) -- (\x,0.53);
    \node[below=2pt, font=\scriptsize, black!65] at (\x,0.37) {\lab};
    \draw[guide] (\x,0.55) -- (\x,5.75);
  }
  \foreach \y/\lab in {0.5/$1$, 2.2/$10$, 3.9/$10^2$, 5.6/$10^3$} {
    \draw[black!45] (0.37,\y) -- (0.53,\y);
    \node[left=2pt, font=\scriptsize, black!65] at (0.37,\y) {\lab};
    \draw[guide] (0.55,\y) -- (9.15,\y);
  }

  \node[academic, label={[font=\tiny, black!70, label distance=-1pt]above:MALT}]
       at (1.85, 2.10) {};
  \node[academic, label={[font=\tiny, black!70, label distance=-1pt]below:MATPO}]
       at (2.30, 1.55) {};
  \node[academic, label={[font=\tiny, black!70, label distance=-1pt]above:M-GRPO}]
       at (2.75, 2.55) {};
  \node[academic, label={[font=\tiny, black!70, label distance=-1pt]below:SHARP}]
       at (3.20, 1.80) {};
  \node[academic, label={[font=\tiny, black!70, label distance=-1pt]above:C3}]
       at (3.65, 2.45) {};
  \node[framework, label={[font=\tiny, black!70, label distance=1pt]right:Agent Lightning}]
       at (4.30, 2.10) {};

  \node[indanchor] at (6.55, 3.9) {};
  \node[anchor=west, font=\tiny, red!70!black, fill=white,
        fill opacity=0.9, text opacity=1, inner sep=0.6pt]
        at (6.72, 3.94) {Kimi K2.5};
  \node[indanchor] at (7.35, 4.72) {};
  \node[anchor=west, font=\tiny, red!70!black, fill=white,
        fill opacity=0.9, text opacity=1, inner sep=0.6pt]
        at (7.52, 4.76) {Kimi K2.6};

  \node[hollow] at (5.10, 2.95) {};
  \node[hollow] at (5.95, 3.55) {};

  \draw[-Latex, very thick, red!55]
    (4.30, 2.65) .. controls (5.20, 3.05) and (5.85, 3.30) .. (6.45, 3.85);
  \node[draw=red!35, fill=red!6, rounded corners=2pt, align=left,
        text width=3.6cm, font=\scriptsize, anchor=north west]
    at (5.55, 1.55)
    {Scale gap: public Kimi reports sit at $10^3$--$10^4$ trace
     length and $10^2$ team size; newer academic width-scaling methods
     narrow but do not close the regime gap.};

  \node[academic] at (1.05, 6.0) {};
  \node[anchor=west, font=\scriptsize, black!70] at (1.18, 6.0) {acad.};
  \node[framework] at (2.25, 6.0) {};
  \node[anchor=west, font=\scriptsize, black!70] at (2.38, 6.0) {framework};
  \node[indanchor] (legkimi) at (4.25, 6.0) {};
  \node[anchor=west, font=\scriptsize, black!70] at (4.38, 6.0) {Kimi};
  \node[hollow] at (5.45, 6.0) {};
  \node[anchor=west, font=\scriptsize, black!70] at (5.58, 6.0) {industrial shape};

\end{tikzpicture}%
}
\caption{Industry--academia scale gap. Reading: blue points summarize
the typical public evaluation regime of academic LLM-MAS RL methods,
while red filled points mark Kimi reports that disclose both team size
and long trace length. Hollow red points indicate industrial
deployment-shape evidence where the public material is useful for
harness and workflow analysis but does not disclose a comparable RL
training scale. Positions are approximate and log-scaled; the figure is
intended to show the regime gap, not a leaderboard.}
\label{fig:scale-gap}
\end{figure}

\begin{table}[htbp]
\centering
\small
\renewcommand{\arraystretch}{1.12}
\begin{tabularx}{\linewidth}{@{}>{\raggedright\arraybackslash}p{2.6cm}
>{\raggedright\arraybackslash}p{3.9cm}X@{}}
\toprule
\textbf{Point in Fig.~\ref{fig:scale-gap}} &
\textbf{Public scale signal} &
\textbf{Source status} \\
\midrule
Academic RL methods
 & mostly $10$--$100$-step traces; small-to-moderate teams
 & Representative operating range read from public evaluations in
   the retained pool. WideSeek-R1 and MARTI-MARS$^2$ begin to stress
   width/self-search scaling, but remain below the disclosed Kimi
   envelope; this row is a regime summary, not a leaderboard. \\
Agent Lightning / harness frameworks
 & $10$--$100$-step traces; small teams
 & Public material is most useful for the training-harness boundary;
   comparable Kimi-reported training traces are not disclosed. \\
Kimi K2.5
 & $1{,}500$ coordinated steps / tool calls; up to $100$ sub-agents
 & Public company report used as the disclosed deployment-envelope
   anchor~\cite{kimi-k2-5-2026}. \\
Kimi K2.6
 & $4{,}000$ coordinated steps; up to $300$ sub-agents
 & Public company report used as the disclosed deployment-envelope
   extension~\cite{kimi-k2-6-2026}. \\
\bottomrule
\end{tabularx}
\caption{Source status for the scale-gap figure. The figure compares
operating regimes disclosed in public material; it does not claim that
all industrial systems train with Kimi-reported RL traces.}
\label{tab:scale-gap-sources}
\end{table}

\begin{itemize}
  \item \textbf{Rollout-cost realism.} Academic methods mostly evaluate on
    $10$--$100$-step orchestration traces; newer width-scaling and
    self-search methods such as WideSeek-R1~\cite{wideseek-r1-2026}
    and MARTI-MARS$^2$~\cite{marti-mars2-2026} narrow the conceptual
    gap but do not disclose Kimi-reported training traces. Public Kimi reports
    disclose $1{,}500$--$4{,}000$-step traces, while other industrial
    reports expose the harness and workflow pressures without
    comparable training-scale disclosure. Credit diffusion
    (\S\ref{sec:credit:fails}) becomes qualitatively worse in this
    regime, and within our curated pool no open academic method publicly
    reports training at the Kimi-reported trace lengths.
  \item \textbf{Harness as fixed context.} All three industrial
    systems expose a stable harness boundary; for any RL or
    post-training layer, that boundary would constrain the learnable
    policy. The academic literature mostly does not make this
    boundary explicit; methods are typically trained in a bespoke
    rollout environment. Agent~Lightning~\cite{agent-lightning2025}
    is the clearest counterexample.
  \item \textbf{Steerability as an RL target.}
    Industrial safety frameworks~\cite{anthropic-trustworthy2025}
    treat mid-execution human intervention as a first-class concern.
    We found no paper in our pool that formulates steerability as an
    RL objective; this is a named open problem in \S\ref{sec:open}.
\end{itemize}

\takeaway{Topology determines which rewards and credit assignments
are even definable; public industrial evidence, most clearly the Kimi
reports, discloses deployment envelopes and harness constraints beyond
most open academic evaluation regimes. The gap is not primarily
algorithmic---it is in \emph{what shape of rollout and harness
academic methods train against}.}

\section{Systems Engineering: Rollout Cost and Harness Boundary}
\label{sec:engineering}

The system forms cataloged in \S\ref{sec:systems} and the event-graph
formalism of \S\ref{sec:formalism} imply concrete engineering
constraints that academic RL methods cannot ignore. Like
\S\ref{sec:formalism}, the material in this section is \emph{not}
an independent theory contribution: it supplies the operational
back-pressure (rollout cost, harness shape, trace length) that
disciplines which taxonomy cells in
\S\ref{sec:rewards}--\S\ref{sec:orchestration} are actually
reachable at industrial scale. This section discusses three such
constraints---rollout cost (\S\ref{sec:eng:rollout}), the harness
boundary (\S\ref{sec:eng:harness}), and trace-length dependence
(\S\ref{sec:eng:trace})---and relates each to a design choice that
recurs in the paper pool.

\subsection{Rollout cost dominates wall-clock training time}
\label{sec:eng:rollout}

A single-agent RL rollout for reasoning-level tasks is typically
$10^{2}$--$10^{3}$ tokens and one or two tool calls. A multi-agent
rollout at industrial scale is substantially more expensive. For a
back-of-envelope estimate in a centralized-orchestrator topology
(\S\ref{sec:systems}), let the orchestrator spawn $K$ sub-agents. The
$i$-th sub-agent consumes $L_i$ context/output tokens and issues
$T_i$ tool calls. Assuming per-token inference cost $c_{\text{tok}}$
and per-tool latency $c_{\text{tool}}$, the expected rollout cost is
\begin{equation}
  C_{\text{rollout}}(G) \;\approx\;
  \sum_{i=1}^{K} \bigl(L_i c_{\text{tok}} + T_i c_{\text{tool}}\bigr)
  \;+\; C_{\text{orch}}(K, |G|),
  \label{eq:rollout-cost}
\end{equation}
where $G$ is the orchestration trace, $|G|$ is its event count, and
$C_{\text{orch}}$ is the orchestrator's own inference and aggregation
cost. We write
\begin{equation}
  T_{\text{total}} \;=\; \sum_{i=1}^{K} T_i
  \label{eq:total-tool-calls}
\end{equation}
to avoid double-counting: if a public report gives total coordinated
steps or total tool calls, that number should be used as
$T_{\text{total}}$, not multiplied again by $K$. Under a simple
per-event proxy, substituting Kimi K2.6's reported operating point
($K{=}300$ and roughly $4{,}000$ coordinated steps / tool
calls)~\cite{kimi-k2-6-2026} yields a rollout that can be one to
several orders larger than a short single-agent reasoning rollout,
depending on token lengths and tool latencies. This schematic proxy
is before the RL-standard group-of-$G$ multiplier: GRPO-style
training with $G{=}8$ rollouts per prompt multiplies rollout
collection by a further $8\times$.

\begin{table}[t]
\centering
\small
\renewcommand{\arraystretch}{1.15}
\begin{tabularx}{\linewidth}{@{}p{2.3cm}p{1.6cm}p{2.3cm}p{1.7cm}X@{}}
\toprule
\textbf{Regime / entry} & \textbf{Team size} & \textbf{Trace length / calls} & \textbf{Source type} & \textbf{How used} \\
\midrule
Single-agent reasoning baseline
 & $K{=}1$ & $10$--$10^2$ reasoning/tool steps & modelling baseline
 & Normalization point for the cost schematic; not a corpus claim. \\
Academic LLM-MAS RL entries
 & small-to-moderate teams & usually $10^1$--$10^2$-scale traces in reported evaluations & arXiv / conference papers
 & Representative regime for retained methods; WideSeek-R1 and MARTI-MARS$^2$ add width/self-search scaling evidence but remain below the disclosed Kimi envelope. \\
Kimi K2.5 Agent Swarm
 & up to $100$ sub-agents & up to $1{,}500$ coordinated steps / tool calls as reported & company technical report
 & Published-training anchor for learned orchestration and scale-gap evidence. \\
Kimi K2.6 / Claw Groups
 & up to $300$ sub-agents & $4{,}000$ coordinated steps as reported & official product / technical blog
 & Deployment-scale evidence; not a fully reproducible training recipe. \\
Anthropic C-compiler case
 & $16$ parallel Claudes & roughly $2{,}000$ sessions in a long eng. project & case study
 & Workflow-shape and harness-pressure evidence, not RL-training evidence. \\
\bottomrule
\end{tabularx}
\caption{Scale and rollout evidence used for Figure~\ref{fig:rollout-cost}.
The table separates extracted public quantities from modelling assumptions;
aggregate academic ranges are representative of the retained method pool and
should not be read as a leaderboard.}
\label{tab:scale-extraction}
\end{table}

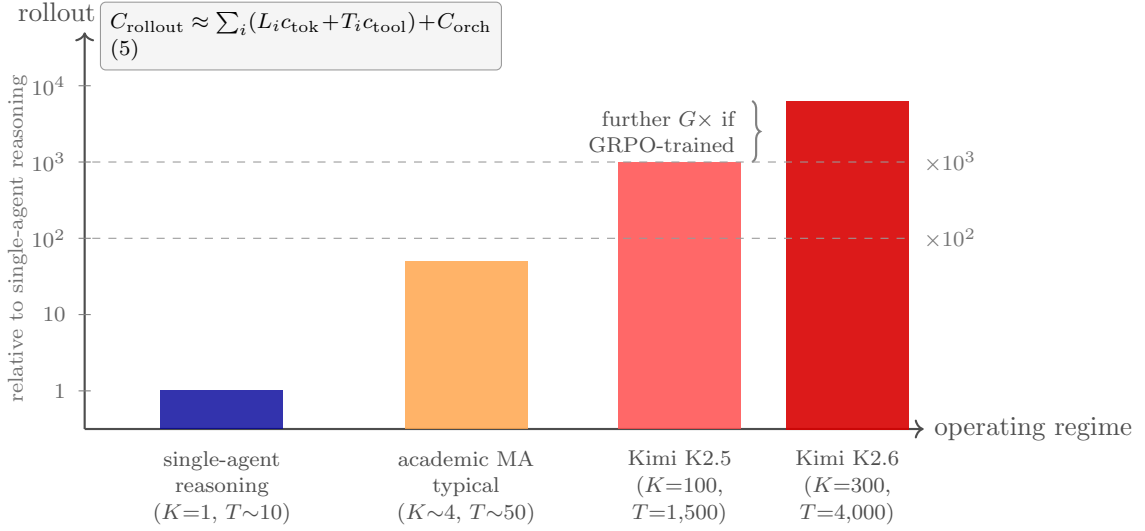
\begin{figure}[t]
\centering
\resizebox{0.95\linewidth}{!}{%
\begin{tikzpicture}[font=\footnotesize]
  \draw[->, thick, black!70] (0, 0) -- (11, 0)
    node[right]{\small operating regime};
  \draw[->, thick, black!70] (0, 0) -- (0, 5.2)
    node[above]{\small rollout cost};
  \foreach \y/\lab in {0.5/1, 1.5/10, 2.5/$10^2$, 3.5/$10^3$, 4.5/$10^4$} {
    \draw[black!50] (-0.08, \y) -- (0.08, \y);
    \node[left=1pt, black!60, font=\scriptsize] at (-0.08, \y) {\lab};
  }
  \node[left=8pt, black!55, font=\scriptsize, rotate=90, anchor=center]
    at (-0.6, 2.5) {relative to single-agent reasoning};

  \fill[blue!60!black, opacity=0.8]
    (1.0, 0) rectangle (2.6, 0.5);
  \node[anchor=north, black!70, align=center, font=\scriptsize]
    at (1.8, -0.15) {single-agent\\reasoning\\($K{=}1$, $T{\sim}10$)};

  \fill[orange!70, opacity=0.85]
    (4.2, 0) rectangle (5.8, 2.2);
  \node[anchor=north, black!70, align=center, font=\scriptsize]
    at (5.0, -0.15) {academic MA\\typical\\($K{\sim}4$, $T{\sim}50$)};

  \fill[red!70, opacity=0.85]
    (7.0, 0) rectangle (8.6, 3.5);
  \node[anchor=north, black!70, align=center, font=\scriptsize]
    at (7.8, -0.15) {Kimi K2.5\\($K{=}100$,\\$T{=}1{,}500$)};

  \fill[red!85!black, opacity=0.9]
    (9.2, 0) rectangle (10.8, 4.3);
  \node[anchor=north, black!70, align=center, font=\scriptsize]
    at (10.0, -0.15) {Kimi K2.6\\($K{=}300$,\\$T{=}4{,}000$)};

  \draw[decorate, decoration={brace, amplitude=3pt, mirror},
        thick, black!50]
    (8.75, 3.5) -- (8.75, 4.3)
    node[midway, left=5pt, black!70, font=\scriptsize, align=right,
         fill=white, fill opacity=0.9, text opacity=1, inner sep=1pt]
    {further $G{\times}$ if\\GRPO-trained};

  \draw[black!40, dashed, thin] (0.1, 2.5) -- (10.8, 2.5);
  \node[right=2pt, black!55, font=\scriptsize] at (10.8, 2.5) {$\times 10^2$};
  \draw[black!40, dashed, thin] (0.1, 3.5) -- (10.8, 3.5);
  \node[right=2pt, black!55, font=\scriptsize] at (10.8, 3.5) {$\times 10^3$};

  \node[draw=black!40, rounded corners=2pt,
        fill=gray!8, text width=5.0cm, font=\scriptsize,
        anchor=south west, align=left]
    at (0.2, 4.7)
    {$C_{\text{rollout}} \approx \sum_i(L_i c_{\text{tok}} + T_i c_{\text{tool}}) + C_{\text{orch}}$
      \hfill \eqref{eq:rollout-cost}};

\end{tikzpicture}%
}
\caption{Rollout cost across representative operating regimes,
shown as a schematic relative-cost proxy rather than a calibrated
dollar or latency estimate. The bars combine representative team size
and total trace length / tool-call counts under the cost form in
\eqref{eq:rollout-cost}; exact ratios depend on token lengths, tool
latencies, and harness overhead. The group-of-$G$ annotation shows the
additional rollout-collection multiplier imposed by GRPO-style
training. The visible gap between academic and industrial regimes is
what disciplines the engineering interventions surveyed in
\S\ref{sec:eng:rollout}: pipeline parallelism, execution--training
decoupling, and context folding all target the cost axis directly.}
\label{fig:rollout-cost}
\end{figure}

The consequences for algorithm design are blunt. \textbf{Methods
that require large $G$ are impractical at industrial rollout cost
without engineering interventions.} Three such interventions recur
in our pool. MarsRL~\cite{marsrl2025} uses \emph{agentic pipeline
parallelism}: different stages of different rollouts execute
concurrently, amortizing sub-agent idle time. Agent
Lightning~\cite{agent-lightning2025} introduces
\emph{execution--training decoupling}: an inference harness
produces rollouts asynchronously into a buffer consumed by the
trainer, removing wall-clock coupling between rollout completion
and gradient steps. Context-Folding~\cite{context-folding2025}
compresses sub-trajectories back into the main trace, reducing the
effective $L$ at aggregation time. WideSeek-R1~\cite{wideseek-r1-2026}
and MARTI-MARS$^2$~\cite{marti-mars2-2026} add a complementary
pressure: they explicitly scale parallel sub-agents or multi-agent
self-search, so their engineering bottleneck is not only depth but
also the width of the rollout graph.

\subsection{The harness boundary as a training-frozen interface}
\label{sec:eng:harness}

In production systems (OpenAI Codex, Anthropic Claude Code) the
\emph{harness}---the shell of model, tool registry, system prompt,
and execution runtime---is the actual unit of deployment
(\S\ref{sec:systems:codex}--\ref{sec:systems:claude-code}). The
model parameters $\theta$ are one component; the harness specifies
the interface through which $\theta$ is accessed and the set of
actions $\theta$ can issue. RL that targets a harness-hosted agent
must therefore obey a constraint that most academic methods do
not:

\begin{equation}
  \pi_{\theta}(\cdot \mid o) \;=\; \text{LLM}_{\theta}(\cdot \mid \mathrm{harness}(o)),
  \quad
  a \in \mathcal{A}_{\text{harness}},
  \label{eq:harness}
\end{equation}
where $\mathrm{harness}(\cdot)$ is a fixed-at-training-time map from
raw observation to prompt, and $\mathcal{A}_{\text{harness}}$ is
the (typically small, finite) set of tools / sub-agent-spawn verbs
the harness exposes. Equation~\eqref{eq:harness} says that the
harness defines both the input distribution the policy sees and
the output grammar it may emit; fine-tuning through a different
harness produces a different policy in the relevant operational
sense.

\begin{figure}[t]
\centering
\resizebox{0.95\linewidth}{!}{%
\begin{tikzpicture}[font=\footnotesize,
  core/.style={rectangle, rounded corners=2pt,
               draw=blue!65, thick, fill=blue!10,
               minimum width=2.6cm, minimum height=1.0cm,
               align=center},
  harnessbox/.style={rectangle, rounded corners=3pt,
                     draw=black!55, thick, dashed,
                     fill=gray!6, inner sep=3.5mm},
  envbox/.style={rectangle, rounded corners=2pt,
                 draw=green!50!black, thick, fill=green!10,
                 minimum width=2.2cm, minimum height=0.75cm,
                 align=center},
  toolbox/.style={rectangle, rounded corners=1pt,
                  draw=orange!75, thick, fill=orange!12,
                  minimum width=1.15cm, minimum height=0.55cm,
                  align=center, font=\scriptsize},
  arr/.style={-{Stealth[length=2mm]}, thick, black!65},
  trainable/.style={red!70!black, very thick, dashed},
  lbl/.style={font=\scriptsize\itshape, black!60}
]

\node[core] (llm) at (0, 0) {
  \textsf{LLM} $\pi_\theta$\\
  \textit{(trainable)}
};

\node[core, fill=white, draw=black!60,
      minimum width=2.0cm, minimum height=0.7cm,
      anchor=west] (prompt) at (3.8, 0.85)
      {prompt template};

\node[core, fill=white, draw=black!60,
      minimum width=2.0cm, minimum height=0.7cm,
      anchor=west] (registry) at (3.8, -0.05)
      {tool registry};

\node[core, fill=white, draw=black!60,
      minimum width=2.0cm, minimum height=0.7cm,
      anchor=west] (runtime) at (3.8, -0.95)
      {exec.\ runtime};

\node[harnessbox, fit={(prompt) (registry) (runtime)}] (harness) {};
\node[anchor=north east, font=\scriptsize\bfseries, black!65]
  at (harness.north east) {harness (frozen)};

\node[envbox, anchor=west] (env) at (9.8, 0.4) {user / task};

\node[toolbox, anchor=west] (t1) at (9.8, -0.5) {web API};
\node[toolbox, anchor=west] (t2) at (9.8, -1.2) {code exec};
\node[toolbox, anchor=west] (t3) at (9.8, -1.9) {MCP tool};

\draw[arr] (env.west) -- node[above, lbl]{obs} (harness.north east);
\draw[arr] (harness.west) -- node[above, lbl]{prompt}
           node[below, lbl]{$\mathrm{harness}(o)$} (llm.east);
\draw[arr] (llm.north east) to[bend left=10]
  node[above, lbl]{action} (prompt.west);
\draw[arr, dashed] (harness.south east) -- (t1.west);
\draw[arr, dashed] (harness.south east) -- (t2.west);
\draw[arr, dashed] (harness.south east) -- (t3.west);

\draw[trainable, rounded corners=2pt]
  ($(llm.north west)+(-0.15, 0.15)$) rectangle
  ($(llm.south east)+(0.15, -0.15)$);
\node[red!70!black, font=\scriptsize\bfseries, align=center,
      anchor=north] at (llm.south) {\\[2pt]RL gradient\\reaches here};

\node[black!60, font=\scriptsize\bfseries, anchor=south, align=center]
  at (harness.south) {\\[-4pt]not touched by RL\\(fixed at training time)};

\node[draw=black!40, rounded corners=2pt, fill=gray!5,
      text width=6.2cm, font=\scriptsize, align=left,
      anchor=north west]
  at (0, -2.6)
  {$\pi_\theta(\cdot \mid o) = \mathrm{LLM}_\theta(\cdot \mid \mathrm{harness}(o))$,
    \quad $a \in \mathcal{A}_{\text{harness}}$ \hfill
    \eqref{eq:harness}};

\end{tikzpicture}%
}
\caption{The harness as a training-frozen interface. The harness
(dashed box) wraps the trainable LLM $\pi_\theta$ with a prompt
template, tool registry, and execution runtime; only $\theta$
receives gradients during RL. The harness defines both the input
distribution $\mathrm{harness}(o)$ that the policy sees and the
output grammar $\mathcal{A}_{\text{harness}}$ it may emit. A policy
fine-tuned through a different harness is a different policy in
the deployment sense. In our pool only
Agent~Lightning~\cite{agent-lightning2025} formulates this boundary
explicitly as a contract the trainer must respect.}
\label{fig:harness-boundary}
\end{figure}
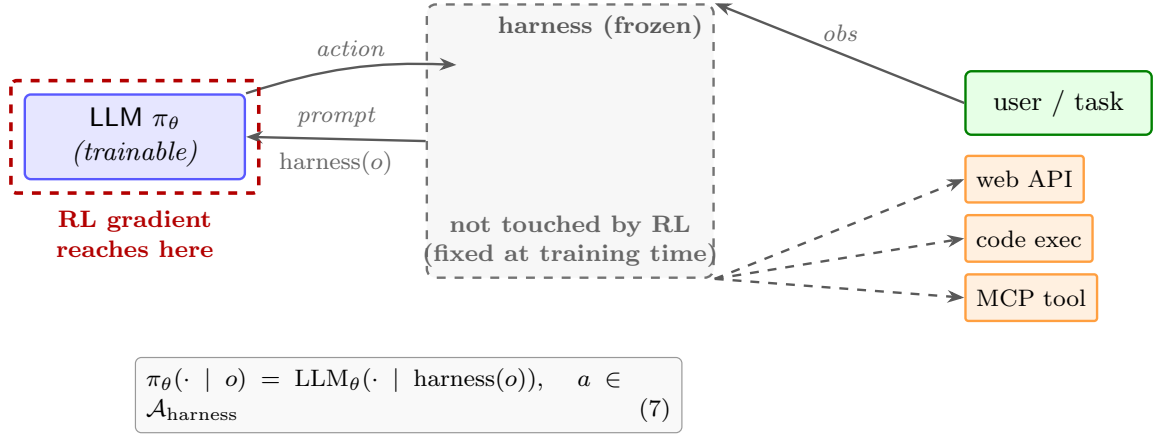

Only Agent~Lightning~\cite{agent-lightning2025} formulates the
harness boundary explicitly as an RL-training contract in our pool;
most methods train in bespoke environments and then deploy into a
different runtime. This is the most under-addressed gap between
academic RL and industrial deployment; we return to it in
\S\ref{sec:open}.

\subsection{Trace length and credit fragility}
\label{sec:eng:trace}

Section~\ref{sec:credit:fails} observed qualitatively, and
\S\ref{sec:formalism:claims} argued via
Observation~\ref{claim:diffusion}, that the effective per-decision
signal under shared-reward uniform credit decreases with trace
length. This section records the same effect from an engineering
standpoint.

\textbf{Observation (credit fragility with trace length).}
Consider a shared-reward orchestration trace of length $T$ decisions
with terminal reward $R \in \{0,1\}$ under uniform credit assignment
($A_t = R - \bar R$ at every step). The per-decision advantage is
bounded by the spread of $R$ about its mean, while the expected
per-decision marginal contribution shrinks as the reward is smeared
across more and more decisions. Consequently, under these assumptions,
the following heuristic \emph{signal-to-noise proxy} at any single
decision can become harder to estimate as the trace length grows:
\begin{equation}
  \mathrm{SNR}(t)
  \;\equiv\; \frac{\big|\mathbb{E}[A_t \mid s_t, a_t]\big|}{\sqrt{\mathrm{Var}[A_t]}}
  \quad\text{is a fragile proxy as $T$ increases.}
  \label{eq:diffusion-bound}
\end{equation}
We deliberately avoid a precise rate in
Equation~\eqref{eq:diffusion-bound}: the exact dependence on $T$ is
determined by the task-specific noise structure of $R$ and the
baseline estimator, and giving a closed form would require
assumptions beyond the scope of this survey
(\S\ref{sec:formalism:claims}). The qualitative failure mode
suffices to explain the empirical failure mode:
Dr.\,MAS~\cite{dr-mas2026} documents that na\"ive GRPO becomes
unstable at multi-agent scale under exactly this regime. Methods
that target role-level (MALT~\cite{malt2025}), message-level
(C3~\cite{c3-2026}), or orchestrator-level
(Puppeteer~\cite{puppeteer2025}) credit effectively shrink the
relevant $T$ for their decomposed sub-problem---which is the
reward--credit dual of \S\ref{sec:formalism:dual} viewed from the
engineering side.

\begin{figure}[t]
\centering
\resizebox{0.95\linewidth}{!}{%
\begin{tikzpicture}[font=\footnotesize]

  \draw[->, thick, black!70] (0, 0) -- (11.2, 0)
    node[right]{\small trace length $T$ (log-scale)};
  \draw[->, thick, black!70] (0, 0) -- (0, 5.0)
    node[above, align=center]{\small per-step\\signal / noise};

  \foreach \x/\lab in {1.5/10, 3.75/$10^2$, 6.0/$10^3$, 8.25/$10^4$, 10.5/$10^5$} {
    \draw[black!50] (\x, -0.08) -- (\x, 0.08);
    \node[below=2pt, black!60, font=\scriptsize] at (\x, -0.1) {\lab};
  }

  \fill[orange!10]
    (1.0, 0.02) rectangle (4.5, 4.8);
  \node[orange!85!black, font=\scriptsize\bfseries, align=center]
    at (2.75, 4.5) {academic\\benchmarks};

  \fill[red!10]
    (5.7, 0.02) rectangle (9.5, 4.8);
  \node[red!85!black, font=\scriptsize\bfseries, align=center]
    at (7.6, 4.5) {industrial\\deployment};

  \draw[blue!65!black, very thick]
    plot[smooth, samples=80, domain=1:10.8]
    (\x, {4.2 / (1 + 0.8*(\x-0.5))});
  \node[blue!65!black, font=\footnotesize, anchor=west]
    at (9.3, 0.85) {uniform credit};
  \node[blue!65!black, font=\scriptsize, anchor=west]
    at (9.3, 0.45) {schematic};

  \draw[green!50!black, very thick, dash pattern=on 4pt off 2pt]
    plot[smooth, samples=80, domain=1:10.8]
    (\x, {3.0 / (1 + 0.25*(\x-0.5))});
  \node[green!50!black, font=\footnotesize, anchor=west]
    at (9.3, 1.95) {role / message};
  \node[green!50!black, font=\scriptsize, anchor=west]
    at (9.3, 1.55) {credit decomp.};

  \draw[red!65!black, very thick, dash pattern=on 2pt off 2pt]
    plot[smooth, samples=80, domain=1:10.8]
    (\x, {2.7 - 0.08*(\x-0.5)});
  \node[red!65!black, font=\footnotesize, anchor=west]
    at (9.3, 2.9) {orch.\ critic};
  \node[red!65!black, font=\scriptsize, anchor=west]
    at (9.3, 2.5) {(Puppeteer)};

  \draw[black!50, dashed] (0.1, 1.0) -- (10.8, 1.0);
  \node[black!55, font=\scriptsize, anchor=west]
    at (0.15, 1.15) {approx.\ training-instability threshold};

  \draw[<->, black!55, thick]
    (1.0, -0.8) -- node[below=-1pt, midway, black!60, font=\scriptsize]
    {academic} (4.5, -0.8);
  \draw[<->, black!55, thick]
    (5.7, -0.8) -- node[below=-1pt, midway, black!60, font=\scriptsize]
    {industrial} (9.5, -0.8);

  \node[draw=black!40, rounded corners=2pt, fill=gray!5,
        text width=5.7cm, font=\scriptsize, anchor=north west,
        align=left]
    at (0.15, -1.2)
    {Uniform-credit warning:
      per-step signal can become low-SNR as $T$ grows
      \hfill\eqref{eq:diffusion-bound}};

\end{tikzpicture}%
}
\caption{Schematic per-step signal-to-noise under three credit
schemes as trace length $T$ grows. The blue curve is not a proven
rate; it visualizes the qualitative warning in
\eqref{eq:diffusion-bound}: uniform terminal credit can become
low-SNR on long shared-reward traces. Role- or message-level
decomposition (dashed green) partitions the trace into shorter
sub-problems; a learned orchestrator critic (dotted red) targets a
smaller set of orchestrator decisions. Curves and thresholds are
illustrative, not fitted empirical laws.}
\label{fig:trace-length}
\end{figure}
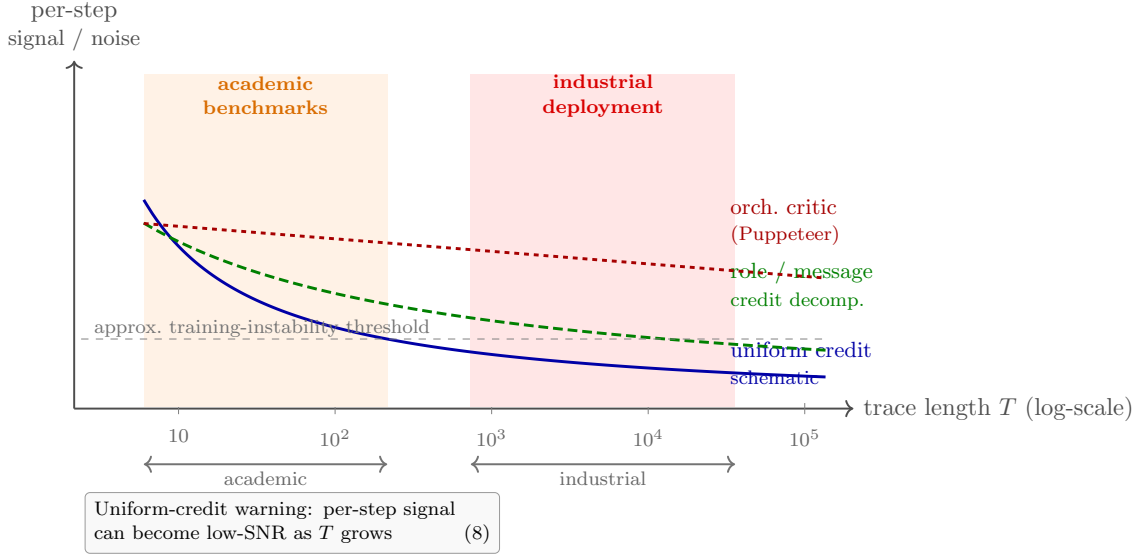

The practical implication: \textbf{the right value of $T$ at which
to benchmark an LLM-MAS RL method is the $T$ at which the method
will be deployed.} Since academic pool entries are mostly trained at
$T \lesssim 10^2$ while the Kimi-reported deployment envelope reaches
$T \sim 10^3$--$10^4$ and other public industrial systems mainly
expose harness pressure rather than comparable training-scale traces,
current academic results may systematically
\emph{overestimate} credit-assignment effectiveness at deployment
scale---the opposite of the usual generalization story.

\takeaway{Three engineering constraints discipline LLM-MAS RL in
ways classical MARL does not: rollout cost scales as
\(\sum_i(L_i c_{\text{tok}} + T_i c_{\text{tool}}) +
C_{\text{orch}}(K, |G|)\) and makes large-$G$
algorithms infeasible at industrial scale; the harness is a
training-frozen interface that most academic methods ignore; and
per-decision signal-to-noise under uniform credit decreases with
trace length, with the decrease most relevant at the long-horizon
$T$ publicly reported by Kimi. Each constraint points to
a specific open problem (\S\ref{sec:open}).}

\section{Reward Design for LLM-based MAS}
\label{sec:rewards}

Reward design is the first choice any LLM-MAS RL practitioner must make,
and it is upstream of credit assignment: what cannot be measured as a
reward cannot be assigned as credit. This section surveys the design
space along eight families and grounds each in representative
entries from the paper pool.

\subsection{Eight families of rewards}
\label{sec:rewards:families}

Table~\ref{tab:reward-families} organizes the design space along five
axes: what signal the reward captures, at what granularity it is
emitted, where the signal comes from, its dominant hacking risk, and
representative entries. Rows follow the order in which a practitioner
typically encounters them: start from a shared outcome (R1), then
decompose per-agent (R2--R3), then densify with process signals
(R4--R6), then add system-level incentives (R7), then combine (R8).

\begin{table}[htbp]
\centering
\small
\renewcommand{\arraystretch}{1.25}
\begin{tabularx}{\linewidth}{@{}c X X X X X@{}}
\toprule
\textbf{ID} & \textbf{Family} & \textbf{Granularity} & \textbf{Source} & \textbf{Dominant hacking risk} & \textbf{Representative methods} \\
\midrule
R1 & Shared team / outcome
   & team (terminal)
   & verifier / ground truth
   & reward diffusion; free-riding
   & MAGRPO~\cite{magrpo2025}, MAPoRL~\cite{maporl2025}, Dr.\,MAS~\cite{dr-mas2026}, CoLLM-MAAC~\cite{collm-maac2026} \\
R2 & Individual agent
   & per-agent (terminal)
   & per-agent outcome
   & credit overfits solvable sub-tasks; lazy-agent
   & MARFT~\cite{marft2025}, Context-Folding~\cite{context-folding2025} \\
R3 & Role-specific
   & per-role (terminal or per-turn)
   & role-specific rubric
   & rubric mismatch across roles
   & MALT~\cite{malt2025}, MATPO~\cite{matpo2025}, LAMO~\cite{lamo2026}, DEPART~\cite{depart2026} \\
R4 & Process (PRM)
   & per-step / per-turn
   & trained PRM or heuristic
   & step-padding; PRM gaming
   & MALT role-PRM~\cite{malt2025}, MarsRL~\cite{marsrl2025} \\
R5 & Tool-use
   & per-tool-call
   & tool execution signal
   & tool-spam; fabricated tool success
   & MATPO~\cite{matpo2025}, Agent~Lightning~\cite{agent-lightning2025} \\
R6 & Debate / verifier
   & per-message / per-turn
   & LLM judge or debate resolution
   & verifier collusion; over-communication
   & Debate-as-Reward~\cite{debate-as-reward2026}, MAE~\cite{mae2025},
     MAGIC~\cite{magic2026}, CriticLean~\cite{criticlean2025} \\
R7 & Orchestration
   & per-orchestrator-decision
   & system metrics (speedup, finish-rate)
   & pseudo-parallelism; reward-shape collapse
   & Kimi PARL~\cite{kimi-k2-5-2026}, Puppeteer~\cite{puppeteer2025}, ParaManager~\cite{paramanager2026}, WideSeek-R1~\cite{wideseek-r1-2026} \\
R8 & Hybrid local--global
   & mixed
   & weighted composition of R1--R7
   & weight drift; signal drowning
   & SHARP~\cite{sharp2026}, M-GRPO~\cite{m-grpo2025}, HERA~\cite{hera2026}, LangMARL~\cite{langmarl2026}, Agent Q-Mix~\cite{agent-qmix2026} \\
\bottomrule
\end{tabularx}
\caption{Eight reward families for LLM-MAS RL. Each row names a
distinct signal a practitioner can attach to a multi-agent rollout;
rows are not mutually exclusive and are commonly combined through R8.}
\label{tab:reward-families}
\end{table}

\noindent Three observations follow from the table.

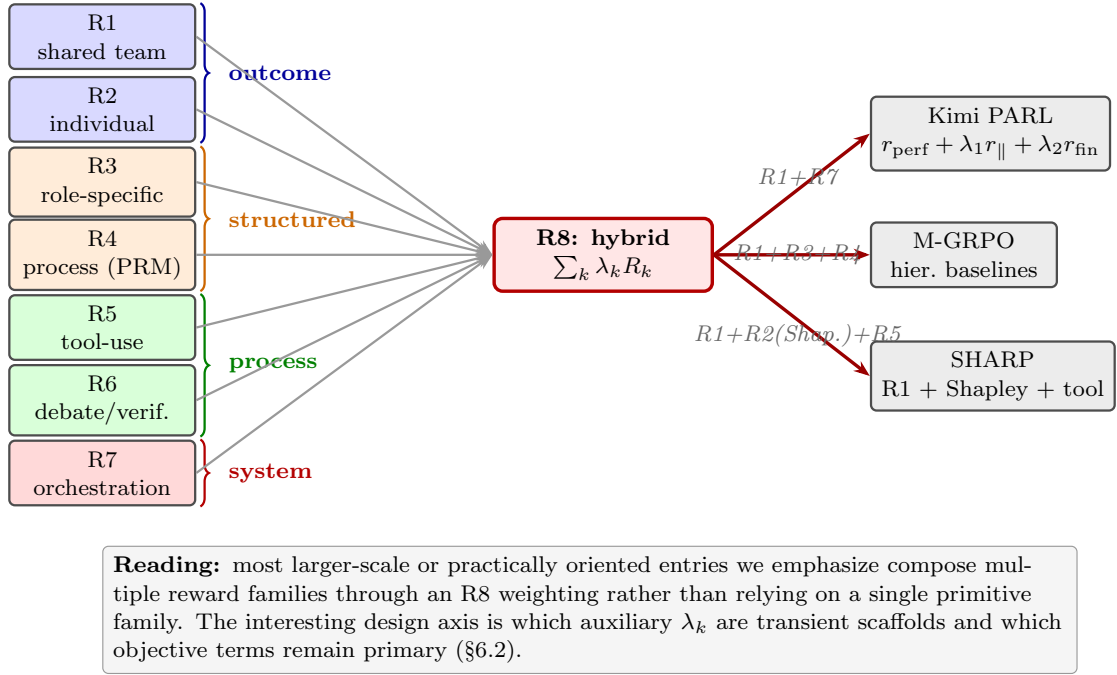
\begin{figure}[t]
\centering
\resizebox{0.92\linewidth}{!}{%
\begin{tikzpicture}[font=\footnotesize,
  rfam/.style={rectangle, rounded corners=2pt,
               draw=black!70, thick, fill=#1!15,
               minimum width=2.3cm, minimum height=0.7cm,
               align=center, font=\scriptsize},
  hybrid/.style={rectangle, rounded corners=3pt,
                 draw=red!70!black, very thick, fill=red!10,
                 minimum width=2.7cm, minimum height=0.9cm,
                 align=center, font=\scriptsize\bfseries},
  arr/.style={-{Stealth[length=2mm]}, thick, black!55},
  annot/.style={font=\scriptsize\itshape, black!60}
]

  \node[rfam=blue]    (r1) at (0, 3.2)  {R1\\shared team};
  \node[rfam=blue]    (r2) at (0, 2.3)  {R2\\individual};
  \node[rfam=orange]  (r3) at (0, 1.4)  {R3\\role-specific};
  \node[rfam=orange]  (r4) at (0, 0.5)  {R4\\process (PRM)};
  \node[rfam=green]   (r5) at (0, -0.4) {R5\\tool-use};
  \node[rfam=green]   (r6) at (0, -1.3) {R6\\debate/verif.};
  \node[rfam=red]     (r7) at (0, -2.2) {R7\\orchestration};

  \draw[decorate, decoration={brace, amplitude=3pt}, thick, blue!60!black]
    ($(r1.north east)+(0.05, 0)$) -- ($(r2.south east)+(0.05, 0)$)
    node[midway, right=6pt, blue!60!black, font=\scriptsize]
    {\textbf{outcome}};
  \draw[decorate, decoration={brace, amplitude=3pt}, thick, orange!80!black]
    ($(r3.north east)+(0.05, 0)$) -- ($(r4.south east)+(0.05, 0)$)
    node[midway, right=6pt, orange!80!black, font=\scriptsize]
    {\textbf{structured}};
  \draw[decorate, decoration={brace, amplitude=3pt}, thick, green!50!black]
    ($(r5.north east)+(0.05, 0)$) -- ($(r6.south east)+(0.05, 0)$)
    node[midway, right=6pt, green!50!black, font=\scriptsize]
    {\textbf{process}};
  \draw[decorate, decoration={brace, amplitude=3pt}, thick, red!70!black]
    ($(r7.north east)+(0.05, 0)$) -- ($(r7.south east)+(0.05, 0)$)
    node[midway, right=6pt, red!70!black, font=\scriptsize]
    {\textbf{system}};

  \node[hybrid] (r8) at (6.2, 0.5) {R8: hybrid\\$\sum_k \lambda_k R_k$};

  \foreach \src in {r1, r2, r3, r4, r5, r6, r7}
    \draw[arr, black!40] (\src.east) -- (r8.west);

  \node[rfam=gray, anchor=west] (p1) at (9.5, 2.0)
    {Kimi PARL\\$r_{\text{perf}} + \lambda_1 r_{\parallel} + \lambda_2 r_{\text{fin}}$};
  \node[rfam=gray, anchor=west] (p2) at (9.5, 0.5)
    {M-GRPO\\hier.\ baselines};
  \node[rfam=gray, anchor=west] (p3) at (9.5, -1.0)
    {SHARP\\R1 + Shapley + tool};

  \draw[arr, red!60!black, very thick] (r8.east) -- (p1.west);
  \draw[arr, red!60!black, very thick] (r8.east) -- (p2.west);
  \draw[arr, red!60!black, very thick] (r8.east) -- (p3.west);

  \node[annot] at (8.6, 1.45)  {R1+R7};
  \node[annot] at (8.6, 0.5)   {R1+R3+R4};
  \node[annot] at (8.6, -0.5)  {R1+R2(Shap.)+R5};

  \node[draw=black!40, rounded corners=2pt, fill=gray!8,
        text width=11.8cm, font=\scriptsize, align=left,
        anchor=north west]
    at (0, -3.1)
    {\textbf{Reading:} most larger-scale or practically oriented entries
      we emphasize compose multiple reward families through an R8
      weighting rather than relying on a single primitive family. The
      interesting design axis is which auxiliary $\lambda_k$ are
      transient scaffolds and which objective terms remain primary
      (\S\ref{sec:rewards:parl}).};

\end{tikzpicture}%
}
\caption{Reward family composition. The seven primitive families
R1--R7 (\S\ref{sec:rewards:families}) group into four semantic
tiers---outcome, structured, process, system---and are composed
through an R8 hybrid weighting to produce method-specific reward
shapes. Three representative compositions from our pool are shown
on the right. The less-studied axis
is schedule semantics: which terms are transient scaffolds, which
terms define the primary objective, and which schedules are disclosed.}
\label{fig:reward-composition}
\end{figure}

\begin{itemize}
  \item \textbf{Terminal $\to$ process is a densification axis.}
    R1--R3 emit one number at episode end; R4--R6 emit signals
    throughout the trace. The latter give stronger gradients but
    introduce new attack surfaces (PRM gaming, judge collusion).
  \item \textbf{R7 is newly central in LLM-MAS.}
    Orchestration rewards have single-agent analogues in compute,
    tool-cost, and process shaping, but no close analogue for
    spawn / delegate / aggregate decisions over multiple agent
    instances. They reward
    \emph{system-level} properties (wall-clock speedup, split
    correctness, finish-rate), not task-level correctness. This is
    where LLM-MAS RL departs most sharply from agentic RL.
  \item \textbf{R8 is the default in practice.}
    The larger-scale or practically oriented entries we emphasize
    (Kimi PARL, M-GRPO, Context-Folding, SHARP, LangMARL,
    Agent Q-Mix, MARSHAL~\cite{marshal2026}, and
    DEPART~\cite{depart2026}) use an R8
    composition rather than a single
    family. Figure~\ref{fig:reward-composition} makes the composition
    pattern explicit. The open question is not \emph{which} family
    to pick but \emph{how to weight} them without one drowning the
    others---a point we return to in \S\ref{sec:open}.
\end{itemize}

\subsection{The Kimi PARL reward decomposition (worked example)}
\label{sec:rewards:parl}

\begin{figure}[t]
\centering
\begin{tikzpicture}[font=\footnotesize]

  \draw[->, thick, black!70] (0,0) -- (8.4, 0) node[right]{\small training step $\to$};
  \draw[->, thick, black!70] (0,0) -- (0, 4.0) node[above]{\small reward weight};

  \foreach \y/\lab in {0/0, 1.5/0.5, 3.0/1.0} {
    \draw[black!50] (-0.08, \y) -- (0.08, \y);
    \node[left=2pt, black!60, font=\scriptsize] at (-0.08, \y) {\lab};
  }

  \foreach \x/\lab in {2.0/early, 4.5/mid, 7.2/late} {
    \draw[black!40, dashed] (\x, 0) -- (\x, 3.5);
    \node[below=2pt, black!60, font=\scriptsize\itshape] at (\x, 0) {\lab};
  }

  \draw[blue!70!black, very thick]
    (0.1, 3.0) -- (8.0, 3.0);
  \node[blue!70!black, font=\footnotesize] at (8.4, 3.0)
    {\hspace{-0.3cm}\small $r_{\text{perf}}$};

  \draw[orange!90!black, very thick]
    plot[smooth, tension=0.6] coordinates {
      (0.1, 0.4) (1.4, 1.4) (2.8, 2.2) (4.0, 2.0) (5.5, 1.0) (7.0, 0.25) (8.0, 0.15)
    };
  \node[orange!90!black, font=\footnotesize, anchor=west,
        fill=white, fill opacity=0.9, text opacity=1, inner sep=1pt]
    at (6.65, 0.95)
    {\small $\lambda_1 r_{\text{parallel}}$};

  \draw[red!75!black, very thick, dashed]
    plot[smooth, tension=0.6] coordinates {
      (0.1, 2.5) (1.0, 2.6) (2.0, 2.4) (3.5, 1.6) (5.0, 0.7) (6.5, 0.2) (8.0, 0.18)
    };
  \node[red!75!black, font=\footnotesize, anchor=west,
        fill=white, fill opacity=0.9, text opacity=1, inner sep=1pt]
    at (7.35, 0.52)
    {\small $\lambda_2 r_{\text{finish}}$};

  \node[align=center, black!70, font=\scriptsize, anchor=north]
    at (1.0, -0.6) {scaffold\\ active};
  \draw[black!50, ->, >=stealth, thin] (1.0, -0.32) -- (1.0, -0.05);

  \node[align=center, black!70, font=\scriptsize, anchor=north]
    at (7.0, -0.6) {scaffold\\ removed};
  \draw[black!50, ->, >=stealth, thin] (7.0, -0.32) -- (7.0, -0.05);

  \node[draw=red!50, fill=red!8, rounded corners=2pt,
        text width=4.6cm, font=\scriptsize, align=left,
        anchor=north west] at (4.2, -1.1)
    {\textbf{Key shape:} both auxiliary terms are reported as
      \emph{transient scaffolds}: useful for learning parallel
      orchestration early, removed so the final policy optimizes
      the primary objective.};

\end{tikzpicture}
\caption{Schematic of Kimi PARL's three-term reward
$r_{\text{orch}} = r_{\text{perf}} + \lambda_1 r_{\text{parallel}}
+ \lambda_2 r_{\text{finish}}$ across training
(\S\ref{sec:rewards:parl}). The task-outcome term $r_{\text{perf}}$
is the primary objective; both auxiliary orchestration-shaping terms
are shown as transient scaffolds because the public Kimi K2.5 report
states that their hyperparameters are annealed to zero over training.
Curves are schematic; exact schedules are not disclosed in the public
PARL report.}
\label{fig:parl-annealing}
\end{figure}
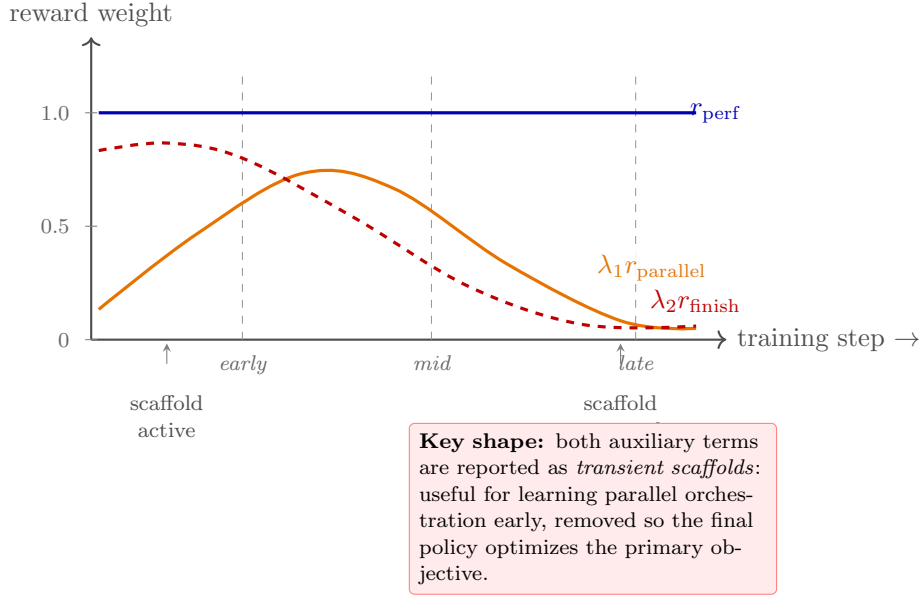

Kimi's PARL~\cite{kimi-k2-5-2026} is the clearest published instance
of an R7+R8 composition and serves as our canonical worked example.
Following the evidence convention in Table~\ref{tab:evidence-status},
we use Kimi here as a company-report anchor: the public material
discloses the reward components and deployment scale, but not enough
optimizer, data, and ablation detail to make PARL independently
reproducible.
The orchestrator reward takes the form
\begin{equation}
  r_{\text{orch}} \;=\; r_{\text{perf}} \;+\; \lambda_1\,r_{\text{parallel}} \;+\; \lambda_2\,r_{\text{finish}},
  \label{eq:parl}
\end{equation}
where $r_{\text{perf}}$ is the downstream task outcome (R1),
$r_{\text{parallel}}$ rewards genuine speedup over a serial baseline
(R7), and $r_{\text{finish}}$ rewards all spawned sub-agents reaching
termination (R7, a shape against pseudo-parallelism). Crucially, the
public Kimi K2.5 description states that the hyperparameters for both
auxiliary rewards are annealed to zero over training so that the final
policy optimizes the primary task objective. Early in training these
terms scaffold exploration of parallel scheduling; late in training
they are removed so the orchestrator cannot farm auxiliary metrics by
over-spawning or padding parallel work. Figure~\ref{fig:parl-annealing}
illustrates the schematic shape.
This staged annealing is, to our knowledge, the clearest explicit
acknowledgement \emph{within our curated pool} that R7 rewards are
inherently \emph{transient scaffolds}---useful to escape the
zero-gradient region where the orchestrator has not yet learned to
spawn, but harmful at convergence.

\subsection{Reward-hacking failure modes specific to MAS}
\label{sec:rewards:hacking}

Five failure modes recur across the pool. Each maps to one or more
reward families, and each has been reported (or, in at least two
cases, directly measured) in published work.

\begin{itemize}
  \item \textbf{Pseudo-parallelism} (R7). The orchestrator spawns
    sub-agents that do no useful work in order to maximize a na\"ive
    parallelism bonus. Mitigated in Kimi PARL by the
    $r_{\text{finish}}$ shape and by a Critical-Steps metric that
    distinguishes real parallel progress from padded traces.
  \item \textbf{Free-riding / lazy agent} (R1). Under shared reward,
    one sub-agent contributes negligibly but absorbs equal credit.
    This is the direct LLM-MAS analogue of the lazy-agent problem in
    classical MARL. SHARP~\cite{sharp2026} targets this via Shapley
    marginal credit; Dr.\,MAS~\cite{dr-mas2026} targets the related
    gradient pathology via agent-wise normalization.
  \item \textbf{Communication padding} (R6). When a judge or PRM
    scores messages, policies inflate message length or verbosity to
    farm partial credit. Observed in debate-style setups and
    implicated in Debate-as-Reward's~\cite{debate-as-reward2026}
    design of resolution-based (not length-based) rewards.
  \item \textbf{Tool-spam} (R5). When tool-call success is rewarded,
    policies call many redundant tools. MATPO~\cite{matpo2025} and
    Agent~Lightning~\cite{agent-lightning2025} handle this by
    conditioning tool reward on downstream task outcome rather than
    call-level success alone.
  \item \textbf{Verifier collusion} (R6). When the verifier is an LLM
    from the same family as the policy, both drift together and the
    verifier reward becomes uninformative. Mitigations in the pool
    are mostly diagnostic rather than algorithmic---dedicated critic
    benchmarks~\cite{codecriticbench2025,artifactsbench2025} surface
    when judges and policies have drifted together but do not by
    themselves prevent collusion. CriticLean~\cite{criticlean2025} is
    one of the few methods to train a critic explicitly via RL on a
    formal verification signal (Lean 4 type-checking), grounding the
    critic in something other than another LLM's preference; this
    remains open more broadly (\S\ref{sec:open}).
\end{itemize}

\subsection{Discussion: the unanswered weighting question}
\label{sec:rewards:discussion}

Three threads run through the families above and converge on a
single question we cannot yet answer.

\textbf{Densification appears to trade signal-to-noise for attack surface.}
Moving along the R1$\to$R4$\to$R6 axis adds gradient signal at
every step, but each new family is a new \emph{hackable
sub-problem}. A reward composed of $r_{\text{perf}} + r_{\text{PRM}} + r_{\text{tool}} + r_{\text{judge}}$
gives the policy four levers it can pull instead of one, and only
$r_{\text{perf}}$ is anchored in ground truth. The empirical
pattern in our pool is suggestive rather than conclusive: methods with rich shaping
(MALT~\cite{malt2025}, SHARP~\cite{sharp2026}) typically report
smaller gains over their own baselines than methods with sparse
shaping (Dr.\,MAS~\cite{dr-mas2026}, Puppeteer~\cite{puppeteer2025})
report over theirs. One possible explanation is that richer shaping
creates more opportunities for PRM or judge gaming, but direct
comparison is impossible (\S\ref{sec:bench:gap}); we therefore treat
this as a hypothesis generated by the survey rather than as an
established empirical law.

\textbf{Interaction-derived reward is becoming a fourth route to
self-evolution.} The coverage audit added several entries that do
not fit cleanly into the older outcome/process/tool/verifier split.
CoMAS~\cite{comas2026} constructs rewards from inter-agent discussion
dynamics; SiriuS~\cite{sirius2025} and Multiagent
Finetuning~\cite{multiagent-finetuning2025} turn successful or
majority-supported interaction traces into reusable experience or
fine-tuning data; MAS-Zero~\cite{mas-zero2025} uses meta-level
feedback to refine MAS designs without outcome supervision. We tag
these as R4/R6/R8-adjacent rather than as a new primitive family,
because the reward signal is still mediated by process, verifier, or
aggregation mechanisms. They nevertheless mark a distinct trend:
the team interaction itself is increasingly used to manufacture the
training signal.

\textbf{R7 weights are not constants---and we have no theory of
their schedule.} Section~\ref{sec:rewards:parl} described
$\lambda_2$ in Kimi PARL as ``annealed toward zero,'' which is
correct but uninformative: when, by how much, on what schedule?
Any orchestration-reward shape that successfully scaffolds
spawning early is, by construction, also a shape the orchestrator
will exploit at convergence---a multi-agent analogue of the
potential-based shaping pathology in single-agent RL. A principled
R7 schedule---ideally derived from a measurable convergence
indicator rather than from training-step count---is, to our
knowledge, absent from our pool.

\textbf{The composition is left to the practitioner.} Every R8
weighting in our pool is hand-tuned. We found no retained entry
with an auto-balancing mechanism that adjusts $\lambda_k$ during
training based on observed gradient magnitudes or reward-component
variance. This is a clear and tractable target for follow-up work,
particularly given that classical RL has well-understood
gradient-balancing heuristics (PCGrad-style methods, GradNorm)
that have not yet been adapted to the MAS-specific setting of
agent-shared losses with per-component noise structures.

\takeaway{Reward design in LLM-MAS is not a choice among eight families but a
weighting over R8 compositions of them. The family that becomes newly
central relative to single-agent LLM RL is R7---orchestration
reward---and its defining feature in the clearest public industrial
example is that auxiliary orchestration-shaping weights are
\emph{not constant}: in Kimi K2.5, they scaffold early training and
are reported as annealed to zero as training progresses. The composition
weighting itself is currently hand-tuned in the entries we reviewed;
auto-balancing R8 weights against measurable training
diagnostics is a clear near-term target.
}

\section{Credit Assignment in LLM-based MAS}
\label{sec:credit}

Credit assignment is where LLM-MAS departs most sharply from both
single-agent LLM RL and classical MARL. Single-agent RL must propagate
credit backwards in time (token $\to$ step $\to$ trajectory). Classical
MARL adds a spatial dimension (which agent). LLM-MAS adds a
\emph{structural} dimension: credit must flow through roles,
messages, tool calls, and---uniquely---through the orchestrator's
decisions about \emph{whether and how to spawn agents in the first
place}. This section makes that structural dimension explicit and
organizes the paper pool around it.

\subsection{The credit- and signal-bearing unit hierarchy}
\label{sec:credit:hierarchy}

We argue that a single final reward in an LLM-MAS trace has \emph{eight}
plausible units to which reward, credit, or design signals can attach,
each one finer-grained than the next:

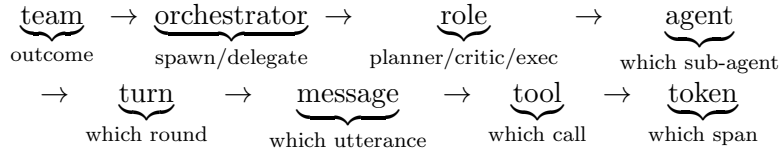
\[
\begin{array}{c}
\underbrace{\text{team}}_{\text{outcome}}
\;\to\;
\underbrace{\text{orchestrator}}_{\text{spawn/delegate}}
\;\to\;
\underbrace{\text{role}}_{\text{planner/critic/exec}}
\;\to\;
\underbrace{\text{agent}}_{\text{which sub-agent}}
\\[6pt]
\;\to\;
\underbrace{\text{turn}}_{\text{which round}}
\;\to\;
\underbrace{\text{message}}_{\text{which utterance}}
\;\to\;
\underbrace{\text{tool}}_{\text{which call}}
\;\to\;
\underbrace{\text{token}}_{\text{which span}}
\end{array}
\]

\begin{figure}[t]
\centering
\begin{tikzpicture}[
    font=\footnotesize,
    level/.style={
      rectangle, rounded corners=2pt, draw=black!70, thick,
      minimum width=3.4cm, minimum height=0.68cm,
      align=center, fill=#1!15
    },
    novel/.style={draw=red!70, line width=0.7pt},
    prop/.style={-{Stealth[length=2.2mm]}, thick, black!60},
    coverage/.style={font=\scriptsize, align=left, black!70}
]

\node[level=red]            (team)  at (0,0)     {\textbf{team}};
\node[level=red,  novel, below=0.35cm of team]  (orch)  {\textbf{orchestrator}};
\node[level=orange,novel, below=0.35cm of orch] (role)  {\textbf{role}};
\node[level=orange, below=0.35cm of role]       (agent) {\textbf{agent}};
\node[level=yellow,below=0.35cm of agent]       (turn)  {\textbf{turn}};
\node[level=green, novel, below=0.35cm of turn] (msg)   {\textbf{message}};
\node[level=green, below=0.35cm of msg]         (tool)  {\textbf{tool call}};
\node[level=blue,  below=0.35cm of tool]        (token) {\textbf{token}};

\foreach \a/\b in {team/orch, orch/role, role/agent, agent/turn, turn/msg, msg/tool, tool/token}
  \draw[prop] (\a) -- (\b);

\node[coverage, right=0.4cm of team]  {shared outcome baseline};
\node[coverage, right=0.4cm of orch]  {Puppeteer, Kimi PARL, WideSeek \textit{(sparse)}};
\node[coverage, right=0.4cm of role]  {MALT, M-GRPO, MATPO, DEPART, LAMO};
\node[coverage, right=0.4cm of agent] {MAGRPO, Dr.\,MAS, SHARP, MAPoRL, LangMARL};
\node[coverage, right=0.4cm of turn]  {MarsRL, Context-Folding, MARSHAL};
\node[coverage, right=0.4cm of msg]   {C3 \textit{(sparse)}};
\node[coverage, right=0.4cm of tool]  {MATPO, SHARP, Agent Lightning};
\node[coverage, right=0.4cm of token] {standard GAE (single-agent inherit)};

\draw[decorate, decoration={brace, amplitude=4pt, mirror}, thick, red!70]
  ($(team.north west)+(-0.25,0.08)$) --
  ($(role.south west)+(-0.25,-0.08)$)
  node[midway, left=7pt, align=center, red!70, font=\scriptsize\bfseries]
  {new for\\LLM-MAS};

\draw[decorate, decoration={brace, amplitude=4pt, mirror}, thick, black!50]
  ($(agent.north west)+(-0.25,0.08)$) --
  ($(token.south west)+(-0.25,-0.08)$)
  node[midway, left=7pt, align=center, black!60, font=\scriptsize]
  {inherited from\\MARL / single-agent};

\end{tikzpicture}
\caption{The eight credit-bearing units in LLM-MAS RL
(\S\ref{sec:credit:hierarchy}), stacked from coarsest (\textbf{team})
to finest (\textbf{token}). Red-outlined levels---\textbf{orchestrator},
\textbf{role}, \textbf{message}---have no clean counterpart in
classical MARL or single-agent LLM RL. Right-column labels list
representative entries that assign credit at each level; the
\textit{sparse} levels (orchestrator, message) mark the most
under-populated research territory.}
\label{fig:credit-hierarchy}
\end{figure}
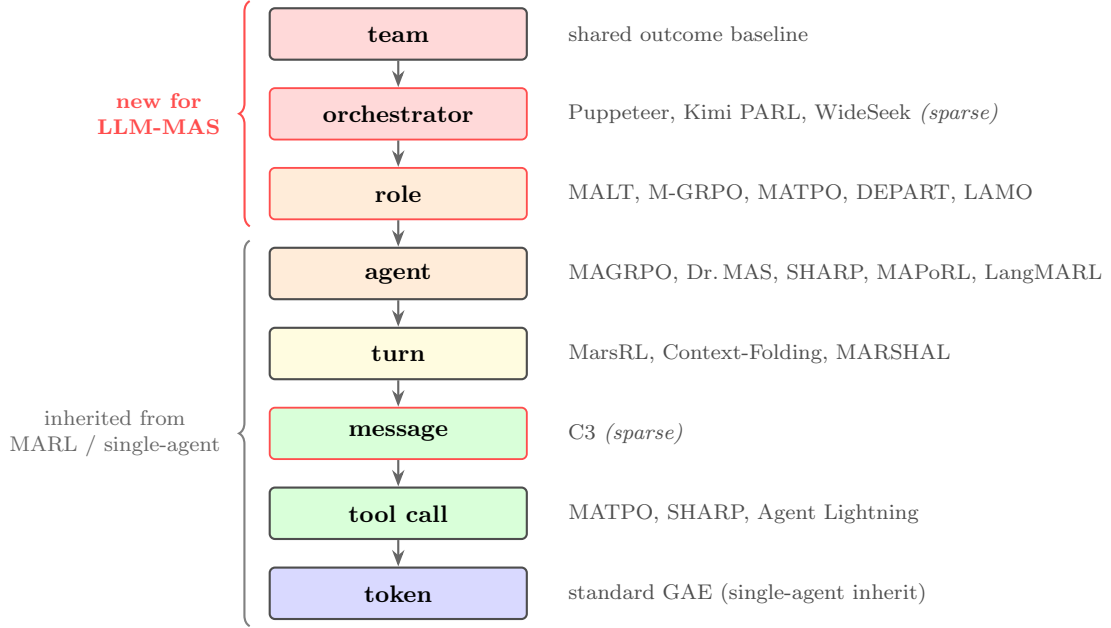

The team and agent units are inherited from cooperative MARL; role
credit has partial analogues in heterogeneous-agent MARL; and
orchestrator-decision credit is the least classical because the
decision changes the future agent set itself. The lower four units
(turn, message, tool, token) correspond to the temporal decomposition
within any one agent's trajectory, but become more consequential when
messages and tool calls mediate inter-agent information flow.
Figure~\ref{fig:credit-hierarchy} visualizes the stack together with
representative entries at each level.
An RL-oriented entry is characterized not only by whether it does
explicit counterfactual credit assignment, but also by
\emph{which level(s) in this hierarchy} carry reward, credit, or
optimization signals, and \emph{by what mechanism}.

\subsection{A two-dimensional taxonomy of entries}
\label{sec:credit:taxonomy}

Table~\ref{tab:credit-taxonomy} is the central organizing device of
this section. Rows are representative entries from the paper
pool plus a shared-outcome baseline; these include methods,
frameworks, and system anchors. Columns are the seven above-token
credit- or signal-bearing units; the token level is treated as inherited
within each agent. Each filled cell names the \emph{mechanism} by
which the entry attaches a reward, credit, or optimization signal at
that level.

\begin{table}[htbp]
\centering
\scriptsize
\renewcommand{\arraystretch}{1.35}
\setlength{\tabcolsep}{2pt}
\begin{tabularx}{\linewidth}{@{}p{1.9cm} | *{7}{>{\centering\arraybackslash}X} | p{1.0cm} >{\raggedright\arraybackslash}p{1.5cm}@{}}
\toprule
\textbf{Entry}
 & \textbf{team}
 & \textbf{orch.}
 & \textbf{role}
 & \textbf{agent}
 & \textbf{turn}
 & \textbf{msg.}
 & \textbf{tool}
 & \textbf{dyn.\ $n$?}
 & \textbf{mech.} \\
\midrule
Shared outcome (GRPO baseline)
 & direct & -- & -- & -- & -- & -- & --
 & --
 & heuristic \\

MAGRPO~\cite{magrpo2025}
 & direct & -- & -- & group adv. & -- & -- & --
 & no
 & group-norm \\

MAPoRL~\cite{maporl2025}
 & direct & -- & -- & broadcast & -- & -- & --
 & no
 & broadcast \\

MARFT~\cite{marft2025}
 & direct & -- & -- & per-agent & -- & -- & --
 & partial
 & PPO-style \\

Dr.\,MAS~\cite{dr-mas2026}
 & direct & -- & -- & \textbf{agent-norm} & -- & -- & --
 & no
 & agent-norm \\

CoLLM-MAAC~\cite{collm-maac2026}
 & direct & -- & -- & \textbf{critic} & -- & -- & --
 & no
 & actor-critic \\

LangMARL \cite{langmarl2026}
 & direct & -- & -- & \textbf{language credit} & -- & -- & --
 & no
 & lang-credit \\

MALT~\cite{malt2025}
 & direct & -- & \textbf{role-PRM} & -- & PRM & -- & --
 & no
 & learned PRM \\

MATPO~\cite{matpo2025}
 & direct & -- & dual-role & -- & -- & -- & tool outcome
 & no
 & shared-wt \\

Puppeteer~\cite{puppeteer2025}
 & direct & \textbf{critic} & -- & -- & -- & -- & --
 & yes
 & critic \\

Agent Q-Mix~\cite{agent-qmix2026}
 & direct & -- & -- & \textbf{QMIX} & -- & graph & --
 & yes
 & CTDE \\

M-GRPO~\cite{m-grpo2025}
 & direct & -- & \textbf{hier.\ base} & indep.\ adv. & -- & -- & --
 & no
 & hier.\ GRPO \\

MarsRL~\cite{marsrl2025}
 & direct & -- & -- & -- & \textbf{pipeline} & -- & --
 & no
 & stage-wise \\

C3~\cite{c3-2026}
 & direct & -- & -- & -- & -- & \textbf{CF} & --
 & no
 & CF \\

SHARP~\cite{sharp2026}
 & direct & -- & -- & \textbf{Shapley} & -- & -- & tool-proc.
 & no
 & Shapley \\

HERA~\cite{hera2026}
 & direct & evolve & -- & -- & -- & -- & --
 & partial
 & evol. \\

Kimi PARL~\cite{kimi-k2-5-2026}
 & direct & \textbf{Crit-Steps} & -- & per-sub & -- & -- & --
 & \textbf{yes}
 & heuristic \\

WideSeek-R1~\cite{wideseek-r1-2026}
 & direct & \textbf{lead/sub} & -- & per-sub & -- & -- & --
 & yes
 & width-RL \\

Context-Folding~\cite{context-folding2025}
 & direct & -- & -- & -- & \textbf{branch} & -- & --
 & partial
 & fold \\

MARSHAL \cite{marshal2026}
 & direct & -- & -- & agent-norm & \textbf{turn adv.} & -- & --
 & no
 & self-play \\

DEPART~\cite{depart2026}
 & direct & -- & \textbf{HIMPO} & -- & -- & -- & --
 & no
 & hier.\ PO \\

Agent~Lightning~\cite{agent-lightning2025}
 & direct & -- & -- & framework & turn & -- & tool
 & yes
 & harness-gen. \\

\bottomrule
\end{tabularx}
\caption{Two-dimensional taxonomy of credit and signal assignment in LLM-MAS RL.
Rows are representative entries plus a shared-outcome baseline;
columns 2--8 are
credit- or signal-bearing units above the token level
(\S\ref{sec:credit:hierarchy}); the token level is omitted because all
entries inherit standard token-level GAE within each agent.
\textbf{dyn.\ $n$?} = does the method accommodate a time-varying agent
count (as in Kimi Agent Swarm)? \textbf{mechanism} = a one-phrase
summary of how credit is mechanically computed. Cell entries name
the decomposition or optimization signal applied at each granularity;
\textbf{bold} marks the method's distinctive contribution.
Not every filled cell is an explicit counterfactual credit-assignment
method; C3 is the retained entry that estimates counterfactual
message-level credit.
``direct'' = the team reward is applied without decomposition;
``--'' = the method does not operate at that granularity.
Empty cells for a given method do \emph{not} mean the method is
incomplete---they mean the method's novelty lives elsewhere in the
hierarchy.}
\label{tab:credit-taxonomy}
\end{table}

Three patterns in the table are worth naming explicitly.

\begin{itemize}
  \item \textbf{Most RL-oriented entries contribute at exactly one novel level.}
    Reading down the ``bold'' cells: Dr.\,MAS, CoLLM-MAAC, LangMARL,
    and Agent Q-Mix at agent / topology-conditioned agent credit,
    MALT at role,
    Puppeteer at orchestrator, M-GRPO at role, MarsRL at turn, C3 at
    message, SHARP at agent, Kimi PARL and WideSeek-R1 at
    orchestrator, and MARSHAL at turn. Each entry
    introduces or documents a technique for \emph{one} layer of the
    hierarchy and leaves the others to standard machinery (GRPO, GAE,
    broadcast).
    This is not a criticism---it is how a research community makes
    progress---but it does mean no single method yet covers the full
    hierarchy.
  \item \textbf{The orchestrator level is sparsely populated.}
    The CSV tags eight retained entries at the orchestrator level,
    but most are design- or evolution-level orchestration signals.
    Puppeteer, Kimi PARL, and WideSeek-R1 are the clearest cases that explicitly
    attach an optimization signal to orchestrator decisions, and they
    do so by very different mechanisms (learned central critic vs.\
    Critical-Steps heuristic). We read explicit orchestrator credit,
    rather than orchestration as a system form, as the underdeveloped
    column in the table.
  \item \textbf{The message level is even sparser.}
    Classical MARL has a long tradition of communication-level
    credit (difference rewards on messages), but our curated LLM-MAS
    pool still contains only two entries tagged at the message level:
    Debate-as-Reward uses message-level debate outcomes as a reward
    signal, while C3~\cite{c3-2026} is the only retained entry that
    explicitly estimates counterfactual message-level credit. This is
    a wide-open research direction.
\end{itemize}

\begin{figure}[t]
\centering
\resizebox{\linewidth}{!}{%
\begin{tikzpicture}[font=\footnotesize,
  q/.style={diamond, draw=black!70, thick, fill=yellow!15,
            aspect=2.2, inner sep=2pt, align=center,
            minimum height=0.8cm, font=\scriptsize},
  leaf/.style={rectangle, rounded corners=2pt,
               draw=blue!65, thick, fill=blue!10,
               minimum width=1.65cm, minimum height=0.82cm,
               text width=1.55cm, align=center, font=\scriptsize},
  arr/.style={-{Stealth[length=2mm]}, thick, black!60},
  ylab/.style={anchor=east, font=\scriptsize, black!65,
               inner sep=1pt, pos=0.5},
  nlab/.style={anchor=west, font=\scriptsize, black!65,
               inner sep=1pt, pos=0.5}
]

  \node[q] (q1) at (7, 6.5) {Dynamic agent count?\\(spawn/despawn at inference)};

  \node[q] (q2a) at (3, 5.0) {Is orchestrator\\the bottleneck?};
  \node[q] (q2b) at (11, 5.0) {Long trace?\\($T \gtrsim 10^3$?)};

  \draw[arr] (q1) -- node[ylab, sloped] {yes} (q2a);
  \draw[arr] (q1) -- node[nlab, sloped] {no} (q2b);

  \node[q] (q3a) at (1, 3.5) {Need fine\\inter-agent\\attribution?};
  \node[q] (q3b) at (5, 3.5) {Heterogeneous\\roles?};
  \node[q] (q3c) at (9, 3.5) {Pipeline\\factorable?};
  \node[q] (q3d) at (13, 3.5) {Debate-\\shaped?};

  \draw[arr] (q2a) -- node[ylab, sloped] {yes} (q3a);
  \draw[arr] (q2a) -- node[nlab, sloped] {no}  (q3b);
  \draw[arr] (q2b) -- node[ylab, sloped] {yes} (q3c);
  \draw[arr] (q2b) -- node[nlab, sloped] {no}  (q3d);

  \node[leaf] (l1) at (-0.2, 1.6) {SHARP\\(Shapley)};
  \node[leaf] (l2) at (2.2,  1.6) {Puppeteer\\(critic)};
  \node[leaf] (l3) at (4.0,  1.6) {Kimi PARL\\(Crit.-Steps)};
  \node[leaf] (l4) at (6.4,  1.6) {MALT\\(role-PRM)};
  \node[leaf] (l5) at (8.4,  1.6) {MarsRL\\(pipeline\\stage)};
  \node[leaf] (l6) at (10.6, 1.6) {Context-\\Folding\\(turn)};
  \node[leaf] (l7) at (12.6, 1.6) {C3\\(msg CF)};
  \node[leaf] (l8) at (14.8, 1.6) {Dr.MAS\\(agent norm)};

  \draw[arr] (q3a) -- node[ylab, sloped] {yes} (l1);
  \draw[arr] (q3a) -- node[nlab, sloped] {no}  (l2);
  \draw[arr] (q3b) -- node[ylab, sloped] {yes} (l4);
  \draw[arr] (q3b) -- node[nlab, sloped] {no}  (l3);
  \draw[arr] (q3c) -- node[ylab, sloped] {yes} (l5);
  \draw[arr] (q3c) -- node[nlab, sloped] {no}  (l6);
  \draw[arr] (q3d) -- node[ylab, sloped] {yes} (l7);
  \draw[arr] (q3d) -- node[nlab, sloped] {no}  (l8);

  \node[draw=black!35, rounded corners=2pt, fill=gray!6,
        text width=14cm, font=\scriptsize, align=left,
        anchor=north]
    at (7, 0.6)
    {\textbf{Usage note.} This tree is a first-pass heuristic mapping
      system properties to the credit-assignment method in our pool
      whose design most closely matches them. Real systems composite
      multiple approaches (e.g., Kimi PARL uses Critical-Steps \emph{and}
      per-subagent credit). ``No'' edges point to the default; ``yes''
      edges to the specialized choice.};

\end{tikzpicture}%
}
\caption{A decision-tree heuristic for selecting a credit-assignment
mechanism from our pool, organized by four system-level questions:
(i)~whether the agent set is dynamic at inference, (ii)~whether the
orchestrator is the identified bottleneck, (iii)~whether traces are
long enough to suffer diffusion, and (iv)~whether roles are
heterogeneous or the structure is debate-shaped. Leaves name the
method whose design target most closely matches the path; in
practice, industrial systems composite multiple choices rather
than pick exactly one.}
\label{fig:credit-decision-tree}
\end{figure}
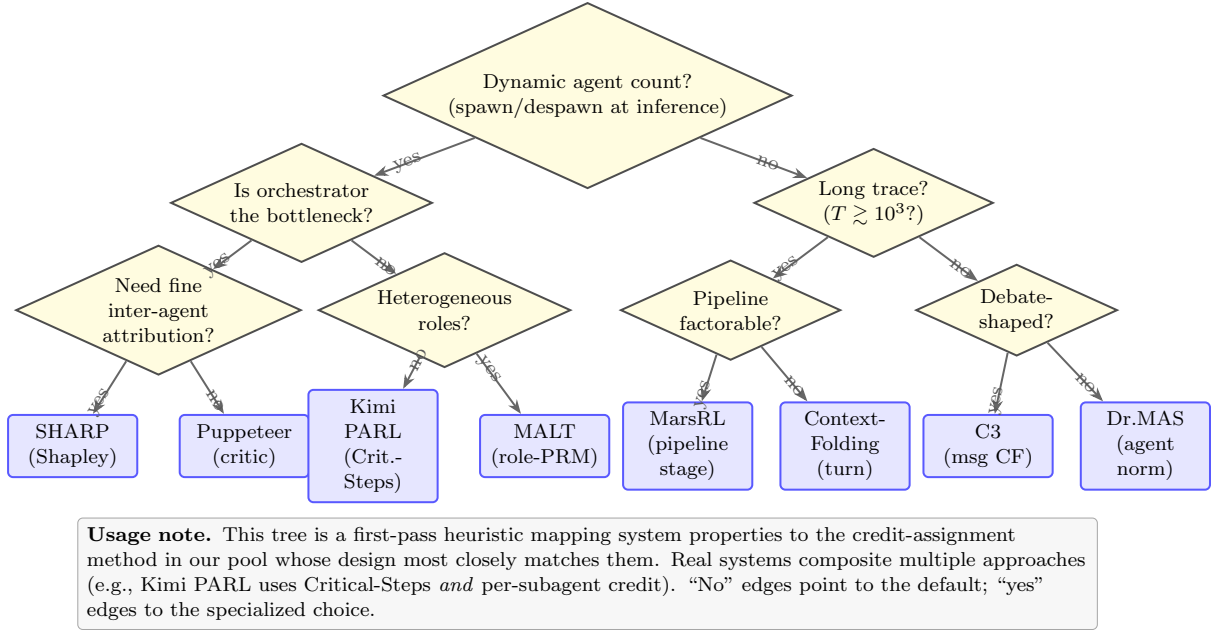

A natural follow-up question is whether these mechanisms compose.
Nothing in the taxonomy prevents stacking, e.g., Puppeteer's
orchestrator critic on top of C3's message counterfactuals on top of
Dr.\,MAS's agent-wise normalization. The May 2026 additions make
agent-, role-, and turn-level credit denser, but we are not aware of
published work in our pool that composes explicit credit mechanisms
across all these levels; we return to this in
\S\ref{sec:open}. For practitioners selecting among the methods,
Figure~\ref{fig:credit-decision-tree} gives a first-pass decision
heuristic keyed to four system-level properties.

\subsection{Why naive single-agent credit assignment fails}
\label{sec:credit:fails}

A tempting position is that team-level reward plus standard GAE
within each agent is enough, and that the rest of
Table~\ref{tab:credit-taxonomy} is decoration. Three failure modes
observed in the pool argue against this position.

\begin{itemize}
  \item \textbf{Reward diffusion.} Kimi's Agent Swarm traces reach
    up to $1{,}500$ coordinated steps / tool calls in the public K2.5
    training anchor~\cite{kimi-k2-5-2026}. Even at that scale, a
    single terminal reward distributed over realized training decisions
    can make the per-decision learning signal fragile and low-SNR.
    K2.6 extends the public deployment envelope to $4{,}000$
    coordinated steps~\cite{kimi-k2-6-2026}; we use this only as
    scale-pressure evidence, not as an independently disclosed
    RL-training trajectory. Dr.\,MAS~\cite{dr-mas2026}
    documents a related empirical symptom as training instability that
    is not cured by hyperparameter tuning alone.
  \item \textbf{Asymmetric contribution.} A single critic
    message---``this plan will not work because $X$''---can flip a
    $1{,}500$-step trace from failure to success. Uniform credit over
    agents (or uniform advantage over messages) assigns this
    pivotal message the same weight as routine executor chatter.
    C3~\cite{c3-2026} argues that this is the right setting for
    counterfactual message-level credit specifically \emph{because}
    contributions are so heavy-tailed.
  \item \textbf{Counterfactual ambiguity about spawning.} When the
    orchestrator spawns a sub-agent and the trace succeeds, was the
    sub-agent responsible, or would the orchestrator have succeeded
    without spawning at all? No reward defined over realized
    trajectories can answer this; it requires a counterfactual over
    an \emph{unrealized} alternative trace. This is why orchestrator
    credit is fundamentally harder than agent credit---and why the
    orchestrator column in Table~\ref{tab:credit-taxonomy} is sparse.
\end{itemize}

\subsection{Open algorithmic questions}
\label{sec:credit:open}

\begin{itemize}
  \item \textbf{Compositionality.} Can the mechanisms in
    Table~\ref{tab:credit-taxonomy} be stacked (\eg Puppeteer's
    orchestrator critic $+$ C3's message counterfactual $+$
    Dr.\,MAS's agent normalization) without their regularizing
    effects cancelling? SHARP's Shapley+tool-process, MARSHAL's
    turn-level estimator plus agent-specific normalization, and
    DEPART's dense role-specific plus sparse task rewards are partial
    composites, but each remains limited to a small subset of the
    hierarchy.
  \item \textbf{Process vs.\ outcome balance.} When a dense PRM
    signal (R4) is combined with a sparse team reward (R1), the
    dense signal typically dominates gradients and the policy drifts
    toward what the PRM rewards rather than what the task rewards.
    MALT~\cite{malt2025} uses role-specific PRMs to reduce this; a
    general principle is missing.
  \item \textbf{Dynamic-agent Shapley.} SHARP computes Shapley credit
    over a fixed agent set. In systems like Kimi Agent Swarm where
    the agent set is itself produced by a policy decision (spawn /
    despawn), classical Shapley axioms no longer hold. A
    Shapley-analogue for dynamic coalitions is open.
  \item \textbf{Credit for the decision \emph{not} to spawn.} The
    orchestrator's policy space includes ``do nothing.'' This
    decision does not produce a realized sub-trace, and the entries in
    Table~\ref{tab:credit-taxonomy} do not address it. A principled
    treatment likely requires off-policy evaluation of unrealized
    branches---a direction for which we found no entry in the pool
    as of May 4, 2026.
\end{itemize}

\subsection{Discussion: composing the sparse and dense levels}
\label{sec:credit:discussion}

The taxonomy in Table~\ref{tab:credit-taxonomy} still reads mostly as
a story about \emph{single-level interventions}: most methods target
one dominant level and inherit standard machinery on the others. Three
synthesis observations follow.

\textbf{Density and visibility trade off.} The token level is the
densest signal source available (every generated token gives a
gradient through GAE) but is also the level most divorced from
team outcome. The team level is the cleanest signal source but
emits one number per trace. All other levels lie on a Pareto
frontier between these endpoints---role and turn levels yielding
medium density and medium outcome-attribution; orchestrator and
message levels yielding low density (few decisions per trace) but
high attribution (each decision is consequential). Newer entries such
as DEPART~\cite{depart2026} and MARSHAL~\cite{marshal2026} mix sparse
task reward with denser role or turn signals, but an exact partition
of credit across the full hierarchy remains absent.

\textbf{Counterfactual-based methods are quadratically expensive
and structurally fragile.} Both C3~\cite{c3-2026} and the spawn
counterfactual hinted at by \S\ref{sec:credit:fails} require
estimating the return of an alternative trace not actually
produced. C3 handles this at the message level by sampling
substitute messages; the cost is at least linear in number of
messages times sample count, and grows quadratically if the
substitution is itself contextual. SHARP's Shapley
sampling~\cite{sharp2026} is similarly Monte-Carlo expensive.
Practical counterfactual credit at Kimi-reported long trace lengths
($T \sim 10^3$, \S\ref{sec:eng:trace}) would require either a
learned counterfactual estimator (a ``what-would-have-happened
model'' trained from off-policy data) or some form of importance
sampling over realized branches. Neither has a published
instance in our pool.

\textbf{Compositionality is not for free.} Stacking, e.g.,
agent-wise normalization (Dr.\,MAS) below a learned orchestrator
critic (Puppeteer) below a message-level counterfactual (C3) is
algebraically possible but introduces a new failure mode:
\emph{credit double-counting}. If a pivotal message is rewarded
once via C3 and again via Puppeteer's critic for the orchestrator's
delegation that produced the message, the policy receives stronger
gradient on that message than on others---a kind of credit
collision that classical MARL avoids by design (each agent has
exactly one credit channel via CTDE). A clean compositional
framework would need to specify a partition of the team reward
across credit channels and enforce that the partition is exact.
We found no such framework in our pool.

\takeaway{The central technical claim of this survey is that LLM-MAS methods
should be read along a \emph{credit- and signal-bearing-unit hierarchy},
not a flat list of ``multi-agent RL tricks.'' Under that reading, the
literature as of May 4, 2026 populates some columns densely (agent
via MAGRPO / Dr.\,MAS / SHARP / LangMARL / CoLLM-MAAC, role via
MALT / M-GRPO / DEPART / LAMO) and others
sparsely (especially explicit counterfactual message credit, and still
explicit orchestrator credit)---and the sparse columns are where
the near-term research opportunity is concentrated. Composing
across columns introduces credit double-counting risks that no
published method in our pool has formally addressed.
}

\section{Learning Orchestration: Trajectory \texorpdfstring{$\to$}{->} Orchestration Trace}
\label{sec:orchestration}

Reward design (\S\ref{sec:rewards}) and credit assignment
(\S\ref{sec:credit}) answer the questions \emph{what to measure} and
\emph{where to assign it}. This section answers \emph{what is being
optimized}: the orchestration trace. We make the object formal
(\S\ref{sec:orch:trace}), organize methods by which orchestration
sub-decision they learn (\S\ref{sec:orch:methods}), discuss training
regimes (\S\ref{sec:orch:regimes}), survey engineering constraints
(\S\ref{sec:orch:engineering}), and enumerate failure modes specific
to orchestration learning (\S\ref{sec:orch:failures}).

\subsection{The orchestration trace as a first-class object}
\label{sec:orch:trace}

A \emph{trajectory} in single-agent RL is a sequence
$\tau = (s_0, a_0, r_0, s_1, a_1, r_1, \ldots, s_T)$. An
\emph{orchestration trace} is a temporal interaction graph
$G = (V, E, \ell)$, where:

\begin{itemize}
  \item $V$ is a set of events:
    orchestrator decisions, sub-agent invocations, tool calls,
    messages, summary returns, and aggregation points.
  \item $E \subseteq V \times V$ is a set of temporal/causal
    dependencies: ``this sub-agent was spawned by that orchestrator
    decision'', ``this aggregator consumed those summaries'',
    ``this tool call followed that planning message''.
  \item $\ell: V \to (\text{agent}, \text{role}, \text{content})$
    labels each event with the executing agent, its role, and
    structured content.
\end{itemize}

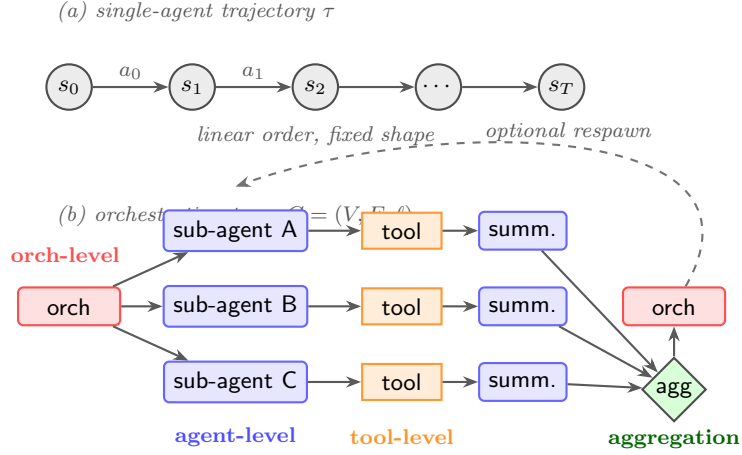
\begin{figure}[t]
\centering
\begin{tikzpicture}[
    font=\footnotesize,
    node distance=0.9cm and 1.0cm,
    trjstep/.style={
      circle, draw=black!70, thick, fill=gray!15,
      minimum size=6mm, inner sep=0pt
    },
    orch/.style={
      rectangle, rounded corners=2pt, draw=red!70, thick, fill=red!12,
      minimum width=1.35cm, minimum height=0.5cm, align=center
    },
    subagent/.style={
      rectangle, rounded corners=2pt, draw=blue!60, thick, fill=blue!10,
      minimum width=1.15cm, minimum height=0.5cm, align=center
    },
    tool/.style={
      rectangle, draw=orange!80, thick, fill=orange!15,
      minimum width=1.05cm, minimum height=0.45cm, align=center
    },
    agg/.style={
      diamond, draw=black!70, thick, fill=green!15,
      minimum size=0.85cm, inner sep=0pt, aspect=1.6
    },
    arr/.style={-{Stealth[length=2mm]}, thick, black!65},
    darr/.style={-{Stealth[length=2mm]}, thick, dashed, black!55},
    lbl/.style={font=\scriptsize\itshape, black!65}
]

\node[lbl, anchor=west] at (-0.3, 2.1) {(a) single-agent trajectory $\tau$};

\node[trjstep] (s0) at (0, 1.1)    {$s_0$};
\node[trjstep, right=of s0] (s1)   {$s_1$};
\node[trjstep, right=of s1] (s2)   {$s_2$};
\node[trjstep, right=of s2] (s3)   {$\cdots$};
\node[trjstep, right=of s3] (sT)   {$s_T$};

\draw[arr] (s0) -- node[above=-1pt]{\scriptsize $a_0$} (s1);
\draw[arr] (s1) -- node[above=-1pt]{\scriptsize $a_1$} (s2);
\draw[arr] (s2) -- (s3);
\draw[arr] (s3) -- (sT);

\node[lbl, below=2pt of s2] {linear order, fixed shape};

\node[lbl, anchor=west] at (-0.3, -0.6) {(b) orchestration trace $G=(V,E,\ell)$};

\node[orch] (o1) at (0,    -1.8) {\textsf{orch}};
\node[orch] (o2) at (8.0,  -1.8) {\textsf{orch}};

\node[subagent] (a1) at (2.2, -0.8)  {\textsf{sub-agent A}};
\node[subagent] (a2) at (2.2, -1.8)  {\textsf{sub-agent B}};
\node[subagent] (a3) at (2.2, -2.8)  {\textsf{sub-agent C}};

\node[tool] (t1) at (4.4, -0.8)  {\textsf{tool}};
\node[tool] (t2) at (4.4, -1.8)  {\textsf{tool}};
\node[tool] (t3) at (4.4, -2.8)  {\textsf{tool}};

\node[subagent, minimum width=0.85cm] (r1) at (6.0, -0.8)  {\textsf{summ.}};
\node[subagent, minimum width=0.85cm] (r2) at (6.0, -1.8)  {\textsf{summ.}};
\node[subagent, minimum width=0.85cm] (r3) at (6.0, -2.8)  {\textsf{summ.}};

\node[agg] (aggnode) at (8.0, -2.9) {\textsf{agg}};

\draw[arr] (o1) -- (a1);
\draw[arr] (o1) -- (a2);
\draw[arr] (o1) -- (a3);

\foreach \a/\t/\r in {a1/t1/r1, a2/t2/r2, a3/t3/r3} {
  \draw[arr] (\a) -- (\t);
  \draw[arr] (\t) -- (\r);
}

\draw[arr] (r1) -- (aggnode);
\draw[arr] (r2) -- (aggnode);
\draw[arr] (r3) -- (aggnode);

\draw[arr] (aggnode) -- (o2);

\draw[darr] (o2) to[out=60, in=20, looseness=1.1]
  node[lbl, above=-2pt, midway]{optional respawn}
  ($(a1.north)+(0,0.3)$);

\node[lbl, red!70, font=\scriptsize\bfseries,
      fill=white, fill opacity=0.9, text opacity=1, inner sep=1pt]
      at (-0.05, -1.12) {orch-level};
\node[lbl, blue!70, font=\scriptsize\bfseries,
      fill=white, fill opacity=0.9, text opacity=1, inner sep=1pt]
      at (2.2, -3.52) {agent-level};
\node[lbl, orange!85, font=\scriptsize\bfseries,
      fill=white, fill opacity=0.9, text opacity=1, inner sep=1pt]
      at (4.4, -3.52) {tool-level};
\node[lbl, green!40!black, font=\scriptsize\bfseries,
      fill=white, fill opacity=0.9, text opacity=1, inner sep=1pt]
      at (8.0, -3.55) {aggregation};

\end{tikzpicture}
\caption{Optimization objects for single-agent LLM RL vs.\ LLM-MAS
RL. \textbf{(a)} A \emph{trajectory} $\tau$ is a linearly ordered
sequence of $(s_t, a_t)$ pairs. \textbf{(b)} An \emph{orchestration
trace} $G=(V,E,\ell)$ is a temporal interaction graph: orchestrator
decisions (red) spawn sub-agents (blue), which issue tool calls
(orange) and return summaries that are aggregated (green diamond)
before the next orchestrator decision. Credit-bearing units
(\S\ref{sec:credit}) attach at distinct substructures of $G$
rather than to time-indexed states. Compared to (a), the trace in
(b) has branching, joining, and variable shape across rollouts.}
\label{fig:orchestration-trace}
\end{figure}

A trajectory is a linearly ordered special case of an orchestration
trace ($|V|$ = episode length, $E$ = successor relation). The
multi-agent case is genuinely graph-structured: branching (parallel
spawn), joining (aggregation), and delegation (orchestrator $\to$
sub-agent) have no trajectory analogue. Figure~\ref{fig:orchestration-trace}
contrasts the two objects.
Consequently, in our taxonomy, the optimization target is naturally
defined over $G$:
\[
  \max_{\theta} \; \mathbb{E}_{G \sim \pi_{\theta}}\!\big[\,R(G)\,\big],
\]
where $R$ is the composite reward from
Table~\ref{tab:reward-families} and $\pi_{\theta}$ is the joint
policy (orchestrator + sub-agents + aggregation). For RL-oriented
entries, we use this as the common comparison object, even when the
original papers do not frame their objectives this way; framework,
benchmark, and industrial-anchor entries are used only where their
public material constrains this comparison.

\subsection{Methods by orchestration sub-decision}
\label{sec:orch:methods}

An orchestration trace is produced by a sequence of sub-decisions,
each of which can, in principle, be the target of a learned policy.
We identify five (Figure~\ref{fig:orch-subdecisions}). The remainder
of this subsection expands each one: what the decision is, what
signal would train it, what retained entries address it, and
what remains open.

\begin{figure}[t]
\centering
\resizebox{\linewidth}{!}{%
\begin{tikzpicture}[
    font=\footnotesize,
    decision/.style={
      rectangle, rounded corners=2pt, draw=black!70, thick,
      minimum width=2.8cm, minimum height=1.0cm, align=center,
      fill=#1!12
    },
    methods/.style={
      font=\scriptsize, align=center, black!75,
      text width=2.6cm
    },
    flow/.style={-{Stealth[length=2mm]}, thick, black!55},
    sparse/.style={dashed, draw=red!70, line width=0.7pt}
]

  \node[decision=red]    (o1) at (0,    0) {\textbf{O1}\\ when to\\ \textbf{spawn}};
  \node[decision=orange] (o2) at (3.0,  0) {\textbf{O2}\\ whom to\\ \textbf{delegate}};
  \node[decision=yellow] (o3) at (6.0,  0) {\textbf{O3}\\ how to\\ \textbf{communicate}};
  \node[decision=green]  (o4) at (9.0,  0) {\textbf{O4}\\ how to\\ \textbf{aggregate}};
  \node[decision=blue, sparse] (o5) at (12.0, 0)
       {\textbf{O5}\\ when to\\ \textbf{stop}};

  \foreach \a/\b in {o1/o2, o2/o3, o3/o4, o4/o5}
    \draw[flow] (\a) -- (\b);

  \node[methods, below=0.4cm of o1]
    {Kimi PARL\\ AgentSpawn\\ HALO};
  \node[methods, below=0.4cm of o2]
    {Puppeteer\\ ParaManager};
  \node[methods, below=0.4cm of o3]
    {Debate-as-Reward\\ LatentMAS};
  \node[methods, below=0.4cm of o4]
    {M-GRPO\\ Context-Folding};
  \node[methods, below=0.4cm of o5, red!70, font=\scriptsize\itshape]
    {\textbf{no published}\\ \textbf{training method}};

  \draw[decorate, decoration={brace, amplitude=4pt}, thick, black!50]
    ($(o1.north west)+(0,0.15)$) --
    ($(o5.north east)+(0,0.15)$)
    node[midway, above=6pt, black!70, font=\footnotesize\itshape]
    {orchestrator's decision chain per task};

  \node[font=\scriptsize, black!55, anchor=north, align=center]
    at (6.0, -2.55) {(rows = representative entries from
    Table~\ref{tab:credit-taxonomy}; the dashed red box marks the
    sub-decision with no entry in our curated pool as of May 4, 2026)};

\end{tikzpicture}%
}
\caption{The five orchestration sub-decisions O1--O5
(\S\ref{sec:orch:methods}). An orchestrator policy makes some or
all of these decisions per task; surveyed entries cover O1--O4 but
\textbf{not O5}. The red dashed box marks ``when to stop'' as a
named open problem (\S\ref{sec:open}): in the entries we surveyed,
termination is either externally signaled (ground-truth answer
found) or triggered by a fixed step-count cap rather than explicitly
trained as a stopping policy.}
\label{fig:orch-subdecisions}
\end{figure}

\subsubsection{O1: When to spawn}
\label{sec:orch:o1}

\textbf{Decision.} Given the current partial trace, does the
orchestrator issue a \texttt{spawn} action, and with what role /
context? The policy's support includes \texttt{no-op}; exercising
\texttt{spawn} commits to downstream rollout cost
(\S\ref{sec:eng:rollout}).

\textbf{What signal would train it?} Ideally, the counterfactual
team return under \texttt{spawn} vs \texttt{no-op}. Because
\texttt{no-op} is never actually rolled out once \texttt{spawn} is
chosen, the counterfactual is unobserved---this is the
non-iden\-tifi\-ability argument stated as Claim~\ref{claim:nonid} in
\S\ref{sec:formalism:claims}.

\textbf{Entries in our pool.} Kimi~PARL~\cite{kimi-k2-5-2026} makes
spawning an action and gives it two reward shapes:
$r_{\text{parallel}}$ (against serial collapse) and
$r_{\text{finish}}$ (against spurious parallelism), with both
auxiliary weights reported as annealed to zero over training
(Fig.~\ref{fig:parl-annealing}). AgentSpawn~\cite{agentspawn2026}
triggers spawn via learned complexity estimators at runtime;
HALO~\cite{halo2025} applies MCTS over spawn decisions at a
hierarchical level, treating spawn as planning rather than as an
RL action.

\textbf{What remains open.} None of the three methods uses an
explicit counterfactual estimator; all use $R7$ shaping or search
heuristics as proxies. A principled off-policy evaluation of
unrealized \texttt{no-op} branches is the obvious missing piece
(\S\ref{sec:open}, P4).

\subsubsection{O2: Whom to delegate to}
\label{sec:orch:o2}

\textbf{Decision.} Conditional on \texttt{spawn} being chosen, which
agent among the currently instantiated pool $\mathcal{I}_t$ (or
newly created agent of a given role) receives the next task chunk?

\textbf{What signal would train it?} A per-delegation return
differential---which agent, in context, would have produced the
best outcome. This is the classical setting for a centralized
critic.

\textbf{Entries in our pool.} Puppeteer~\cite{puppeteer2025} trains
exactly such a learned central critic in a CTDE style
(\S\ref{sec:background:marl}), freezing sub-agents and updating
only the orchestrator. ParaManager~\cite{paramanager2026}
generalizes the support: agent and tool dispatch share a unified
action space
$\mathcal{A}_{\text{delegate}} = \{\text{sub-agent}_i\} \cup \{\text{tool}_j\}$,
which lets the orchestrator trade off between creating a sub-agent
and directly calling a tool with no delegation overhead. WideSeek-R1
jointly optimizes a lead agent and parallel sub-agents for broad
information seeking, making width scaling itself part of the learned
delegation regime~\cite{wideseek-r1-2026}.

\textbf{What remains open.} These academic works still operate far
below the largest disclosed industrial swarms. In industrial swarms
(\S\ref{sec:systems:kimi}) the pool is dynamic and can grow to
hundreds; scaling a learned dispatcher to that regime is
unaddressed.

\subsubsection{O3: How to communicate}
\label{sec:orch:o3}

\textbf{Decision.} What is the content, length, and format of
messages exchanged between orchestrator and sub-agents, and among
sub-agents?

\textbf{What signal would train it?} A reward or credit signal for a
specific message's contribution to team outcome---the message-level
unit of \S\ref{sec:credit:hierarchy}. Explicit counterfactual
message-level credit is the sparsest column of
Table~\ref{tab:credit-taxonomy}.

\textbf{Entries in our pool.} Three retained RL entries engage O3
directly.
Debate-as-Reward~\cite{debate-as-reward2026} rewards
resolution-based messages, disincentivizing length-based farming.
C3~\cite{c3-2026} estimates counterfactual contribution per
message via contextual intervention. Agent Q-Mix~\cite{agent-qmix2026}
learns decentralized communication/topology decisions with a
QMIX-style CTDE objective, treating the round-wise communication graph
as the object to optimize. LatentMAS~\cite{latentmas2025}
takes the opposite route: replace token-level messages with a
continuous latent channel, eliminating the message-level
credit-assignment problem by changing the communication medium
altogether (\emph{and}, empirically, gaining $+14.6\%$ without any
training).

\textbf{What remains open.} The field now has several point-solutions
(token counterfactual, learned topology, latent channel) but still no
unified information-theoretic account. A
principled treatment of \emph{which bits of information} an
orchestrator should exchange---the direct LLM analogue of
Shannon-rate constraints in classical emergent
communication---remains absent from our pool.

\subsubsection{O4: How to aggregate}
\label{sec:orch:o4}

\textbf{Decision.} When sub-agents return partial results, how does
the orchestrator combine them into the trace state that gates the
next decision? Summaries, votes, consensus, or structured merge.

\textbf{What signal would train it?} The aggregation step is
itself a policy output; it can receive either team reward
(slow-moving) or a per-aggregation proxy (\eg whether aggregated
output contains the key fact needed for the next sub-decision).

\textbf{Entries in our pool.}
M-GRPO~\cite{m-grpo2025} formalizes aggregation as a separate main
agent whose policy consumes sub-agent summaries and emits
trajectory continuations. Context-Folding~\cite{context-folding2025}
treats aggregation as an explicit agent action, rewarding branch
outcomes approximately as
$r_{\text{branch}} \approx r_{\text{main}} \pm 0.2$ scope
adjustment.

\textbf{What remains open.} Both methods aggregate via LLM
summarization---lossy and uncalibrated. An aggregator that
explicitly models the uncertainty of sub-agent claims (\eg a
Bayesian combiner) has no entry in our pool.

\subsubsection{O5: When to stop}
\label{sec:orch:o5}

\textbf{Decision.} At which point does the orchestrator halt the
trace and emit the final answer?

\textbf{What signal would train it?} Expected marginal gain of one
more orchestration step vs. the cost of that step
(\S\ref{sec:eng:rollout}). A stopping policy that trades accuracy
for cost is a natural objective.

\textbf{Entries in our pool.} We found no retained entry that trains
this decision directly. Existing entries stop either externally
(\eg ground-truth answer verifier signals completion) or at a fixed
step-count cap. The orchestrator's
\texttt{stop} action is, as far as we can tell from public
material, not explicitly trained as an RL target in any entry in
our curated pool.

\textbf{What remains open.} This is the sub-decision with the
clearest shape of an open research direction: a small
modification to any orchestrator policy that adds a \texttt{stop}
action and trains it against a cost-adjusted return would be the
first entry in this cell of the taxonomy.

\subsection{Orchestrator training regimes}
\label{sec:orch:regimes}

\begin{figure}[t]
\centering
\resizebox{\linewidth}{!}{%
\begin{tikzpicture}[font=\footnotesize,
  orch/.style={rectangle, rounded corners=2pt,
               draw=red!70, thick, fill=red!12,
               minimum width=1.7cm, minimum height=0.7cm,
               align=center, font=\scriptsize},
  sub/.style={rectangle, rounded corners=2pt,
              draw=blue!65, thick, fill=blue!12,
              minimum width=1.3cm, minimum height=0.55cm,
              align=center, font=\scriptsize},
  frozen/.style={dashed, draw=black!60},
  grad/.style={-{Stealth[length=2mm]}, very thick, red!70!black},
  data/.style={-{Stealth[length=2mm]}, thick, black!55},
  note/.style={font=\scriptsize\itshape, black!60, anchor=north, align=center},
  title/.style={font=\footnotesize\bfseries, anchor=south}
]

\begin{scope}[shift={(0, 0)}]
  \node[title] at (1.5, 2.25) {(A) Orchestrator-only};

  \node[orch] (oA)   at (1.5, 1.3)  {orch $\pi_\theta$};
  \node[sub, frozen] (sA1) at (0.3, 0.1)  {sub 1};
  \node[sub, frozen] (sA2) at (1.5, 0.1)  {sub 2};
  \node[sub, frozen] (sA3) at (2.7, 0.1)  {sub 3};

  \draw[data] (oA.south west) -- (sA1.north east);
  \draw[data] (oA.south) -- (sA2.north);
  \draw[data] (oA.south east) -- (sA3.north west);

  \draw[grad] (-0.15, -0.7) -- (oA.west)
    node[midway, left=2pt, red!70!black, font=\scriptsize\bfseries,
         fill=white, fill opacity=0.9, text opacity=1, inner sep=1pt]
    {$\nabla_\theta R$};

  \node[note] at (1.5, -0.95)
    {team reward $R$};

  \node[font=\tiny\itshape, black!55] at (1.5, -0.45) {sub-agents frozen};

  \node[note, anchor=north, align=center, font=\scriptsize]
    at (1.5, -1.3)
    {\textbf{Kimi PARL (stage 1)}\\ \textbf{Puppeteer}};
\end{scope}

\begin{scope}[shift={(5.5, 0)}]
  \node[title] at (1.5, 2.25) {(B) Joint, shared baseline};

  \node[orch] (oB)   at (1.5, 1.3)  {orch $\pi_\theta$};
  \node[sub]  (sB1) at (0.3, 0.1)  {sub 1};
  \node[sub]  (sB2) at (1.5, 0.1)  {sub 2};
  \node[sub]  (sB3) at (2.7, 0.1)  {sub 3};

  \draw[data] (oB.south west) -- (sB1.north east);
  \draw[data] (oB.south) -- (sB2.north);
  \draw[data] (oB.south east) -- (sB3.north west);

  \draw[grad] (0.8, -0.7) -- (sB1.south east);
  \draw[grad] (1.5, -0.7) -- (oB.south);
  \draw[grad] (2.2, -0.7) -- (sB3.south west);

  \node[note] at (1.5, -0.9)
    {shared baseline\\\quad per-agent advantage};

  \node[note, anchor=north, align=center, font=\scriptsize]
    at (1.5, -1.55)
    {\textbf{Context-Folding}\\ \textbf{Dr.\,MAS} (stability fix)};
\end{scope}

\begin{scope}[shift={(11.0, 0)}]
  \node[title] at (1.5, 2.25) {(C) Decoupled, central critic};

  \node[orch] (oC)   at (1.5, 1.3)  {orch $\pi_\theta$};
  \node[sub]  (sC1) at (0.3, 0.1)  {sub 1};
  \node[sub]  (sC2) at (1.5, 0.1)  {sub 2};
  \node[sub]  (sC3) at (2.7, 0.1)  {sub 3};

  \draw[data] (oC.south west) -- (sC1.north east);
  \draw[data] (oC.south) -- (sC2.north);
  \draw[data] (oC.south east) -- (sC3.north west);

  \node[rectangle, rounded corners=2pt, draw=green!55!black, thick,
        fill=green!12, minimum width=1.5cm, minimum height=0.6cm,
        font=\scriptsize] (critic) at (1.5, -0.6) {central critic $V_\phi$};

  \draw[grad] (critic.west) -- ++(-1.1, 0) |- (sC1.west);
  \draw[grad] (critic.north) -- (oC.south);
  \draw[grad] (critic.east) -- ++(1.1, 0) |- (sC3.east);

  \node[note, anchor=north, align=center, font=\scriptsize]
    at (1.5, -1.3)
    {\textbf{M-GRPO} (hier.)\\ \textbf{MATPO} (single-LLM dual-role)};
\end{scope}

\end{tikzpicture}%
}
\caption{Three orchestrator training regimes
(\S\ref{sec:orch:regimes}). \textbf{(A)} Frozen sub-agents:
gradient flows only into the orchestrator; cheapest and most
common. \textbf{(B)} Joint training with shared baseline and
per-agent advantage: all policies update together; requires
stabilization (Dr.\,MAS's agent-wise normalization). \textbf{(C)}
Fully decoupled per-policy training against a central critic
$V_\phi$: most expressive but most engineering-heavy. Solid red
arrows show gradient flow; dashed outlines indicate frozen
components.}
\label{fig:training-regimes}
\end{figure}
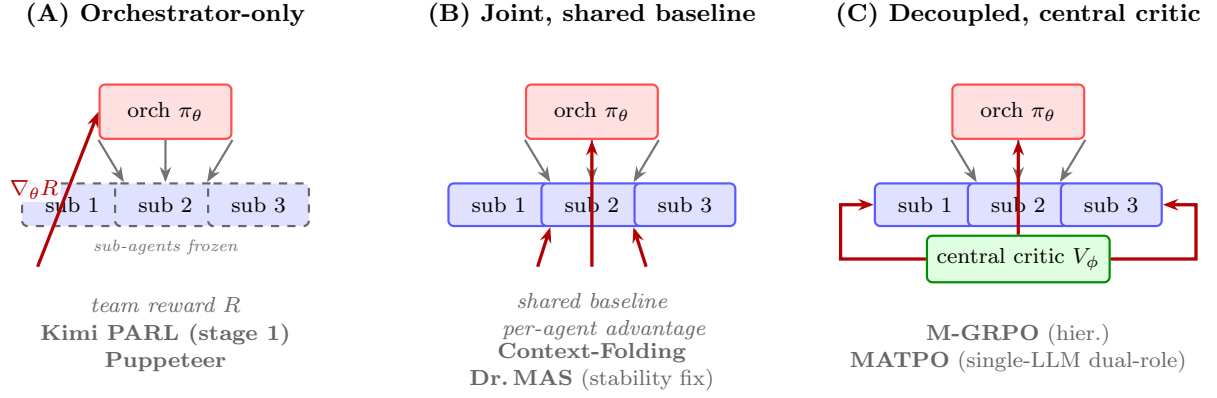

Three regimes appear repeatedly; Figure~\ref{fig:training-regimes}
shows the gradient-flow pattern of each.

\begin{itemize}
  \item \textbf{(A) Frozen sub-agents, train only the orchestrator.}
    Cheapest and the most common in practice. Kimi~PARL's first
    stage~\cite{kimi-k2-5-2026},
    Puppeteer~\cite{puppeteer2025}, and Agent Q-Mix's CTDE-style
    topology learner~\cite{agent-qmix2026} all fit this family in
    different ways. It avoids
    joint-training instability and is the ``credit-assignment
    safest'' choice because only the orchestrator's policy gradient
    flows.
  \item \textbf{(B) Joint training with shared baseline and
    per-agent advantage.} Context-Folding~\cite{context-folding2025}
    is the clearest example: orchestrator and sub-agents are trained
    together, share a team baseline, but compute advantage
    independently. Dr.\,MAS~\cite{dr-mas2026} is the stability
    analysis of why na\"ive joint training under GRPO fails and how
    agent-wise normalization fixes it. WideSeek-R1~\cite{wideseek-r1-2026}
    and MARTI-MARS$^2$~\cite{marti-mars2-2026} add newer examples
    where lead/sub-agent or heterogeneous multi-agent policies are
    optimized together.
  \item \textbf{(C) Fully decoupled per-agent training with a
    central critic.} M-GRPO~\cite{m-grpo2025} is the canonical
    hierarchical instance: top-layer and bottom-layer receive
    separate advantage signals. MATPO~\cite{matpo2025} implements a
    single-LLM analog where the planner and worker share model
    weights but receive role-specific advantages. DEPART~\cite{depart2026}
    alternates planner and executor optimization under dense
    role-specific and sparse task rewards; SPIRAL~\cite{spiral2026}
    and MARSHAL~\cite{marshal2026} show the same decoupling pressure
    in self-play settings.
\end{itemize}

Regimes (A)--(C) span a trade-off between training cost and
expressivity. (A) is cheap but cannot update sub-agent skill; (B)
updates everyone but is unstable without explicit normalization;
(C) is expressive but requires separate replay buffers per role and
is the most engineering-heavy.

\subsection{Engineering: rollout topology and asynchrony}
\label{sec:orch:engineering}

Multi-agent rollouts are cost-dominated by two factors: the slowest
sub-agent and the inter-agent dependency graph. Four engineering
techniques recur.

\begin{itemize}
  \item \textbf{Pipeline parallelism.} MarsRL~\cite{marsrl2025}
    arranges reasoning agents into a pipeline so that different
    stages of different rollouts execute concurrently, amortizing
    rollout cost. WideSeek-R1~\cite{wideseek-r1-2026} and
    MARTI-MARS$^2$~\cite{marti-mars2-2026} add complementary width
    and self-search scaling evidence.
  \item \textbf{Execution--training decoupling.}
    Agent~Lightning~\cite{agent-lightning2025} separates agent
    execution from the trainer; rollouts are produced by an
    inference harness and consumed asynchronously by the trainer.
    This matches the industrial harness boundary discussed in
    \S\ref{sec:systems:codex}.
  \item \textbf{Variable-shape replay buffer.} Orchestration traces
    have variable $|V|$, variable branching, and variable depth. No
    retained entry treats this as a first-class problem; most
    RL-oriented entries in our pool pad or truncate. This is a concrete open engineering
    problem (\S\ref{sec:open}).
  \item \textbf{Reward normalization across trace shapes.}
    Dr.\,MAS's agent-wise normalization~\cite{dr-mas2026} addresses
    this at the advantage level; a trace-level analogue (normalize
    over graph depth / branching factor) is still missing.
\end{itemize}

\subsection{Failure modes of orchestration learning}
\label{sec:orch:failures}

Five failure modes recur when orchestrators are trained directly;
each maps to one of O1--O5.

\begin{itemize}
  \item \textbf{Serial collapse (O1).} As auxiliary orchestration
    rewards decay in Kimi PARL, a naive orchestrator can collapse to
    never spawning, regressing to a single-agent baseline. Mitigated
    by staged annealing rather than an abrupt removal.
  \item \textbf{One-dominant-agent collapse (O2).} Under shared
    reward, the orchestrator routes nearly all delegations to a
    single sub-agent that happens to be slightly above-average.
    Population diversity can collapse. We found no general fix in our
    retained pool; debate-style
    topologies~\cite{debate-as-reward2026} partly sidestep by making
    diversity part of the reward.
  \item \textbf{Over-communication (O3).} Orchestrators inflate
    message volume to farm message-level process rewards. Observed
    whenever the judge scores per-message.
  \item \textbf{Aggregation leakage (O4).} Summary-return content is
    copied verbatim into the main trace, inflating apparent
    progress without real information gain; filtered by
    Context-Folding's scope-adjustment reward.
  \item \textbf{Train--inference topology mismatch.} Policies are
    trained at $k$-agent teams but deployed at $k'$-agent teams
    (Kimi K2.5 discloses a $100$-sub-agent trained-orchestrator
    regime; K2.6 reports a $300$-sub-agent deployment envelope).
    Generalization across team size is under-studied and is an open
    problem (\S\ref{sec:open}).
  \item \textbf{Adaptive deliberation outside the standard hierarchy.}
    Learning to Deliberate~\cite{learning-to-deliberate2025} introduces
    decentralized meta-cognitive actions such as Persist, Refine, and
    Concede. These are not ordinary messages and are not purely central
    orchestrator decisions; they sit between turn-level credit and
    orchestration-policy credit. We keep them in the turn/orchestration
    region of the taxonomy, but they are evidence that future versions
    of the hierarchy may need an explicit meta-policy layer.
\end{itemize}

\takeaway{The trajectory-to-trace shift is not cosmetic: it changes
the optimization target, the training regime choice (A/B/C), and
the engineering stack (pipeline parallelism, decoupled harness,
variable-shape replay). Recent width-scaling, topology-learning, and
self-play entries fill in O2/O3 and role/turn credit, but do not close
the stopping cell. Five sub-decisions (O1--O5) enumerate what
a learnable orchestrator actually does; no single paper in our pool
covers all five, and O5---when to stop---has no entry in our pool
that trains it explicitly.}

\section{Benchmarks and Evaluation}
\label{sec:benchmarks}

A recurring pattern in our paper pool is that LLM-MAS methods
report gains on benchmarks designed for single-agent evaluation.
This can be methodologically hazardous: single-agent benchmarks measure
task success, which any system with enough compute can improve;
they do not measure whether the improvement came from genuine
multi-agent coordination. This section uses an expanded evaluation
surface to audit the current benchmark landscape
(\S\ref{sec:bench:dims}--\S\ref{sec:bench:landscape}) and then gives
benchmark-design recommendations derived from the observed gaps
(\S\ref{sec:bench:good}).

\subsection{Four dimensions for auditing multi-agent evaluation}
\label{sec:bench:dims}

We use four dimensions to audit whether an LLM-MAS benchmark measures
coordination rather than task success alone.

\begin{itemize}
  \item \textbf{(E1) Task success / accuracy.} The standard metric.
    Necessary but not sufficient.
  \item \textbf{(E2) Parallelism efficiency.} Wall-clock speedup
    over a serial baseline; agent utilization (fraction of sub-agents
    doing task-relevant work); Critical-Steps-style metrics that
    distinguish real parallel progress from padded traces
    (\S\ref{sec:rewards:parl}).
  \item \textbf{(E3) Collaboration quality.} Message redundancy,
    consensus quality, debate diversity, and---in the specific case
    of debate-like topologies---whether resolution is reached.
    LatentMAS~\cite{latentmas2025} is the clearest evidence that
    much of (E3) can be achieved without natural-language
    messaging at all.
  \item \textbf{(E4) Protocol overhead.} Token cost per delegation,
    error-amplification ratio (how a single bad message propagates
    through the trace), and safety-related properties such as
    prompt-injection flow.
\end{itemize}

A benchmark is \emph{MAS-native} when it reports at least three of
(E1)--(E4). By this criterion, most benchmarks in our pool are not
MAS-native---they report (E1) only.

\subsection{Benchmark landscape by domain}
\label{sec:bench:landscape}

Table~\ref{tab:benchmarks} organizes the benchmarks referenced in
the pool by domain and MAS-nativeness.

\begin{table}[htbp]
\centering
\small
\renewcommand{\arraystretch}{1.25}
\begin{tabularx}{\linewidth}{@{}p{2.3cm} X p{1.8cm} X@{}}
\toprule
\textbf{Domain} & \textbf{Benchmarks (cited examples)} & \textbf{MAS-native?} & \textbf{Measures (E1--E4)?} \\
\midrule
Coding
 & SWE-Bench~\cite{swe-bench2024}, ArtifactsBench~\cite{artifactsbench2025}, CodeCriticBench~\cite{codecriticbench2025}
 & No
 & E1 only \\
Web / browser
 & WebArena~\cite{webarena2024}, BrowseComp~\cite{browsecomp2025}
 & No
 & E1 only \\
Research / search
 & GAIA~\cite{gaia2023}
 & No
 & E1; occasional E4 (token cost) \\
Tool use
 & ToolBench~\cite{toolbench2023}, $\tau$-bench~\cite{tau-bench2024}, MTU-Bench~\cite{mtu-bench2025}
 & Partial
 & E1 + partial E4 (tool success) \\
Long-horizon OS
 & OSWorld~\cite{osworld2024}
 & No
 & E1 + wall-clock (partial E2) \\
MAS-oriented
 & MultiAgentBench~\cite{multiagentbench2025}, TAMAS~\cite{tamas2025}; reported internal: Kimi Swarm Bench~\cite{kimi-k2-5-2026}
 & Partial / closed
 & Open entries cover subsets; Kimi Swarm Bench is not coded as open evidence for E1--E4. \\
\bottomrule
\end{tabularx}
\caption{Benchmark landscape for LLM-MAS evaluation, restricted
to benchmarks with arXiv-cited entries. Almost all domain
benchmarks report task success (E1) only; among MAS-native
benchmarks in our pool, none covers all four dimensions
(E1--E4) jointly.}
\label{tab:benchmarks}
\end{table}

Two observations.

\begin{itemize}
  \item \textbf{No benchmark in our pool covers all four
    dimensions in an open, auditable way.} Kimi Swarm
    Bench~\cite{kimi-k2-5-2026} is treated as a reported internal
    benchmark, not as open evidence for E1--E4, because it is closed
    and unauditable.
    TAMAS~\cite{tamas2025} covers (E4) safety specifically and
    nothing else. MultiAgentBench~\cite{multiagentbench2025}
    reports (E1) plus partial (E2)/(E3) depending on the task
    instance.
  \item \textbf{Cross-method comparability is limited.} Because
    credit-assignment papers (\S\ref{sec:credit}) each pick a
    different benchmark to evaluate on---C3 on math collaboration,
    SHARP on tool-augmented tasks, M-GRPO on deep research---direct
    comparison of their credit-assignment mechanisms is not
    currently possible.
\end{itemize}

\subsection{The benchmark gap}
\label{sec:bench:gap}

We argue that the shortage of MAS-native benchmarks is not merely
inconvenient; it actively shapes which algorithms succeed. Three
concrete consequences.

\begin{itemize}
  \item \textbf{E1-only benchmarks reward compute, not coordination.}
    A method that improves task success by spawning more sub-agents
    and trying them in parallel is indistinguishable on (E1) from
    a method that improves success by better credit assignment.
    The former is a more-compute scaling story; the latter is the
    claim this survey is organized around. Without (E2) this
    confound cannot be resolved.
  \item \textbf{Safety signals are underrepresented.} Only
    TAMAS~\cite{tamas2025} reports adversarial robustness
    systematically. Inter-agent prompt injection, shared-memory
    poisoning, and tool-parameter escalation are present in
    deployed systems but absent from most eval suites.
  \item \textbf{Kimi-reported long traces are absent from open
    benchmarks.} Kimi reports traces reaching $4{,}000$ steps in
    K2.6~\cite{kimi-k2-6-2026}; no open benchmark evaluates at that
    trace length. The credit-diffusion failure mode
    (\S\ref{sec:credit:fails}) is correspondingly invisible to
    academic evaluation.
\end{itemize}

\subsection{What a good MAS-native benchmark would look like}
\label{sec:bench:good}

The question is not only which existing benchmark to extend, but what
design properties would close the gaps observed above. We sketch
five recommendations derived from the gaps in
\S\ref{sec:bench:gap} and from the engineering constraints of
\S\ref{sec:engineering}.

\textbf{B1. Dimensional completeness.} A MAS-native benchmark should report
all four of (E1)--(E4): task accuracy, parallelism efficiency,
collaboration quality, and protocol overhead. Without this, the
compute-vs-coordination confound (\S\ref{sec:bench:gap}) cannot
be resolved on a per-instance basis.

\begin{table}[t]
\centering
\small
\renewcommand{\arraystretch}{1.15}
\begin{tabularx}{\linewidth}{@{}p{3.1cm}p{3.3cm}X@{}}
\toprule
\textbf{Metric} & \textbf{Operational definition} & \textbf{What it distinguishes} \\
\midrule
Parallelism efficiency
 & $T_{\text{serial}} / T_{\text{parallel}}$ or wall-clock serial baseline divided by MAS wall-clock
 & Real coordination speedup vs. merely spending more inference. \\
Useful-agent utilization
 & task-relevant sub-agent actions divided by total sub-agent actions
 & Productive decomposition vs. idle or redundant spawned agents. \\
Protocol overhead
 & orchestration, message, and tool-management tokens divided by total tokens
 & Coordination cost vs. task-solving content. \\
Message redundancy
 & semantically duplicate messages divided by total inter-agent messages
 & Useful communication vs. verbosity or padding. \\
Error amplification ratio
 & downstream corrupted events divided by the initial corrupted event
 & Whether one bad tool/message contaminates the trace. \\
\bottomrule
\end{tabularx}
\caption{Operational metric definitions for MAS-native evaluation.
Exact implementations will vary by benchmark, but reporting these
quantities would make E1--E4 comparable across methods.}
\label{tab:benchmark-metric-defs}
\end{table}

\textbf{B2. Trace-length stratification.} Tasks should be grouped by
expected trace length so that performance can be reported per
$T \in \{10^1, 10^2, 10^3\}$ band, exposing the credit-diffusion
behavior of \S\ref{sec:eng:trace}. A method that excels at $T=50$
but degrades at $T=500$ is a qualitatively different beast from
one that scales gracefully; current single-number benchmarks hide
this distinction.

\textbf{B3. Topology variability.} Ideally, the same underlying task is
instrumented for multiple topologies from
Table~\ref{tab:topologies}---centralized, debate, swarm,
hierarchical---so that orchestration choices can be ablated. The
goal is not to crown a single best topology but to expose how
much of the gain attributed to a credit-assignment method is
actually attributable to its preferred topology.

\textbf{B4. Adversarial control conditions.} Each task should ship with
controlled adversarial perturbations from the attack vectors of
\S\ref{sec:safety:taxonomy}: indirect prompt injection in tool
output (AV2), inter-agent message pollution (AV3), shared-memory
poisoning (AV4). A method's robustness margin under these
perturbations becomes a first-class metric, not an afterthought.

\textbf{B5. Open data, public leaderboard, replayable traces.}
Instances and orchestration traces are released; runs are
reproducible; scores are reported per-band rather than as a
single average. This addresses the cross-method comparability
problem (\S\ref{sec:bench:gap}): publication of full
orchestration traces lets follow-up work re-evaluate without
re-running rollouts.

\textbf{Minimal trace reporting schema.} To make B1--B5 operational,
an evaluation should log the orchestration trace as a typed event graph
rather than only a final answer and score. The following schema is
minimal: it is not a benchmark proposal by itself, but it is the
smallest artifact that would let later work recompute reward, credit,
parallelism, and safety metrics over the same rollout. The same
structure is provided as a machine-readable JSON Schema in the
artifact repository, together with a minimal valid example trace.
The accompanying Python validator is a lightweight structural
checker for required fields, event types, edge references, duplicate
event identifiers, and non-negative costs; it is not a full
implementation of the JSON Schema standard.

\begin{verbatim}
{
  "trace_id": "...",
  "task_id": "...",
  "events": [
    {"id": "e1", "t": 0, "type": "spawn",
     "agent": "orchestrator", "role": "planner"},
    {"id": "e2", "t": 1, "type": "message",
     "agent": "planner", "from": "planner", "to": "executor"},
    {"id": "e3", "t": 2, "type": "tool_call",
     "agent": "executor", "tool": "browser"}
  ],
  "edges": [{"src": "e1", "dst": "e2", "type": "causal"}],
  "rewards": {"team": 1.0, "orchestration": 0.3, "tool": 0.1},
  "costs": {"tokens": 12000, "wall_clock_s": 420}
}
\end{verbatim}

\begin{table}[t]
\centering
\small
\renewcommand{\arraystretch}{1.12}
\begin{tabularx}{\linewidth}{@{}p{3.0cm}X@{}}
\toprule
\textbf{Reporting item} & \textbf{Minimum information} \\
\midrule
Topology and roles
 & Active roles, spawn/despawn events, fixed vs.\ dynamic team size. \\
Trace scale
 & Number of events, messages, tool calls, sub-agents, wall-clock time, and token cost. \\
Reward channels
 & Team reward plus any process, tool, verifier, or orchestration rewards and whether auxiliary terms are annealed. \\
Credit unit
 & Finest unit receiving an advantage, value estimate, counterfactual score, Shapley score, or learned critic signal. \\
Safety instrumentation
 & Whether untrusted tool output, inter-agent messages, shared memory, and human interventions are separately logged. \\
\bottomrule
\end{tabularx}
\caption{Minimal reporting checklist for MAS-native evaluation. This
checklist is intended to make trace-level claims auditable without
requiring a new benchmark suite.}
\label{tab:trace-reporting-checklist}
\end{table}

The closest non-open reference point is the internally reported Kimi
Swarm Bench~\cite{kimi-k2-5-2026}; we treat it as context, not as an
auditable benchmark row for the E1--E4 gap. The closest open
approximation is MultiAgentBench~\cite{multiagentbench2025}, which
covers a subset of (E1)--(E3) at small $T$. A
benchmark satisfying B1--B5 jointly would directly address the main
comparability bottleneck in this survey: without it, new
credit-assignment methods cannot be evaluated against a shared
coordination-sensitive target.

\takeaway{Within our retained pool as of May 4, 2026, we found no
single open benchmark that
reports (E1)--(E4) jointly at the Kimi-reported long-trace envelope.
This is the
most tractable near-term infrastructure gap: credit-assignment
methods cannot be fairly compared until evaluation measures more
than task success.}

\section{Safety and Adversarial Robustness in LLM-MAS}
\label{sec:safety}

The benchmark gap (\S\ref{sec:bench:gap}) intersects with a second
underdeveloped axis: adversarial robustness. Single-agent agentic
safety is itself unsolved~\cite{greshake2023,agentdojo2024}, but
LLM-MAS introduces failure modes whose multi-agent propagation
patterns have no close single-agent analogue and only a handful of
dedicated benchmarks in our pool. This section
keeps the focus narrow: attacks are organized by the same
credit- and signal-bearing units used in \S\ref{sec:credit:hierarchy}, because
the levels at which credit is assigned are also the levels at which
adversarial influence can enter and propagate.

\subsection{Threat model for LLM-MAS}
\label{sec:safety:threat}

Three properties introduce attack surfaces beyond those of a
single tool-using agent. First, the attack surface scales with team
size: every spawned sub-agent inherits tool access, every message is a
potential injection point, and every shared-memory write is a
potential poison; the number of inter-agent \emph{flows} grows
super-linearly in the number of nodes of the orchestration trace
(\S\ref{sec:credit:hierarchy}). Second, information flows between
LLMs that each treat the other's output as trusted natural
language; a compromised tool output that passes through one
sub-agent's summary becomes an instruction to the next sub-agent or
to the orchestrator~\cite{greshake2023,injecagent2024}. Third,
dynamic-spawn systems (Kimi PARL, AgentSpawn~\cite{agentspawn2026})
create sub-agents at runtime whose isolation cannot be audited in
advance. The May 2026 refresh also adds a training-time safety method:
MAGIC formulates attacker and defender LLMs as a co-evolving
multi-agent RL game~\cite{magic2026}, which is useful evidence for
adversarial safety training even though it does not by itself solve
trace-level constrained optimization for deployed swarms.

We distinguish three threat actors. \emph{(i) External user input}
is the classical jailbreak/prompt-injection channel. \emph{(ii)
Untrusted tool output} is the channel exploited by indirect prompt
injection: a web page, an email body, or a retrieved document
contains adversarial text that is treated as instructions when it
re-enters the LLM context~\cite{greshake2023,agentdojo2024,wasp2025}.
\emph{(iii) Adversarial agent in the team} is novel to MAS:
either a member sub-agent has been compromised at spawn time, or a
team-internal message has been poisoned at runtime, after which
the contagion propagates through shared
memory~\cite{tmcht2025,agents-under-siege2025,tamas2025}. Classical
MARL safety---reward hacking, shielding, safe exploration---is
necessary but not sufficient: it does not address natural-language
information flow between LLMs, which is where most LLM-MAS attacks
land in our pool.

\subsection{Attack taxonomy}
\label{sec:safety:taxonomy}

Table~\ref{tab:attacks} organizes attack vectors against the
credit- and signal-bearing units of \S\ref{sec:credit:hierarchy}. The mapping
is not coincidental: the same structural levels at which credit
must be assigned are the levels at which an attacker can pivot.

\begin{table}[htbp]
\centering
\small
\renewcommand{\arraystretch}{1.25}
\begin{tabularx}{\linewidth}{@{}p{3.2cm} p{1.9cm} p{2.2cm} X@{}}
\toprule
\textbf{Attack vector} & \textbf{Threat actor} & \textbf{Affected level} & \textbf{Representative attack / benchmark; defense status} \\
\midrule
Direct prompt injection
 & user
 & token / turn
 & Long-studied single-agent threat; AgentDojo~\cite{agentdojo2024}
   covers within tool-use loops; defenses are input-side filters. \\
Indirect prompt injection
 & tool output
 & tool / message
 & Greshake et al.~\cite{greshake2023}, InjecAgent~\cite{injecagent2024},
   WASP~\cite{wasp2025}; defenses are tool-output sanitization
   and provenance tracking, both partial. \\
Inter-agent message pollution
 & adversarial agent
 & message
 & Agents Under Siege~\cite{agents-under-siege2025} optimizes
   permutation-invariant attacks across topologies; no published
   defense in our pool. \\
Shared-memory poisoning
 & adversarial agent
 & agent / team
 & TMCHT~\cite{tmcht2025} demonstrates contagious jailbreak via
   memory writes that survive across turns and propagate to
   honest agents. \\
Tool-parameter escalation
 & user / tool output
 & tool
 & A compromised sub-agent issues tool calls outside its declared
   capability set; mitigated heuristically by per-sub-agent tool
   allowlists~\cite{claude-code-subagents-2025}. \\
Orchestrator misdirection
 & user / message
 & orchestrator
 & A poisoned summary causes the orchestrator to spawn the wrong
   sub-agent or route the wrong task; observed in
   TAMAS~\cite{tamas2025}; no published RL-time defense. \\
Reward hacking across agents
 & policy itself
 & agent / role
 & Pseudo-parallelism, communication padding, verifier collusion
   (\S\ref{sec:rewards:hacking}); MAGIC~\cite{magic2026} trains an
   attacker--defender game for safety alignment, but trace-level
   constrained optimization remains open. \\
Sandbox escape under spawn
 & adversarial agent
 & system
 & Per-sub-agent isolation in dynamic-spawn systems is not
   publicly specified; flagged as P14 in \S\ref{sec:open:safety}. \\
\bottomrule
\end{tabularx}
\caption{Attack vectors against LLM-MAS, organized by the
credit- or signal-bearing level at which the attack lands
(\S\ref{sec:credit:hierarchy}). Most published attacks target
\emph{message} or \emph{tool} levels; \emph{orchestrator} and
\emph{role} attacks are barely studied in our pool, and
\emph{sandbox/system} attacks are not publicly documented at all.}
\label{tab:attacks}
\end{table}

\begin{figure}[t]
\centering
\begin{tikzpicture}[font=\footnotesize, scale=0.92, transform shape,
  orch/.style={rectangle, rounded corners=2pt,
               draw=red!70, very thick, fill=red!12,
               minimum width=1.6cm, minimum height=0.7cm,
               align=center, font=\scriptsize\bfseries},
  sub/.style={rectangle, rounded corners=2pt,
              draw=blue!65, thick, fill=blue!10,
              minimum width=1.4cm, minimum height=0.6cm,
              align=center, font=\scriptsize},
  tool/.style={rectangle, draw=orange!85, thick, fill=orange!15,
               minimum width=1.1cm, minimum height=0.5cm,
               align=center, font=\scriptsize},
  mem/.style={rectangle, rounded corners=2pt,
              draw=purple!70, thick, fill=purple!12,
              minimum width=1.0cm, minimum height=0.7cm,
              align=center, font=\scriptsize, inner sep=2pt},
  external/.style={rectangle, rounded corners=8pt,
                   draw=black!50, thick, fill=gray!10,
                   minimum width=1.5cm, minimum height=0.6cm,
                   font=\scriptsize, align=center},
  msg/.style={-{Stealth[length=1.6mm]}, thick, black!55},
  attack/.style={font=\scriptsize\bfseries, red!80!black, align=left,
                 inner sep=1.5pt},
  alabel/.style={fill=red!15, draw=red!60, rounded corners=1pt,
                 inner sep=1.4pt, font=\tiny\bfseries,
                 text=red!75!black, align=center,
                 text width=1.25cm}
]

\node[external] (user) at (-0.3, 3.0) {user task};
\node[orch] (oA) at (3.5, 3.0) {orch};
\node[sub] (s1) at (1.6, 1.6) {sub-agent\\A};
\node[sub] (s2) at (3.5, 1.6) {sub-agent\\B};
\node[sub] (s3) at (5.4, 1.6) {sub-agent\\C};
\node[tool] (t1) at (1.0, 0.1) {web API};
\node[tool] (t2) at (3.5, 0.1) {code exec};
\node[tool] (t3) at (5.4, 0.1) {MCP tool};
\node[mem] (mem) at (8.5, 1.6) {shared\\memory};

\draw[msg] (user) -- node[above=-1pt, font=\tiny\itshape, black!60]{prompt} (oA);
\draw[msg] (oA) -- (s1);
\draw[msg] (oA) -- (s2);
\draw[msg] (oA) -- (s3);
\draw[msg] (s1) -- (t1);
\draw[msg] (s2) -- (t2);
\draw[msg] (s3) -- (t3);
\draw[msg, <->] (s2) -- (mem);
\draw[msg, <->] (s3) -- (mem);
\draw[msg, dashed, black!50] (s1) -- node[above=-1pt, font=\tiny\itshape, black!60]{summary} (s2);
\draw[msg, dashed, black!50] (s2) -- (s3);

\node[alabel] (av1) at (-0.3, 4.0) {AV1\\direct prompt inj.};
\draw[red!60, thick, ->, >=stealth] (av1.south) -- (user.north);

\node[alabel] (av2) at (0.75, -1.15) {AV2\\indirect prompt inj.};
\draw[red!60, thick, ->, >=stealth] (av2.north) -- (t1.south);

\node[alabel] (av3) at (2.2, 0.58) {AV3\\msg pollution};
\draw[red!60, thick, ->, >=stealth] (av3.east) to[bend left=10] ($(s1.east)+(0.1,0.0)$);

\node[alabel] (av4) at (8.5, 3.5) {AV4\\memory poisoning};
\draw[red!60, thick, ->, >=stealth] (av4.south) -- (mem.north);

\node[alabel] (av5) at (6.1, -1.15) {AV5\\tool param escalation};
\draw[red!60, thick, ->, >=stealth] (av5.north) -- (t3.south);

\node[alabel] (av6) at (3.5, 4.3) {AV6\\orch.\ misdirection};
\draw[red!60, thick, ->, >=stealth] (av6.south) -- (oA.north);

\node[alabel] (av7) at (7.45, 0.65) {AV7\\reward-hack collusion};
\draw[red!60, thick, ->, >=stealth, dashed] (av7.west) to[bend right=20] (s3.east);

\node[alabel] (av8) at (3.45, -1.15) {AV8\\sandbox escape};
\draw[red!60, thick, ->, >=stealth] (av8.north) -- (t2.south);

\node[draw=black!40, rounded corners=2pt, fill=gray!8,
      text width=4.2cm, font=\scriptsize, align=left,
      anchor=north west]
  at (10.3, 4.3)
  {\textbf{Credit-level mapping}\\
   (cross-ref Fig.~\ref{fig:credit-hierarchy}):\\[2pt]
   $\bullet$ AV1, AV6 $\to$ \textit{orch.\ level}\\
   $\bullet$ AV3 $\to$ \textit{message level}\\
   $\bullet$ AV4 $\to$ \textit{role / agent level}\\
   $\bullet$ AV2, AV5, AV8 $\to$ \textit{tool level}\\
   $\bullet$ AV7 $\to$ \textit{team level}};

\end{tikzpicture}
\caption{Attack-surface map for an LLM-MAS orchestration trace.
The substrate (orchestrator $\to$ sub-agents $\to$ tools $\to$ shared
memory) is the same as Fig.~\ref{fig:orchestration-trace};
red labels mark eight attack vectors (AV1--AV8) cataloged in
Table~\ref{tab:attacks}, each anchored to its locus of action.
Right legend shows which credit-bearing unit
(\S\ref{sec:credit:hierarchy}) is the locus of attack: this
mapping is what makes credit-assignment progress and security
progress mutually informative.
AV1--AV2 derive from single-agent indirect prompt
injection~\cite{greshake2023,injecagent2024,agentdojo2024};
AV3--AV4 are MAS-specific~\cite{tmcht2025,agents-under-siege2025};
AV5/AV8 are tool-runtime concerns; AV6/AV7 emerge only in
orchestrated systems and remain under-benchmarked.}
\label{fig:attack-surface}
\end{figure}
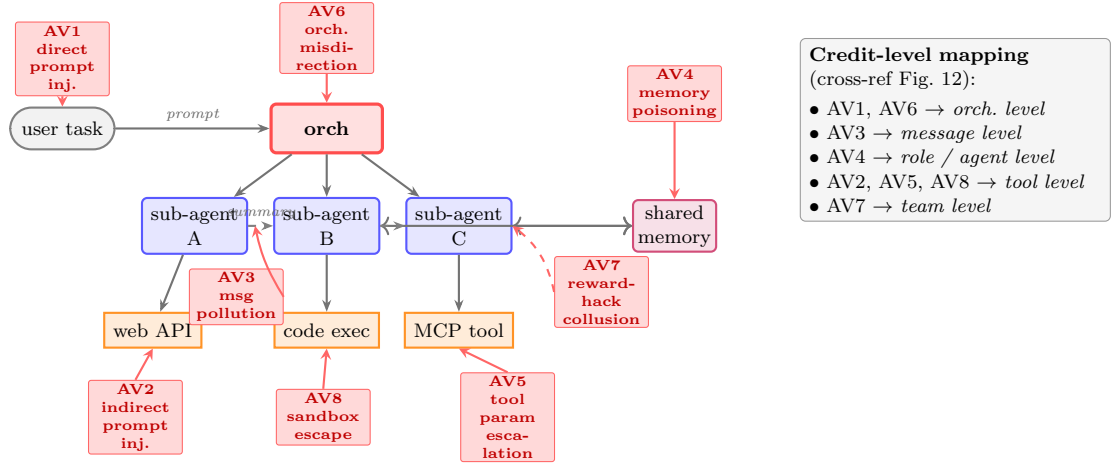

\subsection{Defense landscape and benchmarks}
\label{sec:safety:defense}

Inference-time defenses in our pool are mostly \emph{ad hoc mitigations}: input and
tool-output sanitization~\cite{agentdojo2024,greshake2023},
tool-parameter allowlists~\cite{claude-code-subagents-2025},
per-task or per-sub-agent sandboxing~\cite{openai-codex2025,claude-code-subagents-2025},
reward-model verification, heterogeneous verifiers, and
organizational controls from trustworthy-agent frameworks
(\eg human checkpoints and scoped credentials)~\cite{anthropic-trustworthy2025}.
MAGIC~\cite{magic2026} is the clearest retained training-time
counterexample: it uses co-evolving attacker and defender agents to
manufacture adversarial safety data and optimize the defender.
Within our pool, we found no trace-level constrained-optimization
formulation comparable to constrained MDPs or shielded RL. This is
the safety-side analogue of the credit-assignment gap: the defense
must decide which orchestrator, role, message, or tool event should
be constrained, edited, or blamed.

The benchmark situation is similarly sparse. TAMAS~\cite{tamas2025}
is the most MAS-specific benchmark in our pool; AgentDojo,
InjecAgent, and WASP cover adjacent tool-use or web-agent injection
settings~\cite{agentdojo2024,injecagent2024,wasp2025}; and two ACL
2025 works move into multi-agent attack propagation through topology
and shared memory~\cite{agents-under-siege2025,tmcht2025}. None of
these benchmarks jointly reports safety with collaboration quality
(E3) and parallelism efficiency (E2), so they cannot yet support the
kind of reward--credit--safety comparison needed by this survey.

\subsection{The under-addressed problem: steerability}
\label{sec:safety:steer}

Anthropic's trustworthy-agents framework~\cite{anthropic-trustworthy2025}
isolates a property they call mid-trace \emph{steerability}: the
ability of a human supervisor to inspect a partially completed
orchestration trace, intervene at a specific point, and have the
intervention propagate sensibly forward. This is a
credit-assignment-shaped problem in disguise. To intervene
informatively the supervisor must answer the same questions the
RL trainer asks: which earlier orchestrator decision is responsible
for the current state, and which downstream sub-agent decisions will
be invalidated by changing it? An intervention at the wrong
decision is wasted (the system reverts) or destructive (downstream
sub-agents continue on stale assumptions). The hierarchy that
organizes credit (orchestrator, role, agent, message, tool;
\S\ref{sec:credit:hierarchy}) thus also organizes intervention
points, and the counterfactual machinery used by
C3~\cite{c3-2026} to attribute credit to messages could in principle
attribute the consequences of a human edit to messages. We are not
aware of a published RL formulation of steerability in our pool;
the closest existing work treats it as a UI/HCI question rather than
as an RL objective. This is the connection point to P13 in
\S\ref{sec:open:safety}.

\takeaway{LLM-MAS safety inherits all of single-agent agentic
safety and adds three structural threats whose propagation patterns
are multi-agent-specific: inter-agent message pollution,
shared-memory contagion, and orchestrator misdirection. The defense landscape is uniformly
ad hoc at inference time, while MAGIC shows that adversarial
attacker--defender RL is beginning to enter the training side. The
benchmark coverage remains shallow (TAMAS, AgentDojo,
InjecAgent, WASP, plus two ACL 2025 multi-agent attack papers),
and steerability---one operationally important safety
property---has not been formalized as an RL objective in any work
we are aware of.}

\section{Open Problems}
\label{sec:open}

We close the survey with fifteen open problems, organized along five
axes: algorithmic (P1--P5), reward (P6--P8), systems (P9--P11),
safety (P12--P14), and evaluation (P15). Each problem is stated
compactly with a pointer to where in the survey it was developed and
to the closest published work (if any) that addresses it.

\subsection{Algorithmic}
\label{sec:open:algo}

\textbf{P1. Credit diffusion under long traces.}
Terminal-only team reward over $10^3$--$10^4$ orchestration steps
can make the per-decision signal fragile or low-SNR
(\S\ref{sec:credit:fails}). Dr.\,MAS~\cite{dr-mas2026} addresses a
symptom (training instability) through agent-wise normalization, while
LangMARL~\cite{langmarl2026}, CoLLM-MAAC~\cite{collm-maac2026}, and
MARSHAL~\cite{marshal2026} add denser language, critic, or turn-level
signals. A principled account of how these signals scale to
Kimi-reported long traces is still missing in our retained pool.

\textbf{P2. Free-riding under shared reward.}
Under R1 shared reward, silent or near-silent sub-agents receive
equal credit. SHARP~\cite{sharp2026} offers Shapley-based marginal
credit; an open question is whether Shapley approximation remains
tractable at production team sizes
($n \gtrsim 100$, \S\ref{sec:systems:kimi}).

\textbf{P3. Coordination collapse and one-dominant-agent.}
Under joint training, population diversity often collapses
(\S\ref{sec:orch:failures}); the orchestrator routes most delegations
to a single sub-agent. Within our retained LLM-MAS RL pool, we found
no method that rewards agent diversity directly across dynamic swarms.
MARTI-MARS$^2$~\cite{marti-mars2-2026} shows policy-diversity gains
from heterogeneous self-search training, but diversity is not yet a
general-purpose orchestration objective.

\textbf{P4. Counterfactual credit over unrealized branches.}
The orchestrator's policy includes ``do not spawn.'' No realized
trace exists to attribute credit against. C3~\cite{c3-2026} handles
message-level counterfactuals \emph{within} a realized trace, not
across realized/unrealized alternatives. Off-policy evaluation of
unrealized branches is an open direction.

\textbf{P5. Train--inference topology mismatch.}
Methods are trained at $k$ agents but deployed at $k'$: Kimi K2.5
discloses a trained-orchestrator regime up to $100$ sub-agents, while
K2.6 reports a deployment envelope up to $300$ sub-agents~\cite{kimi-k2-5-2026,kimi-k2-6-2026}.
Whether a trained orchestrator policy generalizes across team size,
and under what conditions, is under-studied.

\subsection{Reward}
\label{sec:open:reward}

\textbf{P6. Reward hacking in tool environments.}
Tool-spam, fabricated tool success, and padding
(\S\ref{sec:rewards:hacking}). Agent~Lightning~\cite{agent-lightning2025}
and MATPO~\cite{matpo2025} condition tool reward on downstream
outcome; a general principle for pricing tool calls is absent in our
retained pool.

\textbf{P7. Verifier--policy collusion.}
When a verifier LLM is drawn from the same family as the policy,
both drift together and the verifier reward becomes uninformative.
We are not aware of a fix in our pool beyond using heterogeneous
verifier families, which is brittle.

\textbf{P8. Process--outcome reward balance.}
Dense PRM (R4) combined with sparse team outcome (R1) lets the
dense signal dominate gradients, causing policy drift toward what
the PRM rewards rather than what the task rewards. MALT~\cite{malt2025}
uses role-specific PRMs as a mitigation; a general principle is
missing in our retained pool.

\subsection{Systems and engineering}
\label{sec:open:systems}

\textbf{P9. Rollout cost dominance.}
Multi-agent rollouts are $10$--$100\times$ more expensive than
single-agent and dominate wall-clock RL time
(\S\ref{sec:orch:engineering}). Pipeline parallelism
(MarsRL~\cite{marsrl2025}) and execution--training decoupling
(Agent~Lightning~\cite{agent-lightning2025}) are partial answers;
further gains likely require hierarchical rollout scheduling.

\textbf{P10. Variable-shape replay.}
Orchestration traces have variable $|V|$, branching, and depth.
Standard replay buffers pad or truncate. A graph-native buffer and
a matching advantage normalization (cf.\ Dr.\,MAS~\cite{dr-mas2026}
at the agent level) are still missing in our retained pool.

\textbf{P11. Straggler-robust training.}
The slowest sub-agent gates the whole trace. Bias correction for
asynchronous rollouts---on-policy vs.\ near-on-policy---is not
addressed in any retained LLM-MAS method we found.

\subsection{Safety}
\label{sec:open:safety}

\textbf{P12. Inter-agent prompt injection.}
Untrusted tool output flows through the team; one compromised
message can pivot the orchestrator.
Foundational indirect prompt
injection~\cite{greshake2023,injecagent2024} already afflicts
single-agent settings; LLM-MAS compounds the problem by letting
the injection propagate across sub-agents as trusted summaries
(\S\ref{sec:safety}). TAMAS~\cite{tamas2025}, AgentDojo~\cite{agentdojo2024},
and WASP~\cite{wasp2025} establish benchmarks at different scales;
MAGIC~\cite{magic2026} adds a training-time attacker--defender RL
countermeasure, but deployed trace-level defenses remain mostly ad
hoc (\S\ref{sec:safety:defense}).

\textbf{P13. Mid-trace steerability.}
Anthropic's framework~\cite{anthropic-trustworthy2025} flags that
humans cannot easily intervene mid-orchestration. This is
credit-assignment-shaped: where in the trace can a human inject,
and what are the downstream consequences?
Multi-agent attacks that exploit this gap---including
permutation-invariant topology attacks~\cite{agents-under-siege2025}
and contagious memory poisoning~\cite{tmcht2025}---are now
documented; we found no paper in our pool that formalizes
steerability as an RL objective.

\textbf{P14. Sandbox-isolation under dynamic spawn.}
Each new sub-agent needs its own sandbox; failure modes scale with
team size. Dynamic-spawn systems such as Kimi PARL and
AgentSpawn~\cite{agentspawn2026} do not publicly discuss how
per-sub-agent isolation is guaranteed.

\subsection{Evaluation}
\label{sec:open:eval}

\textbf{P15. MAS-native benchmark at Kimi-reported trace lengths.}
As argued in \S\ref{sec:bench:gap}, we found no open benchmark in our
retained pool that covers (E1)--(E4) jointly at $\gtrsim 10^3$-step
traces. WideSeek-R1~\cite{wideseek-r1-2026} and
MARTI-MARS$^2$~\cite{marti-mars2-2026} make academic width/self-search
scaling more serious, but they do not replace an open trace-level
benchmark at industrial lengths. This is the single most
tractable infrastructure gap: without it, credit-assignment methods
cannot be fairly compared and scaling claims cannot be cross-validated.
Likewise, no explicit RL training method in the retained pool targets
the stopping decision as a learned O5 policy; this does not rule out
heuristic, budgeted-inference, or non-RL halting mechanisms.

\takeaway{Of the fifteen problems, P1 (credit diffusion), P4
(unrealized-branch counterfactuals), and P15 (MAS-native benchmarks)
are the most load-bearing: progress on them would unlock progress on
many of the others. P5, P10, and P13 are the most deployment-relevant
and the most under-published.}

\section{Limitations}
\label{sec:limitations}

This survey is intended as a curated taxonomy and position paper, not
as an exhaustive systematic review. Four limitations are therefore
important for interpreting its claims.

\textbf{Curated rather than exhaustive corpus.} The retained pool
contains $84$ entries selected for their relevance to reward design,
credit assignment, orchestration learning, systems constraints,
benchmarks, or safety in LLM-MAS RL. It is not a complete bibliography
of all LLM-MAS, agentic-RL, hierarchical-RL, or MARL work. The
inclusion protocol and screening-decision log make the curation
auditable at record level for this manuscript. They do not make the
review fully reproducible in the PRISMA sense because we do not provide
database-result exports, deduplication logs, or a multi-annotator
screening protocol.

\textbf{Single-author tagging.} The taxonomy tags in the retained-entry
CSV were assigned by manual reading of
abstracts, methods, and public artifacts. We did not run a blinded
multi-annotator protocol or report inter-annotator agreement. Borderline
entries---for example surveys with RL relevance, industrial systems
with undisclosed training details, and self-evolution frameworks that
do not optimize a conventional RL objective---should be read as
taxonomy judgements rather than objective labels.

\textbf{Industrial sources are not reproducible algorithms.} Kimi
K2.5 is the only industrial source in our pool that explicitly
discloses RL training of the orchestrator. Codex, Claude Code, Kimi
K2.6, and Anthropic engineering case studies are used for deployment
shape, scale, harness boundaries, and workflow pressure unless a public
source explicitly discloses the training objective. We do not infer
undisclosed multi-agent RL objectives from product behavior.

\textbf{Formal claims are organizing arguments.} The dynamic-Dec-POMDP
and orchestration-trace definitions fix the vocabulary used by the
survey. The two conceptual claims about credit diffusion and
non-identifiability are not theorems, and Figure~\ref{fig:trace-length}
is schematic rather than a fitted empirical law. A full theory would
need explicit assumptions on noise, baselines, graph dynamics,
off-policy branch coverage, and value-function approximation over
variable-shape traces.

\textbf{The literature is moving quickly.} The cutoff for this version
is May 4, 2026. New industrial reports, OpenReview submissions,
benchmarks, and safety evaluations can change both the coverage map and
the sparsity conclusions, especially in the message-credit,
orchestrator-credit, adaptive-deliberation, and trace-level safety
cells.

\section{Reproducibility and Artifact Statement}
\label{sec:artifact-statement}

The manuscript is accompanied by a supplementary artifact snapshot,
mirrored in the artifact repository (\artifactrepo) with
repository-path normalization (for example, repository
\texttt{scripts/} and \texttt{trace-schema/} paths correspond to the
manuscript-bundle \texttt{artifact/} paths), intended to make the
taxonomy auditable. The artifact contains four components.

\begin{itemize}
  \item \textbf{Corpus metadata.} A retained-entry CSV records the
    $84$ retained entries with 18 controlled fields. An exclusion-log
    CSV records the $32$ screened-but-excluded decisions, each with a
    public identifier or documentation handle and URL.
  \item \textbf{Scripted statistics.}
    A repository script regenerates the retained /
    excluded counts, controlled-field histograms, and cross-tabs used
    in Table~\ref{tab:coverage-statistics},
    Table~\ref{tab:reward-credit-crosstab}, and
    Table~\ref{tab:orch-credit-crosstab}.
  \item \textbf{Trace schema.} A machine-readable JSON Schema specifies
    typed orchestration traces, and a companion example provides a
    minimal valid trace.
  \item \textbf{Validation.} A dependency-free structural checker
    verifies
    required fields, event types, edge references, duplicate event
    identifiers, and non-negative cost fields for a trace JSON file.
    It is intended as a lightweight sanity check, not as a complete
    JSON Schema implementation.
\end{itemize}

These files do not make the literature review exhaustive, nor do they
replace a multi-annotator systematic review. They do make the central
claims of the paper mechanically inspectable: readers can check which
entries support a taxonomy cell, regenerate the sparsity counts, and
test whether a new benchmark log satisfies the minimal orchestration
trace schema.

\section{Conclusion}
\label{sec:conclusion}

We surveyed reinforcement learning and post-training for LLM-based
multi-agent systems as of May 4, 2026, organized around a single
thesis: the field is usefully analyzed through orchestration traces
rather than only through per-agent trajectories.
\S\ref{sec:formalism} formalized this
object as a working abstraction for taxonomy and auditability, using
an event graph drawn from a dynamic-Dec-POMDP extension
and stated two informal observations---credit diffusion under
uniform credit and non-iden\-tifi\-ability of orchestrator spawn
decisions---that organize the rest of the paper. We stress that
these observations are motivating arguments, not formal theorems;
tight rates and full proofs are deferred to follow-up work.

Three taxonomies operationalize the thesis. \S\ref{sec:rewards}
partitions the reward design space into eight families, with
orchestration reward (R7) identified as the family that most directly
targets spawn / delegate / aggregate decisions over multiple agent
instances, and with the defining property that its useful weight is
non-constant over training. \S\ref{sec:credit} organizes
entries along an eight-level credit- or signal-bearing-unit
hierarchy---team / orchestrator / role / agent / turn / message /
tool / token---and shows that explicit counterfactual message-level
credit remains especially sparse, while newer
agent-, role-, turn-, and orchestrator-level entries have started to
fill in the surrounding taxonomy cells. \S\ref{sec:orchestration}
decomposes orchestration learning into five sub-decisions (when to
spawn, whom to delegate to, how to communicate, how to aggregate,
when to stop), and finds that within our curated pool no method
explicitly trains the when-to-stop decision as an RL target.

The industrial--academic bridge is asymmetric. Kimi Agent Swarm
publicly trains an orchestrator (PARL); OpenAI Codex and Anthropic
Claude~Code publicly document their deployment shape (parallel
workflows, harness boundaries, dynamic spawn) but---to our
knowledge---not whether the orchestration itself is an RL training
target. Section~\ref{sec:engineering} identified three engineering
constraints---rollout cost scaling as
\(\sum_i (L_i c_{\text{tok}} + T_i c_{\text{tool}}) +
C_{\text{orch}}(K, |G|)\), the harness as a
training-frozen interface, and per-decision signal decay under long
traces---that together explain why academic methods mostly evaluated at
$T \lesssim 10^2$ cannot be assumed to transfer to the disclosed
Kimi-reported deployment envelope at $T \sim 10^3$--$10^4$. Even taking
only Kimi as the public trained-orchestrator anchor, the open literature is still typically evaluated
with fixed or moderate-size teams rather than hundreds of sub-agents,
although WideSeek-R1 and MARTI-MARS$^2$ now make academic width scaling
more concrete. Closing
this gap is less an algorithmic challenge than an infrastructural
one: variable-shape replay (P10), rollout cost (P9), and MAS-native
benchmarks at Kimi-like scale (P15) are engineering problems without
solutions in our pool.

Three directions follow most directly from this survey:

\begin{enumerate}
  \item \textbf{A unified credit-assignment formalism over
    orchestration graphs.} Table~\ref{tab:credit-taxonomy}'s sparse
    cells (especially explicit counterfactual message credit and
    explicit orchestrator credit) are the most tractable research
    targets; compositionality across cells is an open question
    (P1--P4).
  \item \textbf{Benchmarks that measure coordination, not just
    success.} (E1) is a weak discriminator of whether gains come from
    compute or from coordination. An open MAS-native benchmark
    covering (E1)--(E4) at Kimi-reported trace lengths would allow
    credit-assignment methods to be fairly compared for the first
    time (P15).
  \item \textbf{Safe and steerable long-horizon orchestrators.}
    Mid-trace human intervention (P13), inter-agent prompt injection
    (P12), and sandbox-isolation under dynamic spawn (P14) are
    deployment-shaped problems that academic work has only begun to
    address through training-time adversarial games such as
    MAGIC~\cite{magic2026}. They will not stay deferrable: Claw Groups~\cite{kimi-k2-6-2026}
    and the Anthropic trustworthy-agents
    framework~\cite{anthropic-trustworthy2025} already treat them
    as first-class.
\end{enumerate}

The field is now in the window where useful abstractions---the
orchestration trace, the credit- and signal-bearing-unit hierarchy, the
five-way orchestration sub-decision---can still be chosen cleanly,
before conventions calcify around the first generation of
industrial systems. This survey is an argument for why these
abstractions are a productive starting point.

\bibliography{papers}

\appendix
\section{Entry Cards: Core RL Methods and Anchors for LLM-MAS}
\label{app:method-cards}

This appendix gives one-card summaries of thirteen
\emph{core} methods, frameworks, and industrial anchors that are most
frequently referenced by the credit-assignment taxonomy
(Table~\ref{tab:credit-taxonomy}). Each card uses a uniform layout:
(1) one-line claim, (2) reward shape, (3) credit-assignment
mechanism, (4) orchestration form, (5) headline empirical result,
(6) key limitation. Cards are ordered by credit- or signal-bearing unit, from
team/orchestrator to message/tool, mirroring the hierarchy of
\S\ref{sec:credit:hierarchy}.

\newcommand{\methodcard}[7]{%
  \par\medskip\noindent
  \fcolorbox{ThesisBlue}{SoftGray!30!white}{%
    \begin{minipage}{\dimexpr\linewidth-2\fboxsep-2\fboxrule}
      \textbf{\large #1}\hfill\textcolor{ThesisBlue}{\small\itshape #2}\\[2pt]
      \textbf{Claim.} #3\\[1pt]
      \textbf{Reward.} #4\\[1pt]
      \textbf{Credit assignment.} #5\\[1pt]
      \textbf{Orchestration.} #6\\[1pt]
      \textbf{Headline result / limitation.} #7
    \end{minipage}%
  }\par\medskip
}


\subsection{Orchestrator-level credit}

\methodcard
  {Puppeteer~\cite{puppeteer2025}}
  {NeurIPS 2025; Tsinghua / OpenBMB}
  {A learned central orchestrator chooses which sub-agent takes the
   next turn, treating delegation as a learnable action.}
  {Team outcome (R1) only; orchestrator credit comes from a learned
   central critic over orchestrator decisions.}
  {Orchestrator-level: a centralized critic scores each
   delegation against the trace return; sub-agent policies are
   frozen during orchestrator training.}
  {Centralized orchestrator + frozen sub-agents (Regime A,
   \S\ref{sec:orch:regimes}).}
  {Among entries in our pool, the earliest explicit treatment of
   the orchestrator as the unit of RL training. Limitation:
   sub-agent skill is held fixed, so gains are bounded by the
   existing sub-agent pool.}

\methodcard
  {Kimi PARL~\cite{kimi-k2-5-2026,kimi-k2-6-2026}}
  {K2.5 technical report plus K2.6 deployment-envelope extension; Moonshot AI}
  {K2.5 reports PARL training of a learned orchestrator that spawns
   up to $100$ sub-agents and coordinates up to $1{,}500$ reported
   steps / tool-call events. K2.6 extends the public deployment
   envelope to $300$ sub-agents and $4{,}000$ reported coordinated
   steps, but is not used here as an independent RL-training claim.}
  {Composite R7+R8:
   $r_{\text{perf}} + \lambda_1 r_{\text{parallel}}
   + \lambda_2 r_{\text{finish}}$,
   with both auxiliary weights reported in K2.5 as annealed
   (Figure~\ref{fig:parl-annealing}).}
  {Critical-Steps metric distinguishes real parallel progress
   from padded traces; functions as orchestrator-level credit.}
  {Parallel swarm with dynamic spawn; staged training (frozen-then-joint).}
  {Public trained-orchestrator evidence in K2.5,
   with K2.6 used only for deployment-envelope pressure.
   Limitation: full algorithmic details are not public;
   reproducibility is limited.}

\subsection{Role-level credit}

\methodcard
  {MALT~\cite{malt2025}}
  {COLM 2025; University of Oxford}
  {Train a generator--verifier--refiner role triple end-to-end on
   reasoning tasks, with role-specific PRMs.}
  {Role-specific process reward (R3+R4); each role has its own
   PRM rubric.}
  {Role-level: per-role advantage from per-role PRM, summed into
   policy gradient per role.}
  {Planner--executor--critic with explicit role separation.}
  {$+14.14\%$ over single-agent baseline on reasoning tasks.
   Limitation: role rubric design is hand-engineered per task.}

\methodcard
  {M-GRPO~\cite{m-grpo2025}}
  {arXiv (Ant Group); Nov 2025}
  {Hierarchical GRPO that decouples planner (main agent) and
   sub-agent advantage estimation.}
  {Hybrid R8: top-layer team reward, bottom-layer sub-task reward,
   with separate baselines.}
  {Role-level: hierarchical baseline---top and bottom layer
   compute advantages independently against their own group
   baselines.}
  {Centralized orchestrator + sub-agents on deep-research tasks.}
  {Reported AIME $86.5 \to 93.3$ when used as multi-agent deep
   research training. Limitation: assumes a fixed two-level
   hierarchy.}

\methodcard
  {MATPO~\cite{matpo2025}}
  {arXiv (NTU); Oct 2025}
  {A single LLM plays both planner and worker roles with
   role-specific tool integration; the same weights are trained
   for both roles via shared rollouts.}
  {Role-specific reward (R3) plus tool-use reward (R5)
   conditioned on downstream success.}
  {Role-level: dual-role advantage---each rollout contributes
   policy gradient under both role labels with role-specific
   shaping.}
  {Planner--executor with single-LLM weight sharing.}
  {$+18.38\%$ over single-agent baseline. Limitation: dual-role
   training requires careful balancing to avoid one role
   dominating.}

\subsection{Agent-level credit}

\methodcard
  {MAGRPO~\cite{magrpo2025}}
  {arXiv (Northeastern, with C.\ Amato); Aug 2025}
  {Cast LLM collaboration as a cooperative MARL problem and
   propose a multi-agent, multi-turn variant of GRPO.}
  {Shared team reward (R1).}
  {Agent-level: group-relative advantage computed per agent
   within multi-agent rollouts.}
  {Centralized: writing / coding collaboration tasks.}
  {Among entries in our pool, the earliest systematic formalization
   of LLM collaboration as
   cooperative MARL with a matched RL algorithm. Limitation:
   shared-reward setting; free-riding not directly addressed.}

\methodcard
  {MAPoRL~\cite{maporl2025}}
  {ACL 2025; MIT}
  {First post-co-training paradigm explicitly training collaboration
   behavior in LLMs (rather than as an emergent property of
   prompting).}
  {Shared team outcome (R1) broadcast to all participating agents.}
  {Agent-level: PPO-style updates with team reward broadcast;
   no per-agent decomposition.}
  {Centralized; multiple LLMs trained jointly on collaborative tasks.}
  {Establishes the post-training framing now standard in the field.
   Limitation: predates the credit-assignment literature it inspired.}

\methodcard
  {Dr.\,MAS~\cite{dr-mas2026}}
  {arXiv (NTU); Feb 2026}
  {Diagnose multi-agent GRPO instability and propose
   \textbf{agent-wise advantage normalization} as the fix.}
  {Shared team reward (R1).}
  {Agent-level: instead of group-normalizing advantages across
   the rollout group, normalize per-agent within each rollout to
   prevent cross-agent variance from poisoning gradients.}
  {Centralized; general MAS workloads.}
  {Establishes that na\"ive GRPO is unstable when used unchanged
   in multi-agent settings; agent-wise normalization restores
   convergence. Limitation: the fix is empirical---no theoretical
   guarantee.}

\methodcard
  {SHARP~\cite{sharp2026}}
  {arXiv; Feb 2026}
  {Apply Shapley-value credit allocation to multi-agent LLM
   systems with tool-augmented agents.}
  {Hybrid R8: global team reward $+$ Shapley marginal credit per
   agent $+$ tool-process reward.}
  {Agent-level (Shapley) and tool-level: marginal contribution
   of each agent computed via Shapley sampling; tools receive
   process-style rewards.}
  {Hierarchical; tool-augmented multi-agent system.}
  {Most principled credit-attribution method in the pool.
   Limitation: Shapley sampling cost grows combinatorially with
   agent set size; intractable at industrial team sizes
   (\S\ref{sec:open}).}

\subsection{Turn-level credit}

\methodcard
  {MarsRL~\cite{marsrl2025}}
  {arXiv; Nov 2025}
  {Train a multi-reasoning-agent pipeline with RL using
   \textbf{agentic pipeline parallelism} so different rollout
   stages execute concurrently.}
  {Hybrid R8: per-stage rewards along the reasoning pipeline.}
  {Turn-level (per-pipeline-stage): each stage has its own
   advantage signal.}
  {Hierarchical pipeline of reasoning agents.}
  {Demonstrates that pipeline-parallel rollouts amortize the
   cost of multi-agent RL. Limitation: requires reasoning task
   to be cleanly factorable into stages.}

\methodcard
  {Context-Folding~\cite{context-folding2025}}
  {arXiv (ByteDance Seed / CMU); Oct 2025}
  {Agent actively manages its own context by folding sub-trajectories
   back into the main trace, enabling long-horizon agent training.}
  {Hybrid R8: branch reward $\approx$ main reward $\pm 0.2$ scope
   adjustment per folded sub-trajectory.}
  {Turn-level: each fold/unfold action receives advantage
   computed against a shared baseline with independent per-branch
   advantage.}
  {Hierarchical (orchestrator $\to$ branches $\to$ aggregation).}
  {Shows that long-horizon training is feasible with explicit
   context management. Limitation: tied to an agent-specific
   harness; not framework-agnostic.}

\subsection{Message-level credit}

\methodcard
  {C3~\cite{c3-2026}}
  {arXiv (HK PolyU); Mar 2026}
  {The only retained entry in our pool that performs counterfactual
   message-level credit assignment for LLM
   multi-agent systems, using contextual counterfactual
   intervention.}
  {Shared team reward (R1).}
  {Message-level: for each utterance, estimate counterfactual
   trace return under intervention (replacing or removing the
   message). Pivotal messages receive proportionally larger credit.}
  {Centralized; reasoning collaboration.}
  {Only retained entry in our pool that explicitly estimates
   counterfactual message-level credit.
   Limitation: counterfactual estimation cost grows with trace
   length and message count.}

\subsection{Framework-level (cross-cutting)}

\methodcard
  {Agent~Lightning~\cite{agent-lightning2025}}
  {arXiv (Microsoft Research); Aug 2025}
  {Generic RL training framework that decouples agent execution
   from the trainer; supports any agent harness.}
  {Hybrid R8: framework-level reward dispatch with per-agent and
   per-tool-call shaping.}
  {Agent-level and tool-level: framework provides credit-assignment
   primitives; specific decomposition is application-defined.}
  {Harness-based; designed to wrap existing agent runtimes.}
  {Aligns academic RL training with the harness boundary deployed
   in industrial systems (Codex, Claude Code).
   Limitation: a framework, not an algorithm; provides plumbing,
   not an answer to credit assignment per se.}

\section{Paper Pool Summary Table}
\label{app:summary-table}

Table~\ref{tab:pool-summary} lists all $84$ entries in our paper
pool, organized into manuscript evidence buckets rather than by the
raw \texttt{category} field of the retained-entry CSV. These buckets
separate focal LLM-MAS entries from supporting foundations,
benchmarks, safety entries, and critic / verifier references so that
the appendix follows the argument of the paper. The CSV remains the
source of truth for machine-readable counts and controlled
vocabulary; its \texttt{category} histogram is reported in
Table~\ref{tab:coverage-statistics}. Columns abbreviate the taxonomy
tags used throughout the paper; for the full $18$-column schema see
the repository artifact. Entries are sorted within bucket by year,
then by key.

\noindent\textit{Column legend.}
\textbf{RL-rel}: relevance to RL/post-training
(Y = direct RL/post-training method, P = partial/framework/case-level
relevance, N = not RL-centered).
\textbf{Reward}: dominant family from Table~\ref{tab:reward-families}
(shr = shared, ind = individual, role, proc = process, tool,
dbt = debate, verif, orch, hyb = hybrid).
\textbf{Credit}: finest credit level from
\S\ref{sec:credit:hierarchy} (tm, or, ro, ag, tn, msg, tool, tok).
\textbf{Orch}: orchestration form (cntr = centralized,
pec = planner-exec-critic, dbt = debate, swm = swarm,
hier = hierarchical, hrn = harness).
\textbf{Scen}: target scenario (cod = coding, web, rsh = research,
mth = math, tl = tool use, gen = general).

\renewcommand{\arraystretch}{1.15}
\setlength{\tabcolsep}{3pt}
{\footnotesize
\begin{longtable}{@{}p{2.3cm} p{2.0cm} c c c c c c p{3.8cm}@{}}
\toprule
\textbf{Key} & \textbf{Year / venue} & \textbf{RL-rel} & \textbf{Reward} & \textbf{Credit} & \textbf{Orch} & \textbf{Scen} & \textbf{Core?} & \textbf{One-liner} \\
\midrule
\endfirsthead

\multicolumn{9}{@{}l}{\footnotesize\itshape (continued)} \\
\toprule
\textbf{Key} & \textbf{Year / venue} & \textbf{RL-rel} & \textbf{Reward} & \textbf{Credit} & \textbf{Orch} & \textbf{Scen} & \textbf{Core?} & \textbf{One-liner} \\
\midrule
\endhead

\multicolumn{9}{@{}r}{\footnotesize\itshape (continued on next page)} \\
\endfoot

\bottomrule
\endlastfoot

\multicolumn{9}{@{}l}{\textbf{\textit{A. Focal LLM-MAS training, benchmark, and adjacent framework entries} (40 entries)}} \\
\midrule
magrpo            & 2025 / arXiv    & Y & shr  & ag  & cntr & gen & core & LLM collab as coop MARL; MAGRPO \\
marft             & 2025 / arXiv    & Y & hyb  & ag  & cntr & gen & core & Multi-agent reinforcement fine-tuning \\
m-grpo            & 2025 / arXiv    & Y & hyb  & ro  & hier & rsh & core & Hierarchical GRPO, decoupled planner/sub \\
agent-lightning   & 2025 / arXiv    & Y & hyb  & ag  & hrn  & gen & core & Generic RL framework; exec--train decoupling \\
matpo             & 2025 / arXiv    & Y & role & ro  & pec  & tl  & core & Dual-role planner+worker; tool-integrated PO \\
puppeteer         & 2025 / NeurIPS  & Y & orch & or  & cntr & gen & core & Learned central orchestrator \\
halo              & 2025 / arXiv    & P & orch & ro  & hier & gen & supp & MCTS-based 3-layer hierarchical MAS \\
marsrl            & 2025 / arXiv    & Y & hyb  & tn  & hier & mth & core & Agentic pipeline-parallel RL \\
malt              & 2024 / COLM'25  & Y & role & ro  & pec  & mth & core & Generator-verifier-refiner w/ role-PRM \\
maporl            & 2025 / ACL      & Y & shr  & ag  & cntr & gen & core & Post-co-training RL for collaboration \\
latentmas         & 2025 / arXiv    & N & NA   & NA  & dbt  & gen & supp & Training-free latent-space MAS \\
mae               & 2025 / arXiv    & Y & verif& ro  & pec  & gen & supp & Proposer-solver-judge co-evolution \\
context-folding   & 2025 / ICLR'26s & Y & hyb  & tn  & hier & cod & core & Agent-managed context folding \\
dr-mas            & 2026 / arXiv    & Y & shr  & ag  & cntr & gen & core & Diagnose GRPO instability; agent-wise norm \\
c3                & 2026 / arXiv    & Y & shr  & msg & cntr & gen & core & Contextual counterfactual msg-level credit \\
sharp             & 2026 / arXiv    & Y & hyb  & ag  & hier & tl  & core & Shapley-value credit allocation \\
debate-as-reward  & 2026 / arXiv    & Y & dbt  & msg & dbt  & rsh & core & Multi-agent debate as RL reward \\
paramanager       & 2026 / arXiv    & Y & orch & or  & cntr & gen & core & Small-model master orchestrator \\
hera              & 2026 / arXiv    & P & orch & or  & cntr & rsh & supp & Evolving orch policy for MAS-RAG \\
tamas             & 2025 / ICML'25W & N & NA   & NA  & NA   & gen & supp & Adversarial robustness benchmark for MAS \\
agentspawn        & 2026 / arXiv    & P & orch & or  & swm  & cod & supp & Runtime dynamic spawn + memory transfer \\
rema2025          & 2025 / NeurIPS  & Y & hyb  & ro  & hier & mth & core & Meta-thinking + reasoning agents via MARL \\
collabui\hspace{0pt}agents2025 & 2025 / COLM    & Y & proc & ag  & cntr & web & core & Credit re-assignment for UI/web generalization \\
comas2026         & 2026 / ICLR     & Y & dbt  & ag  & dbt  & gen & core & Co-evolution via interaction rewards \\
owl2025           & 2025 / NeurIPS  & Y & hyb  & or  & hier & rsh & core & Optimized Workforce planner learning \\
sirius2025        & 2025 / NeurIPS  & P & proc & tn  & hier & gen & supp & Bootstrapped reasoning experience library \\
multiagent-finetuning2025 & 2025 / ICLR & P & dbt & ag & dbt & mth & supp & Multiagent self-improvement via diverse chains \\
mas-zero2025      & 2025 / arXiv    & N & NA   & or  & hier & gen & supp & Zero-supervision inference-time MAS design \\
learning-to-deliberate2025 & 2025 / arXiv & Y & hyb & tn & dbt & mth & core & Meta-policy deliberation actions + SoftRankPO \\
collm-maac2026    & 2026 / arXiv    & Y & shr  & ag  & cntr & gen & core & Actor-critic decentralized LLM collaboration \\
wideseek-r1-2026  & 2026 / arXiv    & Y & orch & or  & hier & rsh & core & Width scaling with lead/subagent MARL \\
magic2026         & 2026 / arXiv    & Y & dbt  & ag  & dbt  & gen & core & Attacker-defender MARL for safety \\
marti-mars2-2026  & 2026 / arXiv    & Y & hyb  & ag  & hier & cod & core & Multi-agent self-search RL for code \\
spiral2026        & 2026 / ICLR     & Y & hyb  & ro  & dbt  & gen & core & Online self-play with role-conditioned advantage \\
marshal2026       & 2026 / ICLR     & Y & hyb  & tn  & dbt  & gen & core & Strategic self-play with turn-level advantage \\
depart2026        & 2026 / OpenReview & Y & hyb & ro & hier & web & core & HIMPO planner/executor post-training \\
agent-qmix2026    & 2026 / arXiv    & Y & hyb  & ag  & cntr & gen & core & QMIX topology and communication learning \\
langmarl2026      & 2026 / arXiv    & Y & hyb  & ag  & cntr & gen & core & Language-space agent credit assignment \\
lamo2026          & 2026 / ACL Findings & Y & role & ro & hier & web & core & Lightweight GUI multi-role orchestration \\
sage2026          & 2026 / ARR      & Y & verif& ro  & pec  & mth & core & Challenger-planner-solver-critic co-evolution \\
\midrule

\multicolumn{9}{@{}l}{\textbf{\textit{B. Related surveys used for gap analysis} (5 entries)}} \\
\midrule
survey-mas        & 2024 / arXiv & P & NA & NA & NA & gen & core & LLM-MAS architecture survey \\
survey-collab     & 2025 / arXiv & N & NA & NA & NA & gen & core & MAS collaboration mechanisms \\
survey-rl-meets-llm & 2025 / arXiv & Y & NA & NA & NA & gen & supp & RL across LLM lifecycle \\
survey-agentic-rl & 2025 / TMLR  & Y & NA & NA & NA & gen & core & 500+ works on agentic RL \\
survey-agentic-\hspace{0pt}reasoning & 2026 / arXiv & P & NA & NA & NA & gen & supp & Agentic reasoning roadmap \\
\midrule

\multicolumn{9}{@{}l}{\textbf{\textit{C. Industrial systems (cases)} (6 entries)}} \\
\midrule
kimi-k2-5         & 2026 / Tech rep. & Y & orch & or & swm & rsh & case & Agent Swarm + PARL \\
kimi-k2-6         & 2026 / Tech blog & P & NA & NA & swm & cod & case & Deployment-scale evidence; Claw Groups \\
openai-codex      & 2025 / Blog      & P & NA   & NA & hrn & cod & case & Cloud-parallel SE agent \\
claude-code-subagents & 2025 / Docs  & N & NA   & NA & hrn & cod & case & Claude Code sub-agent API \\
anthropic-trustworthy & 2025 / Blog  & N & NA   & NA & NA  & gen & case & Safe and trustworthy agent framework \\
anthropic-c-compiler & 2026 / Eng.blog & N & NA & NA & swm & cod & case & 16 parallel Claudes build C compiler \\
\midrule

\multicolumn{9}{@{}l}{\textbf{\textit{D. Classical MARL (conceptual toolkit)} (10 entries)}} \\
\midrule
markov-games1994  & 1994 / ICML   & Y & ind & ag & NA & gen & supp & Markov games as MARL framework \\
difference-rewards2001 & 2001 / ACS & -- & NA & -- & NA & gen & supp & Difference rewards / WLU \\
dec-pomdp2002     & 2002 / MOR    & N & NA  & -- & NA & gen & supp & Complexity of decentralized control \\
maddpg2017        & 2017 / NeurIPS & Y & ind & ag & NA & gen & supp & Multi-agent DDPG (CTDE) \\
coma2018          & 2018 / AAAI   & Y & shr & ag & NA & gen & supp & Counterfactual multi-agent PG \\
vdn2018           & 2018 / AAMAS  & Y & shr & ag & NA & gen & supp & Value decomposition networks \\
qmix2018          & 2018 / ICML   & Y & shr & ag & NA & gen & supp & Monotonic value factorisation \\
ippo2020          & 2020 / arXiv  & Y & ind & ag & NA & gen & supp & Independent PPO in StarCraft \\
shapley-q2020     & 2020 / AAAI   & Y & shr & ag & NA & gen & supp & Shapley Q-value \\
mappo2022         & 2022 / NeurIPS D\&B & Y & shr & ag & NA & gen & supp & Surprising effectiveness of MAPPO \\
\midrule

\multicolumn{9}{@{}l}{\textbf{\textit{E. Benchmarks cited in Sec.~\ref{sec:benchmarks}} (8 entries)}} \\
\midrule
gaia2023          & 2023 / arXiv  & N & NA & NA & NA & rsh & supp & General AI assistants benchmark \\
toolbench2023     & 2023 / ICLR'24 & N & NA & NA & NA & tl & supp & Tool-learning 16k+ APIs \\
swe-bench2024     & 2024 / ICLR  & N & NA & NA & NA & cod & supp & Real-world GitHub issues \\
webarena2024      & 2024 / ICLR  & N & NA & NA & NA & web & supp & Realistic web agent environment \\
tau-bench2024     & 2024 / arXiv & N & NA & NA & NA & tl & supp & Tool-agent-user interaction \\
osworld2024       & 2024 / NeurIPS D\&B & N & NA & NA & NA & cod & supp & Multimodal computer-use tasks \\
browse\-comp2025    & 2025 / arXiv & N & NA & NA & NA & web & supp & OpenAI browsing benchmark \\
multi\-agent\-bench2025 & 2025 / ACL & N & NA & NA & NA & gen & supp & MAS collab+competition benchmark \\
\midrule

\multicolumn{9}{@{}l}{\textbf{\textit{F. Single-agent RL \& LLM-RL foundations} (5 entries)}} \\
\midrule
ppo2017           & 2017 / arXiv  & Y & NA & NA & NA & gen & supp & PPO (Schulman et al.) \\
instructgpt2022   & 2022 / NeurIPS & Y & NA & NA & NA & gen & supp & RLHF / InstructGPT \\
react2023         & 2023 / ICLR   & N & NA & NA & NA & gen & supp & Reasoning + acting prompting \\
deep\-seekmath2024  & 2024 / arXiv  & Y & NA & NA & NA & mth & supp & DeepSeekMath; introduces GRPO \\
deepseek-r12025   & 2025 / Nature & Y & NA & NA & NA & gen & supp & Incentivizing reasoning via RL \\
\midrule

\multicolumn{9}{@{}l}{\textbf{\textit{G. Safety / Adversarial Robustness (cited in Sec.~\ref{sec:safety})} (6 entries)}} \\
\midrule
greshake2023         & 2023 / AISec     & N & NA & NA & NA  & gen & supp & Indirect prompt injection (foundational) \\
agentdojo2024        & 2024 / NeurIPS'24 D\&B & N & NA & NA & hrn & tl  & supp & 97 tasks + 629 security cases \\
injecagent2024       & 2024 / ACL Findings    & N & NA & NA & NA  & tl  & supp & 1,054 IPI test cases \\
agents-under-\hspace{0pt}siege2025 & 2025 / ACL     & N & NA & NA & NA  & gen & supp & Permutation-invariant topology attacks \\
tmcht2025            & 2025 / ACL       & N & NA & NA & cntr & gen & supp & Contagious jailbreak via memory poisoning \\
wasp2025             & 2025 / arXiv     & N & NA & NA & hrn & web & supp & Web-agent prompt-injection benchmark \\
\midrule

\multicolumn{9}{@{}l}{\textbf{\textit{H. Critic / tool-use evaluation cited in Sec.~\ref{sec:rewards}--\ref{sec:benchmarks}} (4 entries)}} \\
\midrule
mtu-bench2025        & 2025 / ICLR    & N & NA   & NA & NA & tl  & supp & 5-granularity tool-use benchmark \\
codecritic\hspace{0pt}bench2025 & 2025 / arXiv & N & verif & NA & NA & cod & supp & Code critique benchmark; checklists \\
artifacts\hspace{0pt}bench2025 & 2025 / arXiv & N & verif & NA & NA & cod & supp & MLLM-as-Judge over 1,825 visual code tasks \\
criticlean2025       & 2025 / arXiv   & Y & verif & NA & NA & mth & supp & RL-trained critic for Lean 4 formalization \\

\caption{Complete paper pool ($84$ entries), organized into eight
manuscript evidence buckets. Entries carry one of three status labels in the
\textbf{Core?} column: \textbf{core} marks entries central to the
survey's argument (including the thirteen method / framework /
industrial-anchor cards in Appendix~\ref{app:method-cards}, plus the
most directly adjacent surveys);
\textbf{case} marks industrial cases that motivate the survey's
framing but are not algorithmic contributions;
\textbf{supp} marks entries that support the taxonomy (classical
MARL, benchmarks, safety, single-agent foundations, critic /
tool-use evaluations, remaining surveys) without being themselves
central contributions. The \textbf{core} label is therefore a
\emph{relevance} flag for the survey's argument, not a synonym for
``RL method''---industrial cases are flagged separately, and
several classical-MARL entries (e.g., COMA, Shapley-Q) remain
\textbf{supp} despite being RL works. ``--'' denotes a field not
applicable (e.g., credit granularity for pre-LLM classical works).
The retained-entry CSV in the artifact repository is the authoritative
source and carries nine additional columns omitted here for space;
the bucket labels in this appendix are for readability and are not a
replacement for the CSV's controlled \texttt{category} field.}
\label{tab:pool-summary} \\
\end{longtable}
}

\noindent\textbf{Counts by manuscript evidence bucket.}
A: 40 entries (focal LLM-MAS training, benchmark, and adjacent framework entries);
B: 5 entries (related surveys);
C: 6 entries (industrial cases);
D: 10 entries (classical MARL);
E: 8 entries (benchmarks);
F: 5 entries (single-agent foundations);
G: 6 entries (safety / adversarial robustness);
H: 4 entries (critic / tool-use evaluation).
These buckets intentionally differ from the CSV \texttt{category}
histogram because some entries play a different evidence role in the
manuscript than their coarse source type would suggest. For
machine-readable counts, use the artifact script and the CSV fields
directly.

\section{Artifact, Search Protocol, and Trace Schema}
\label{app:artifact-protocol}

This appendix records the manuscript-level evidence protocol behind
the paper pool. It is included to make the taxonomy easier to audit and
extend. The protocol is quasi-systematic: it exposes the search
families, screened records, exclusion stages, tag schema, and
borderline decisions, but it does not claim PRISMA-level
reproducibility because raw database exports, deduplication logs, and
multi-annotator agreement are not provided.

The supplementary artifact snapshot and repository (\artifactrepo),
with repository-path normalization between the manuscript bundle
(\texttt{artifact/}) and the GitHub layout (\texttt{scripts/},
\texttt{trace-schema/}, and \texttt{docs/}), turn the appendix into a
reusable object rather than only prose. They contain a retained-entry
CSV, an exclusion-log CSV, a statistics script that regenerates all
corpus counts and cross-tabs, a
machine-readable orchestration-trace schema used in
\S\ref{sec:bench:good}, a minimal valid trace, and a dependency-free
trace validator.

\subsection{Search strings and screening counts}

The search cutoff for this version is May 4, 2026. Searches were
performed over arXiv, ACL Anthology, OpenReview, Semantic Scholar /
citation links, official project pages, company technical reports, and
product documentation. The representative query families were:

\begin{itemize}
  \item \texttt{multi-agent LLM} $\times$ \texttt{reinforcement learning}
  \item \texttt{multi-agent LLM} $\times$ \texttt{post-training}
  \item \texttt{multi-agent LLM} $\times$ \texttt{credit assignment}
  \item \texttt{LLM agent} $\times$ \texttt{orchestration} /
        \texttt{agent swarm} / \texttt{dynamic spawn}
  \item \texttt{tool-use LLM} $\times$ \texttt{reinforcement learning}
  \item \texttt{LLM agent} $\times$ \texttt{prompt injection} /
        \texttt{safety} / \texttt{jailbreak}
  \item \texttt{multi-agent reinforcement learning} $\times$
        \texttt{communication} / \texttt{credit assignment}
\end{itemize}

\begin{table}[t]
\centering
\small
\renewcommand{\arraystretch}{1.12}
\begin{tabularx}{\linewidth}{@{}p{3.0cm}p{2.0cm}X@{}}
\toprule
\textbf{Screening stage} & \textbf{Records} & \textbf{Notes} \\
\midrule
Candidate records considered
 & 116
 & Union of seed surveys, keyword search, citation following, and public-system audit. \\
Retained entries
 & 84
 & Tagged in the retained-entry CSV with 18 fields and cited in the manuscript. \\
Excluded records
 & 32
 & Logged in the exclusion-log CSV; all rows include a public identifier or documentation handle and URL. \\
Abstract-screen exclusions
 & 16
 & Excluded because the abstract/public page did not expose LLM-MAS RL, credit, orchestration, benchmark, safety, or load-bearing formalism. \\
Full-text-screen exclusions
 & 9
 & Read beyond abstract but not retained as taxonomy evidence. \\
Artifact-screen exclusions
 & 6
 & Documentation, code, or product pages without enough methodological detail for coding. \\
Duplicate / overlapping signal
 & 1
 & Excluded because the retained pool already contained the load-bearing benchmark/framework signal. \\
\bottomrule
\end{tabularx}
\caption{Screening counts for the $116$ audited records. Counts are
manuscript audit counts, not raw search-engine result counts.}
\label{tab:appendix-screening-counts}
\end{table}

\subsection{Tag schema}

Each retained row in the retained-entry CSV has 18 fields:
\texttt{key}, \texttt{title}, \texttt{first\_author},
\texttt{affiliation}, \texttt{year}, \texttt{arxiv\_id},
\texttt{venue}, \texttt{url}, \texttt{category}, \texttt{is\_rl},
\texttt{reward\_type}, \texttt{credit\_granularity},
\texttt{orchestration\_form}, \texttt{scenario}, \texttt{is\_core},
\texttt{one\_liner}, \texttt{verified}, and \texttt{notes}. The most
important controlled fields are:

\begin{table}[t]
\centering
\small
\renewcommand{\arraystretch}{1.12}
\begin{tabularx}{\linewidth}{@{}p{4.1cm}X@{}}
\toprule
\textbf{Field} & \textbf{Controlled values / interpretation} \\
\midrule
\texttt{category}
 & \texttt{rl\_method}, \texttt{survey}, \texttt{benchmark}, \texttt{framework}, \texttt{industry}, \texttt{classical\_marl}. \\
\texttt{is\_rl}
 & \texttt{yes} for explicit RL/post-training, \texttt{partial} for self-evolution, RL-adjacent optimization, or industrial deployment evidence that constrains RL design without independently disclosing a training objective, \texttt{no} for background/safety/benchmark/system entries. \\
\texttt{reward\_type}
 & One of the reward families R1--R8, or \texttt{NA} when the entry is not a reward-design method. \\
\texttt{credit\_granularity}
 & Finest level at which the entry exposes a reward, credit, or design signal: token, turn, message, tool, agent, role, orchestrator, team, or \texttt{NA}. Explicit credit-assignment mechanisms are a narrower subset discussed in \S\ref{sec:credit}; for example, a message-level debate reward is tagged at the message level, while C3 is the retained entry with counterfactual message-level credit. \\
\texttt{orchestration\_form}
 & centralized, planner--executor--critic, debate, swarm, hierarchical, harness, or \texttt{NA}. \\
\texttt{verified}
 & \texttt{yes} when bibliographic metadata and public artifact were checked; \texttt{partial} when the entry depends on a submission/project page or evolving public material. \\
\bottomrule
\end{tabularx}
\caption{Tagging schema used by the retained pool. Ambiguous entries
are intentionally marked \texttt{partial} rather than forced into
binary labels.}
\label{tab:appendix-tag-schema}
\end{table}

\paragraph{Partial and evolving evidence.}
Eight retained entries have \texttt{verified=partial}: ReMA,
CollabUIAgents, CoMAS, OWL, SiriuS, Multiagent Finetuning, DEPART,
and SAGE. These
entries were added during the coverage audit from OpenReview or
project-page material and are used to populate adjacent taxonomy
cells, not as sole support for the paper's central claims. Kimi K2.6
is instead \texttt{verified=yes} but \texttt{is\_rl=partial}: the
public material is stable enough to support deployment-envelope
claims, while its row is not used as an independent RL-training
claim.

\subsection{Borderline decisions}

\begin{table}[t]
\centering
\small
\renewcommand{\arraystretch}{1.12}
\begin{tabularx}{\linewidth}{@{}p{2.6cm}p{1.6cm}X@{}}
\toprule
\textbf{Record} & \textbf{Decision} & \textbf{Reason} \\
\midrule
AutoGen / CAMEL / MetaGPT
 & exclude
 & Important LLM-MAS frameworks, but the screened records do not provide RL/post-training or credit-assignment mechanisms used as load-bearing evidence here. \\
MAS-Zero
 & retain
 & Not RL in the strict sense, but directly relevant to zero-supervision MAS design and orchestrator search. \\
SiriuS / Multiagent Finetuning
 & retain
 & Self-evolution and interaction-generated training signals are adjacent to RL and affect the reward taxonomy. \\
Kimi K2.5
 & retain
 & Public industrial source explicitly discloses RL training of the orchestrator; reproducibility remains limited. \\
Codex / Claude Code
 & retain as cases
 & Used only for deployment-shape and harness evidence, not as public multi-agent RL training evidence. \\
AgentBench / AgentBoard
 & exclude
 & Strong agent benchmarks, but less directly tied to MAS-native E2--E4 instrumentation than retained benchmark entries. \\
\bottomrule
\end{tabularx}
\caption{Representative borderline inclusion/exclusion decisions.
These examples are included so that future versions can revise the
pool consistently rather than silently changing the taxonomy boundary.}
\label{tab:borderline-decisions}
\end{table}

\subsection{Evidence matrix}

\begin{figure}[t]
\centering
\small
\renewcommand{\arraystretch}{1.12}
\begin{tabularx}{0.96\linewidth}{@{}p{3.1cm}ccccc@{}}
\toprule
\textbf{Source class} & \textbf{Alg.} & \textbf{Train} & \textbf{Deploy} & \textbf{Scale} & \textbf{Reprod.} \\
\midrule
Peer-reviewed / arXiv methods & yes & partial & partial & partial & partial \\
OpenReview submissions / posters & partial & partial & partial & partial & limited \\
Company technical reports & partial & partial & yes & yes & limited \\
Product docs / launch blogs & no & no & yes & partial & no \\
Engineering case studies & no & no & yes & partial & no \\
Benchmarks / leaderboards & no & no & partial & partial & partial \\
\bottomrule
\end{tabularx}
\caption{Evidence-level matrix used when interpreting the corpus.
``Alg.'' denotes support for algorithmic mechanism claims; ``Train''
denotes public training-objective or post-training evidence; ``Deploy''
denotes deployment-shape evidence; ``Scale'' denotes public scale or
horizon evidence. This matrix prevents product documentation from being
treated as equivalent to reproducible algorithmic evidence.}
\label{fig:evidence-matrix}
\end{figure}

\subsection{Claim-to-artifact ledger}

Table~\ref{tab:claim-artifact-ledger} records how the main empirical
claims in the survey can be checked against the artifact. The ledger
is intentionally narrow: it ties claims to retained-pool fields,
scripted counts, or explicitly bounded industrial evidence, and it
does not turn the curated pool into a field-wide prevalence estimate.

\begin{table}[t]
\centering
\small
\renewcommand{\arraystretch}{1.15}
\begin{tabularx}{\linewidth}{@{}p{3.0cm}p{3.4cm}X@{}}
\toprule
\textbf{Claim} & \textbf{Artifact check} & \textbf{Boundary} \\
\midrule
Message-level credit is sparse
 & Message-level tag count is two; only C3 is counterfactual message credit
 & Tag count includes message-level reward/signal, not only explicit credit mechanisms. \\
Orchestrator-level credit is sparse
 & Orchestrator-level tag count is eight after K2.6 is treated as deployment evidence and WideSeek-R1 is added
 & Explicit RL credit mechanisms are narrower than orchestrator-level design or evolution signals. \\
Kimi provides the public trained-orchestrator anchor
 & K2.5 is tagged as explicit RL with orchestration reward; K2.6 is tagged as partial / NA
 & K2.6 is used for deployment-envelope pressure, not as an independent training claim. \\
Open MAS-native evaluation remains incomplete
 & Benchmark rows and Table~\ref{tab:benchmarks}; trace schema and reporting checklist in \S\ref{sec:bench:good}
 & The claim is restricted to open, auditable retained entries under the stated protocol. \\
Trace reporting is mechanically inspectable
 & Trace schema, example trace, and structural checker
 & The checker validates core structural constraints, not the full JSON Schema standard. \\
\bottomrule
\end{tabularx}
\caption{Claim-to-artifact ledger. Each row identifies the artifact
object that supports a central survey claim and the boundary on that
claim.}
\label{tab:claim-artifact-ledger}
\end{table}

Because \texttt{credit\_granularity} records the finest level at which
an entry exposes a reward, credit, or design signal, it is broader than
``explicit counterfactual credit.'' Table~\ref{tab:sparse-tag-rationale}
therefore lists the retained rows behind the two sparsest credit cells.

\begin{table}[t]
\centering
\scriptsize
\renewcommand{\arraystretch}{1.08}
\begin{tabularx}{\linewidth}{@{}p{2.35cm}p{5.0cm}X@{}}
\toprule
\textbf{Tagged row} & \textbf{Tag rationale} & \textbf{Mechanism boundary} \\
\midrule
puppeteer
 & Learned central critic over orchestrator delegation.
 & Explicit RL credit. \\
paramanager
 & Unified agent/tool orchestration action space.
 & Orchestrator-level design signal; not counterfactual credit. \\
hera
 & Evolving orchestration policy and prompts.
 & Evolution signal over orchestration choices. \\
agentspawn
 & Runtime spawn decisions and memory transfer.
 & Runtime design signal; no disclosed RL credit estimator. \\
kimi-k2-5
 & PARL with Critical-Steps reward for Agent Swarm.
 & Explicit public training signal; full traces are not released. \\
owl2025
 & Planner/workforce optimization for modular agents.
 & Planner-level training signal. \\
mas-zero2025
 & Meta-level MAS design feedback and verification.
 & Non-RL design-search signal. \\
wideseek-r1-2026
 & Lead-agent/subagent width scaling with MARL.
 & Explicit orchestration-training signal, but not dynamic-spawn counterfactual credit. \\
c3
 & Counterfactual causal credit at message level.
 & Explicit counterfactual message credit. \\
debate-as-reward
 & Debate messages supply a reward signal.
 & Message-level reward signal; not counterfactual credit. \\
\bottomrule
\end{tabularx}
\caption{Rationale for sparse credit-granularity tags. The table makes
the distinction between tag level and explicit credit mechanism
auditable for the eight orchestrator-tagged rows and two
message-tagged rows.}
\label{tab:sparse-tag-rationale}
\end{table}

\subsection{Scripted meta-analysis}

The headline counts in Table~\ref{tab:coverage-statistics} and the
reward--credit cross-tab in Table~\ref{tab:reward-credit-crosstab}
are generated from the CSV artifact rather than hand-entered during
analysis. Running the repository statistics script prints
retained/excluded counts, controlled-field histograms, and
three cross-tabs:
\texttt{reward\_type}~$\times$~\texttt{credit\_granularity},
\texttt{orchestration\_form}~$\times$~\texttt{credit\_granularity},
and \texttt{category}~$\times$~\texttt{verified}. The first cross-tab
is reproduced in the introduction because it directly supports the
paper's claim that message-level and orchestrator-level tags are
sparse, and that explicit message/orchestrator credit mechanisms are
rarer still.

\begin{table}[t]
\centering
\small
\renewcommand{\arraystretch}{1.08}
\begin{tabular}{@{}lrrrrrr@{}}
\toprule
\textbf{Orchestration form} & \textbf{NA} & \textbf{agent} & \textbf{msg.} & \textbf{orch.} & \textbf{role} & \textbf{turn} \\
\midrule
centralized & 0  & 14 & 1 & 3 & 0 & 0 \\
hierarchical & 0 & 2  & 0 & 3 & 5 & 3 \\
debate & 1       & 3  & 1 & 0 & 1 & 2 \\
swarm & 2        & 0  & 0 & 2 & 0 & 0 \\
harness & 2      & 1  & 0 & 0 & 0 & 0 \\
planner--executor--critic & 0 & 0 & 0 & 0 & 4 & 0 \\
NA & 31          & 3  & 0 & 0 & 0 & 0 \\
\bottomrule
\end{tabular}
\caption{Orchestration-form by credit-granularity cross-tab generated
from the artifact. Centralized entries dominate agent-level credit,
while swarm entries split between deployment-shape evidence and
orchestrator-level signals. This supports the paper's claim that
topology constrains which credit unit can be made explicit.}
\label{tab:orch-credit-crosstab}
\end{table}

\subsection{Machine-readable trace schema}

The JSON Schema in the artifact repository encodes the
minimal trace object used by \S\ref{sec:bench:good}. It requires
\texttt{trace\_id}, \texttt{task\_id}, \texttt{events}, \texttt{edges},
\texttt{rewards}, and \texttt{costs}. Event types include
\texttt{orchestrator\_decision}, \texttt{spawn}, \texttt{despawn},
\texttt{message}, \texttt{tool\_call}, \texttt{tool\_result},
\texttt{return}, \texttt{aggregate}, \texttt{human\_intervention}, and
\texttt{safety\_event}. Edges distinguish temporal, causal, spawn,
message, tool-dependency, return, aggregation, and safety-flow
relations.

The schema deliberately stores prompt/tool content through
\texttt{content\_ref} rather than requiring raw content. This lets a
benchmark release replayable trace topology, reward channels, cost
metadata, and safety-flow information while redacting private prompts,
credentials, or tool outputs. The goal is not to standardize every
agent framework, but to define the minimum object needed to recompute
the evaluation quantities in Table~\ref{tab:benchmark-metric-defs}
and the reporting checklist in Table~\ref{tab:trace-reporting-checklist}.
The example trace can be checked with the repository validator.

\subsection{Artifact update protocol}

Future updates should add new records in three steps: (1) append the
candidate to the exclusion-log CSV with identifier, URL, screening
stage, and exclusion reason; (2) promote it to the retained-entry CSV
only if it changes a reward, credit, orchestration, evaluation,
safety, industrial-evidence, or formalism cell; (3) update
Figure~\ref{fig:taxonomy-heatmap},
Table~\ref{tab:coverage-statistics},
Table~\ref{tab:reward-credit-crosstab}, and the appendix summary table
by rerunning the repository statistics script if the promoted entry
changes a controlled tag count. This keeps the pool curated while
making curation decisions auditable.

\end{document}